\PassOptionsToPackage{hyphens}{url}
\documentclass[10pt, twocolumn, copyright, gr]{google}

\usepackage[authoryear, sort&compress, round]{natbib}
\usepackage[T1]{fontenc}

\usepackage{microtype}
\usepackage{graphicx}
\usepackage{subcaption}
\usepackage{booktabs} 
\usepackage{needspace}

\usepackage{hyperref}

\usepackage{inconsolata}
\usepackage[smartEllipses]{markdown}
\usepackage{xfrac}
\usepackage{colortbl}
\usepackage{adjustbox}
\usepackage{diagbox}
\usepackage{amsmath,amssymb, amsthm, mathtools}
\usepackage{array,multirow,makecell}
\usepackage{tabularx}
\usepackage[skins]{tcolorbox}
\tcbuselibrary{breakable}
\usepackage{anyfontsize}
\usepackage{framed}
\usepackage{tabto}
\usepackage{tabulary}
\usepackage{listings}
\usepackage{nccmath}

\usepackage[toc,page,header]{appendix}
\usepackage{minitoc}

\theoremstyle{plain}

\theoremstyle{definition}

\theoremstyle{remark}

\newtcolorbox[auto counter, number within=section]{prompt}[3][]{%
  enhanced,
  breakable,
  colback=#2!5!white,
  colframe=#2!75!black,
  title=\textbf{Box \thetcbcounter: #3},
  fontupper=\footnotesize\fontfamily{cmr}\selectfont,
  #1
}

\definecolor{cCompletion}{HTML}{9E0142}
\definecolor{cClosedBook}{HTML}{D53E4F}
\definecolor{cMultipleChoice}{HTML}{F46D43}
\definecolor{cFull}{HTML}{FDAE61}
\definecolor{cMinimal}{HTML}{FEE08B}
\definecolor{cExact}{HTML}{FFFFBF}
\definecolor{cRephrased}{HTML}{E6F598}
\definecolor{cHigh}{HTML}{ABDDA4}
\definecolor{cNatural}{HTML}{66C2A5}
\definecolor{cDirect}{HTML}{3288BD}
\definecolor{cReverse}{HTML}{5E4FA2}

\newcommand{\wikiprofile}{\textsc{WikiProfile}}

\uselogo{} 

\title{Empty Shelves or Lost Keys? Recall Is the Bottleneck for Parametric Factuality}

\correspondingauthor{nitayc@google.com, galyona@google.com. \textbf{Published in the Proceedings of the 43$^{rd}$ International Conference on Machine Learning (ICML 2026)}. \wikiprofile{} is available at \url{https://huggingface.co/datasets/google/WikiProfile}.}

\reportnumber{} 

\renewcommand{\today}{}

\author[1,2]{Nitay Calderon}
\author[1]{Eyal Ben-David}
\author[1]{Zorik Gekhman}
\author[1]{Eran Ofek}
\author[1]{Gal Yona}

\affil[1]{\thepa{}{}}
\affil[2]{Technion -- Israel Institute of Technology}

\begin{abstract}
Standard factuality evaluations of LLMs treat all errors alike, obscuring whether failures arise from missing knowledge (empty shelves) or from limited access to encoded facts (lost keys). We propose a behavioral framework that profiles factual knowledge at the level of facts rather than questions, characterizing each fact by whether it is encoded, and then by how accessible it is: cannot be recalled, can be directly recalled, or can only be recalled with inference-time computation (thinking). To support such profiling, we introduce WikiProfile, a new benchmark constructed via an automated pipeline with a prompted LLM grounded in web search. Across 4 million responses from 13 LLMs, we find that encoding is nearly saturated in frontier models on our benchmark, with GPT-5 and Gemini-3 encoding 95--98\% of facts. However, recall remains a major bottleneck: many errors previously attributed to missing knowledge instead stem from failures to access it.
These failures are systematic and disproportionately affect long-tail facts and reverse questions.
Finally, we show that thinking improves recall and can recover a substantial fraction of failures, indicating that future gains may rely less on scaling and more on methods that improve how models utilize what they already encode.
\end{abstract}

\begin{document}
\maketitle



\doparttoc
\faketableofcontents


\begin{figure}[!h]
    \centering
    \includegraphics[width=0.48\textwidth]{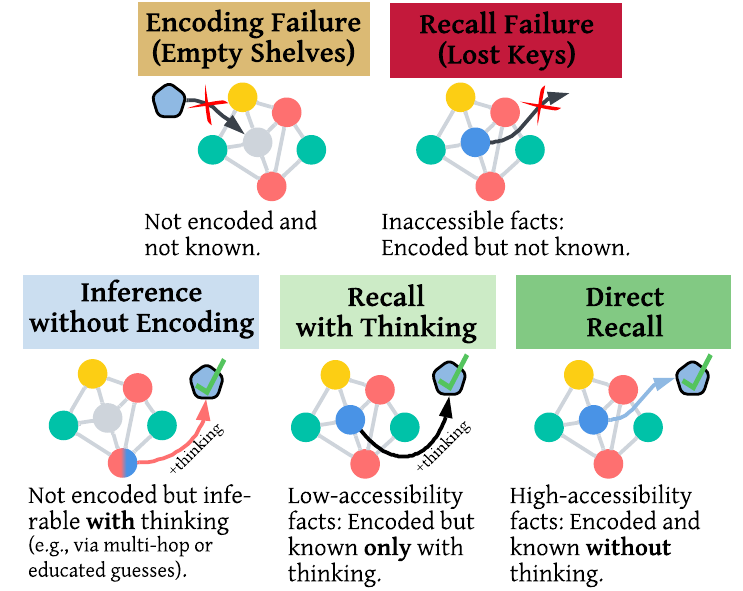}
    \par
    \vspace{-0.2em}
    \includegraphics[width=0.47\textwidth]{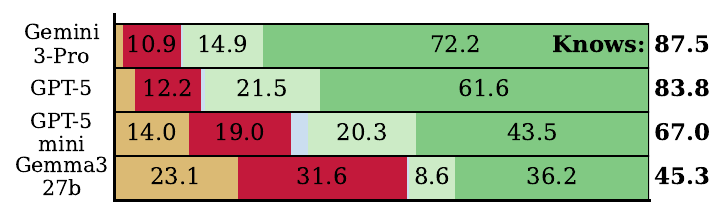}
    \caption{\emph{Top:} We propose five knowledge profiles that characterize facts. \emph{Bottom:} Percentages of these profiles across selected LLMs, revealing: (1) Scaling fills ``empty shelves'' by reducing encoding failures: \textbf{frontier LLMs encode nearly all facts} in our data. (2) \textbf{Recall failures remain abundant} despite scaling, leaving substantial room for improvement. 
    (3) \textbf{Thinking acts as a recovery mechanism} of facts that would otherwise remain \emph{``lost''}.
    }
    \label{fig:intro_fig}
\vspace{-0.8em}
\end{figure}

\section{Introduction}
\label{sec:intro}

Large Language Models (LLMs) encode a vast amount of factual information in their parameters \citep{triviaqa, PetroniRRLBWM19, KwiatkowskiPRCP19, MallenAZDKH23}, yet factual errors remain a persistent challenge \citep{simpleqa, simpleqa_ver, RavichanderGW025}. While existing evaluations consider a response as either correct or incorrect, accuracy alone provides little insight into the source of the error. If an LLM incorrectly answers the question \textit{``Which famous band played their first gig at the Boardwalk club?''} (the answer is Oasis), can we tell why? One possibility is that the LLM \emph{does not encode} this information in its parameters (e.g., due to limited data coverage or model capacity). Alternatively, a fact may be \emph{encoded but not accessible} under different conditions than how it was learned. One example is the reversal curse \citep{reversal, LinFL0L00WY24}, where an LLM can answer that Oasis played their first gig at the Boardwalk club, yet fails to answer who played their first gig there.

Encoding and recall failures are indistinguishable under accuracy metrics, yet they imply different limitations and solutions. Encoding failures call for pre-training interventions, such as scaling model size or data coverage. Recall failures suggest post-training interventions that often improve how models utilize what they already encode \cite{zhou2023nips, GekhmanYAEFRH24, lin2024iclr}.

We therefore propose shifting the unit of analysis from individual questions to \emph{facts}. We introduce \emph{knowledge profiling}: a framework that categorizes facts into one of five profiles (see Figure~\ref{fig:intro_fig}), based on whether the fact is encoded, and by how accessible it is: cannot be recalled, can be directly recalled, or can be recalled only with inference-time computation (thinking). We use \emph{thinking} to refer to inference-time techniques that elicit intermediate computations before the final answer, including both chain-of-thought (CoT) prompting and dedicated thought tokens generated by thinking-optimized LLMs. We operationalize the definitions of encoding and recall behaviorally \citep{Holtzman25prompting}. We measure \emph{encoding} by priming the LLM to complete a factual proposition using a context that resembles how the fact was likely to be encountered during pre-training. We measure \emph{recall}\footnote{By analogy to human memory, we use \emph{encoding} to denote parametric representation, and \emph{recall} to accessing encoded facts.} by asking questions about these encoded facts across different contexts, including phrasings or relational orderings. This enables the study of frontier LLMs despite the inaccessibility of their weights and training data.

To support knowledge profiling, we introduce \wikiprofile{}, a new benchmark comprising 2,150 facts, each paired with 10 questions that probe encoding, recall, and recognition (multiple-choice). \wikiprofile{} is constructed via a fully automated multi-step pipeline that employs a prompted LLM grounded in web search to generate, refine, and verify questions. Unlike other benchmarks that rely on synthetic or schema-restricted knowledge-base triplets \citep{LuoVPH23, Kumar2024Automatic, Yuan2024Holistic, WuPXZ0MD0LW25}, our facts are extracted from natural documents. This enables evaluation in more realistic settings, where facts are expressed indirectly, and may involve complex, context-dependent relations \citep{WangWH00C0Y25}.

Using \wikiprofile{}, we evaluate 13 LLMs, resulting in more than 4 million graded answers. We find that scaling primarily improves encoding, while recall remains a major bottleneck. For frontier LLMs, including Gemini-3-Pro and GPT-5, encoding is nearly saturated, with 95--98\% of facts encoded. Yet the same models still cannot recall 25--33\% of facts without thinking. In addition, we show that scaling model size alone increases what the model encodes but does not necessarily improve its recall.

Why do LLMs struggle to access knowledge they have encoded? Our findings suggest that recall is tightly coupled to the conditions under which facts were learned, degrading when queries diverge from training-time patterns. We manifest this through two cases: fact popularity and the reversal curse. 
While the encoding gap between popular and long-tail facts for frontier LLMs is small (a few percentage points), the recall gap is substantially larger (often exceeding 25\%). Reverse questions reveal a surprising pattern: LLMs struggle to answer reverse questions compared to direct ones, despite showing no disadvantage in recognizing the correct answer under multiple-choice evaluation.

Together, our results refine prior accounts of long-tail errors and the reversal curse \citep{KandpalDRWR23, reversal, MallenAZDKH23, He2025pop}. Whereas these failures are often interpreted as limitations of memorization or bidirectional encoding, our results suggest a different picture: rare facts are often encoded but inaccessible, and reverse facts can be recognized even when they cannot be generated. This reframes both phenomena as recall failures rather than \emph{`missing knowledge'}. Our results also complement prior work on \emph{`hidden knowledge'} \citep{Gekhman25Inside, OrgadTGRSKB25}: what appears to be knowledge accessible only through model internals may reflect encoded facts whose recall is context-dependent, such that the right prompt or context can elicit \citep{JiangXAN20}.

Given that encoding in frontier LLMs is nearing saturation while substantial headroom remains for recall, future improvements are likely to come from better utilization of existing knowledge. One mechanism that already demonstrates this is thinking: our results show that thinking recovers 40--65\% of encoded-but-not-directly-known facts. The gains are most pronounced in challenging settings, such as long-tail facts and reverse questions. This parallels the ``tip-of-the-tongue'' phenomenon in human memory \citep{Brown1966, Schwartz2002}, in which additional effort can surface knowledge that is present but momentarily inaccessible. Post-training and inference-time methods may thus play a central role in factuality by improving how models utilize what they already encode.

In summary, our contributions: (1) We propose a behavioral framework that characterizes factual knowledge in LLMs by separating encoding from recall via knowledge profiles; (2) We introduce \wikiprofile{}, a benchmark for knowledge profiling, along with a fully automated pipeline for constructing such benchmarks; and (3) Through evaluation of 13 LLMs on Wikipedia-derived facts, we show that the bottleneck for factuality is recall, not encoding, and that thinking helps recover otherwise inaccessible knowledge.

\section{Knowledge Profiling}
\label{sec:profiling}

We focus on \emph{factual knowledge} as the ability to answer short-form questions correctly without external tools. Our aim is to provide a systematic, model-agnostic characterization of factual behavior that applies to both closed- and open-weight LLMs. To this end, we introduce two complementary operational notions, defined purely in terms of observable behavior: \emph{encoding} and \emph{knowledge}. Together, they allow us to determine whether errors arise from encoding or recall, thereby informing different paths to improvement. 

\subsection{Operationalizing Encoding and Knowledge}
\label{sub:measuring}

\begin{figure}[t]
    \centering
    \includegraphics[width=0.48\textwidth]{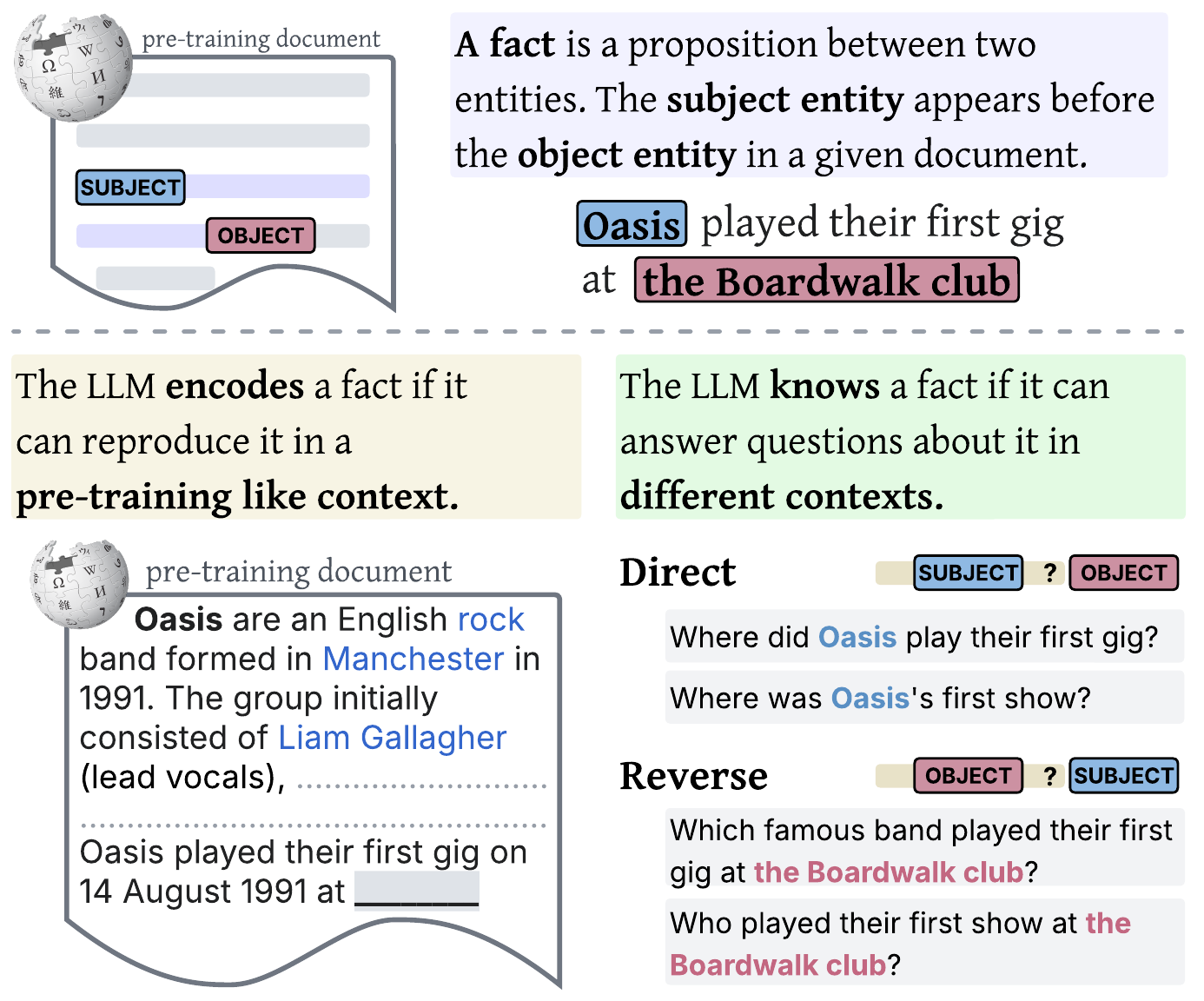}
        \caption{\emph{Top:} We extract facts from Wikipedia, a predominant source of pre-training data. \emph{Left:} We measure encoding by prompting the LLM to reproduce facts within their original context, testing whether they are stored in the model's parameters. \emph{Right:} We measure knowledge by asking questions across varied phrasings and relational directions, with and without thinking.}
    \label{fig:setup}
\vspace{-0.8em}
\end{figure}

\medskip\noindent\textbf{Facts and Questions:}
We define a \emph{fact} as a proposition\footnote{Although facts are often described as subject-relation-object triplets, in our data they may involve multiple or complex relations and are therefore treated as \textit{propositions}.} involving an ordered pair of entities: a subject and an object. The roles of subject and object are determined by the source text from which the fact was extracted (e.g., a Wikipedia document): the subject is the entity that appears first in the text, and the object appears subsequently. A question whose answer is the object is termed a \emph{direct question}, while a question whose answer is the subject is termed a \emph{reverse question}. Examples are shown in Figure~\ref{fig:setup}. 

For each question $q$, we generate $n$ responses from the LLM and use a prompted LLM grader (autorater) to compare each response to the gold answer and label it as correct or incorrect.\footnote{The grader may assign `partially' or `other' labels when a response cannot be reliably evaluated. See \S\ref{sec:experimental}.}
Let $c_q$ and $i_q$ denote the number of responses to question $q$ graded as correct and incorrect. We define the \emph{question grade} as the accuracy over gradable responses:
\[
g(q) = \frac{c_q}{c_q + i_q}.
\]

\medskip\noindent\textbf{Encoding:}
The notion of \emph{encoding} has been defined in multiple ways in prior work, typically referring to information stored in the model's internal parametric representations. Existing approaches to measuring encoding often rely on access to internal states, a requirement that does not align with our focus on evaluating frontier LLMs. Moreover, the extent to which these methods reliably capture whether a model truly encodes a fact remains an open question \citep{HaseBKG23, Ma2024bird, Huang2024Demys, WeiYWMZ0024, ChenC00025, Haller2025brittle}. 

We therefore adopt an \emph{encoding-via-memorization} perspective: an LLM \textbf{encodes} a fact if it can correctly reproduce that fact in a pre-training-like context. Intuitively, successful reproduction under training-like conditions provides sufficient evidence that the fact is encoded (though it may not be known). Since encoding, under this definition, is assessed behaviorally, we operationalize it using tasks that strongly prime the model with the context in which the fact $f$ originally appeared. We denote the set of such encoding tasks by $\mathcal{E}_f$. We say an LLM \emph{encodes} the fact $f$ if:
\[
\exists q \in \mathcal{E}_f, \quad g(q) > \tau,
\]
where $\tau$ is a predefined threshold. We set $\tau = 0.5$ in all experiments as a natural majority threshold. We show in Appendix~\ref{app_sub:thresholds} that our findings are robust to this choice.

In our setup, $\mathcal{E}_f$ consists of two tasks.
The first task is \emph{proposition completion} (see Figure~\ref{fig:setup}), in which the model is given the \emph{left context} (the entire source text truncated immediately before the object entity), and is asked to complete the factual statement. This task directly mimics the pre-training objective for which the LLM was optimized. Notably, the answer (the object entity) does not appear in the left context. 

The second task, which we refer to as \emph{contextual questioning}, uses the same left context but replaces the final sentence with a direct question about the fact. We include both tasks because completion-based evaluation can be ambiguous for post-trained LLMs, which are optimized to answer questions rather than continue pre-training documents. LLMs may continue a sentence in multiple plausible ways that do not explicitly state the target fact. The contextual questioning task mitigates this ambiguity by explicitly eliciting the target fact in question form while preserving the original source context, which primes the model.

We deliberately exclude \emph{thinking} from the encoding evaluation. Correct reproduction can arise not only from direct encoding but also from inference, including multi-hop reasoning or educated guessing based on other encoded facts. To avoid conflating encoding with inference, encoding is always measured without thinking. Knowledge, in contrast, is evaluated both with and without thinking.

\medskip\noindent\textbf{Knowledge:}
We say that an LLM \textbf{knows} a fact if it can correctly answer questions about it across a range of semantically equivalent contexts. This definition captures the intuition that factual knowledge should be robust to superficial changes, such as in phrasing and to alternative relational directions \citep{ElazarKRRHSG21, MengBAB22, Zucchet2025dynamics}. Let $\mathcal{Q}_f$ denote a set of questions associated with a fact $f$. We say an LLM \emph{knows} a fact $f$ if:
\[
\forall q \in \mathcal{Q}_f, \quad g(q) > \tau,
\]
The contrast with encoding is intentional: encoding uses existential quantification ($\exists$) because reproducing a fact in any priming context suffices as evidence of storage, whereas knowledge uses universal quantification ($\forall$) because robust recall should not depend on phrasing or query direction. In our setup, $\mathcal{Q}_f$ consists of four questions: two direct and two reverse, each in a different phrasing. This tests invariance to both surface form and relational direction. 

\subsection{Knowledge Profiles}
\label{sub:profiles}

Given our operational definitions, each fact can be characterized along three dimensions: whether it is \emph{encoded}, whether it is \emph{known without thinking}, and whether it is \emph{known only with thinking}. The interaction of these dimensions yields qualitatively distinct \emph{knowledge profiles}. In particular, we identify five profiles, illustrated in Figure~\ref{fig:intro_fig}. 

\smallskip\noindent\textbf{Encoding Failure:}
The fact is neither encoded nor known, with and without thinking. A high rate of encoding failures indicates limitations in model capacity or pre-training data coverage, suggesting interventions such as increasing model size or expanding the training set.

\smallskip\noindent\textbf{Recall Failure:}
The fact is encoded but not known, even with thinking. This indicates that the fact is stored but inaccessible at inference time, and calls for improvements beyond pre-training, such as in post-training.

\smallskip\noindent\textbf{Direct Recall:}
The fact is encoded and known without thinking. This straightforward recall reflects highly accessible knowledge, which is desirable in many applications (e.g., when integrating facts into long-form responses).

\smallskip\noindent\textbf{Recall with Thinking:}
The fact is encoded but known only with thinking. This indicates that the fact is not readily accessible and requires additional computation to be recalled. This profile can be viewed as a recovery mechanism for otherwise inaccessible facts.

\smallskip\noindent\textbf{Inference without Encoding:}
The fact is not encoded but is known when thinking is enabled. In this case, correctness may arise from inference over other encoded facts, such as multi-hop reasoning or educated guessing. It is also possible that the fact is encoded but missed by our encoding tasks. This profile is less reliable; relying on it may promote hallucinations, and it occurs less frequently in practice.

Finally, we do not define a profile for facts that are not encoded but known without thinking. While such cases are possible, they occur in fewer than 0.5\% of instances across all evaluated models and are likely attributable to sampling noise or grader error. We therefore exclude this profile.
\begin{figure*}[t]
    \centering
    \includegraphics[width=0.97\textwidth]{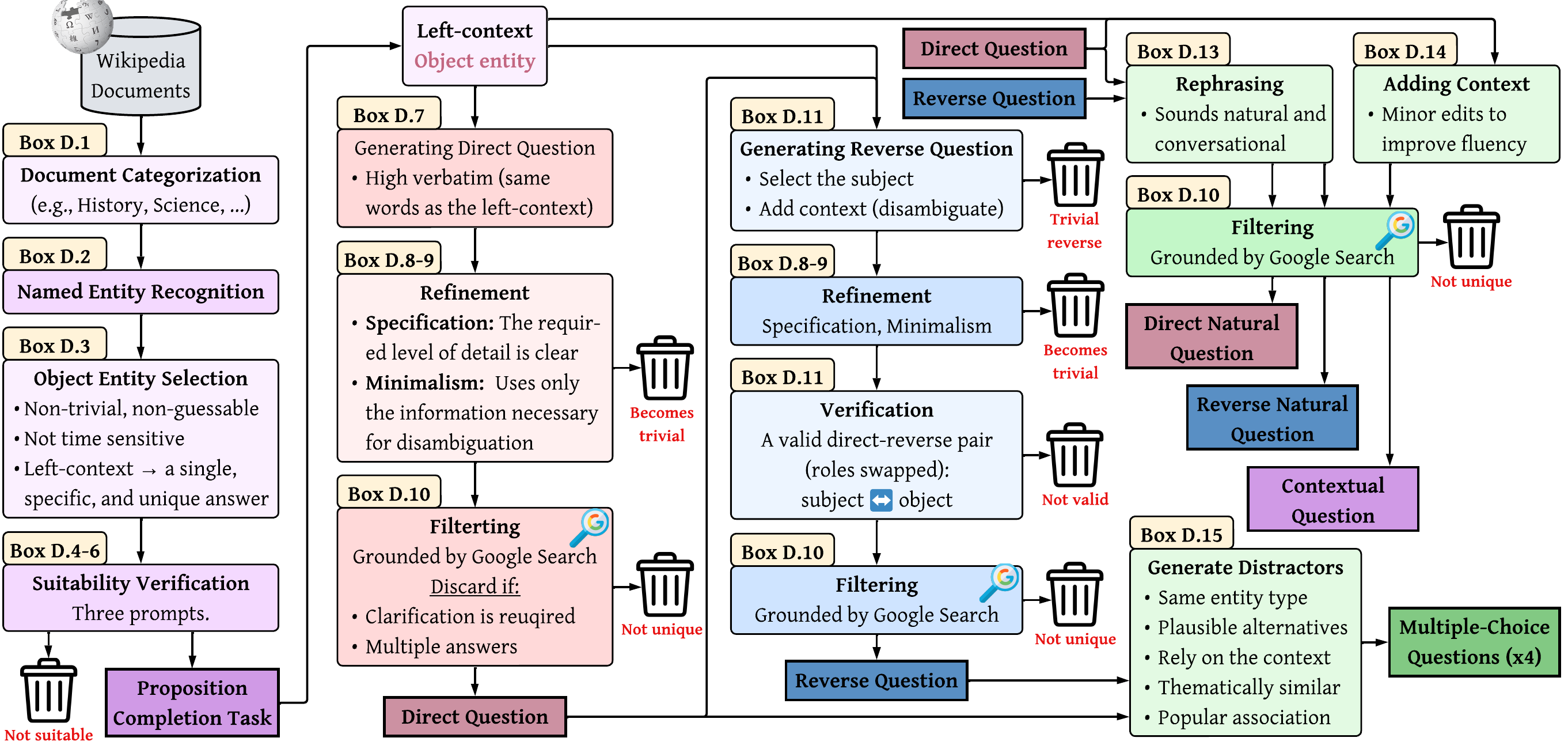}
    \caption{\textbf{The WikiProfile Creation Pipeline:} We propose a fully automated pipeline based on prompted LLMs. The yellow boxes denote the specific prompts used at each step (see \S\ref{app_sec:prompts}). \emph{Left (purple):} Fact extraction and construction of the proposition completion task. \emph{Center (red and blue):} Construction of direct and reverse questions via generation, refinement, and filtering (grounded by Google Search). \emph{Right (green):} Creation of remaining questions (natural phrasing, contextual, and multiple-choice versions) based on the direct/reverse pairs. Facts are discarded throughout the pipeline if their associated questions are rejected. Full details in Appendix~\ref{sec:dataset}. The encoding tasks are `Proposition Completion' and `Contextual Question'. The knowledge tasks are `Direct/Reverse', `Direct/Reverse Natural' questions.}
    \label{fig:pipeline}
\vspace{-0.8em}
\end{figure*}

\section{The \wikiprofile{} Benchmark}
\label{sec:data}

To operationalize the knowledge profiles, we need a benchmark that supports measuring both encoding and knowledge. We hence introduce \wikiprofile{}, a benchmark for evaluating and profiling factual knowledge in LLMs. Since profiling involves measuring encoding, we extract facts from Wikipedia, a trusted reference that pre-training pipelines are likely to prioritize, making it both a strong candidate for encoded knowledge and a reliable ground truth. \wikiprofile{} consists of 2150 facts, each with ten corresponding questions. These include two questions for measuring encoding (proposition-completion and contextual questions); four questions for measuring knowledge (direct and reverse questions, each instantiated with two distinct phrasings); and four multiple-choice variants for measuring fact verification capabilities. 
Examples of tasks are in Table~\ref{tab:examples}.

All questions in \wikiprofile{} have a single, unambiguous gold answer, enabling automatic validation with an LLM grader (see \S\ref{sec:experimental}). \wikiprofile{} is constructed through a fully automated pipeline (Gemini-2.5-Pro with thinking; \citealt{gemini}) that may be applicable to other corpora for domain-specific knowledge profiling.

\medskip\noindent\textbf{Pipeline Overview:}
The full pipeline spans dozens of steps and fifteen prompts; here we provide a high-level overview (Figure~\ref{fig:pipeline}), with complete details and prompts in Appendix~\ref{sec:dataset} and~\ref{app_sec:prompts}. 
All prompts in the pipeline were carefully developed through manual optimization and error analysis on a held-out subset of tens of documents, from which the examples used in the prompts are drawn.

\medskip\noindent\textbf{Fact Extraction:}
To extract facts, we sample 10,000 Wikipedia pages, record their visit counts, and categorize them into nine topics (see Figures~\ref{fig:years} and~\ref{fig:categories} for distributions). Next, we perform NER to identify entities and their types (e.g., DATE, LOCATION, PERSON; see Figure~\ref{fig:entities}). For each document, we select up to 3 candidate object entities (not from the first sentences), each of which defines a distinct fact. These candidates are selected based on strict criteria: an entity is suitable only if it represents a non-trivial, non-guessable, and time-independent (i.e., not subject to change) completion of its left context. Furthermore, the left context must constrain the completion to a single, unambiguous answer. After a final verification step, we downsample to 5,000 candidates, balancing categories and entity types. The median is 100 words, the mean is 116 words ($\pm$56 std).

\medskip\noindent\textbf{Question Generation:}
We generate direct and reverse questions through a three-step process of generation, refinement, and filtering, ensuring that each question is unambiguous, specific, minimal, and has a unique answer. Direct questions are generated first, and reverse questions are then derived by swapping subject and object roles. Because our goal is to test encoding, we initially generate high-verbatim questions that closely match the source text; we then create natural rephrasings of each to test robustness to phrasing variation. Contextual questions are constructed by appending the direct question to its left context, providing maximal priming. For each question, we also generate a multiple-choice variant with three plausible distractors matched by entity type and thematic similarity. All questions undergo filtering grounded in Google Search. We prompt an LLM with each question and discard cases where multiple answers are returned or clarification is needed. This filtering is strict: if any question is rejected, the entire fact is discarded, leaving us with approximately 2,200 facts. 

Finally, we manually validated facts and their associated questions, discarding fewer than 2\% of facts and yielding a final set of 2,150 facts. See Appendix~\ref{app_sub:manual} for details.
\section{Experimental Setup}
\label{sec:experimental}

\noindent\textbf{LLMs:}
We evaluate 13 LLMs from five families, covering commercial and open-weight models: Gemini-3 and Gemini-2.5 (Pro and Flash variants) \citep{gemini}; GPT-5.2, GPT-5, and GPT-5-mini \citep{gpt5}; GPT-4.1 and GPT-4.1-mini \citep{gpt41}; and Gemma3 (1B, 4B, 12B, 27B) \citep{gemma3}. Each model is evaluated with and without thinking. Gemini-2.5, Gemini-3, and GPT-5 are thinking-optimized; we use default settings when thinking is enabled and set effort/budget to zero to disable it. For GPT-4.1 and Gemma3, we use CoT prompting.

\medskip\noindent\textbf{Response Generation:}
For each example (an incomplete text or a question), we generate eight responses per LLM with a temperature of 1. In Appendix~\ref{app_sub:n_sensitivity}, we justify the choice of eight responses. We employ three prompts, one for each task format (completion, closed-book QA, and multiple-choice QA). Each prompt instructs the LLM to produce concise, answer-only responses (see Box~\ref{box:responses}). For multiple-choice questions, answer choices are shuffled each time, ensuring that each option is the correct answer exactly twice. In total, we evaluate $\approx$4.5 million responses (13$\times$2 LLMs$ \times$2150 facts $\times$10 tasks $\times$8 responses).

\medskip\noindent\textbf{Response Evaluation:}
We use two prompted LLM graders (autoraters) \citep{simpleqa}: one for the completion tasks (Box~\ref{box:autorater_comp}) and one for the closed-book QA tasks (Box~\ref{box:autorater_qa}). Both graders use Gemini-2.5-Pro with thinking. The graders assign one of four labels: CORRECT, INCORRECT (factual contradiction), PARTIALLY (correct but at a different granularity; \citealt{YonaAG24}), or OTHER (unattempted or unverifiable answers). Including PARTIALLY improved grader stability in preliminary experiments. 

As defined in \S\ref{sub:measuring}, the question grade is computed using CORRECT and INCORRECT responses only, excluding PARTIALLY and OTHER, since assigning them weights is subjective \citep{Kalai25why}. In practice, such responses are rare (less than 5\%; see Table~\ref{tab:nan_strategy}). When all responses are PARTIALLY or OTHER, the fact is excluded from aggregate statistics. Alternative handling strategies yield negligible differences (see our analysis in Appendix~\ref{app_sub:nans}).

Finally, in Appendix~\ref{app_sub:grader}, we assess cross-family grader consistency by comparing graders based on Gemini-2.5-Pro and GPT-5, and find 98.2\% agreement. The remaining disagreements primarily involve OTHER or PARTIALLY labels, which are excluded from our analysis; thus, our results are not sensitive to the choice of underlying LLM.
\begin{figure}[t!]
    \centering
    \includegraphics[width=0.48\textwidth]{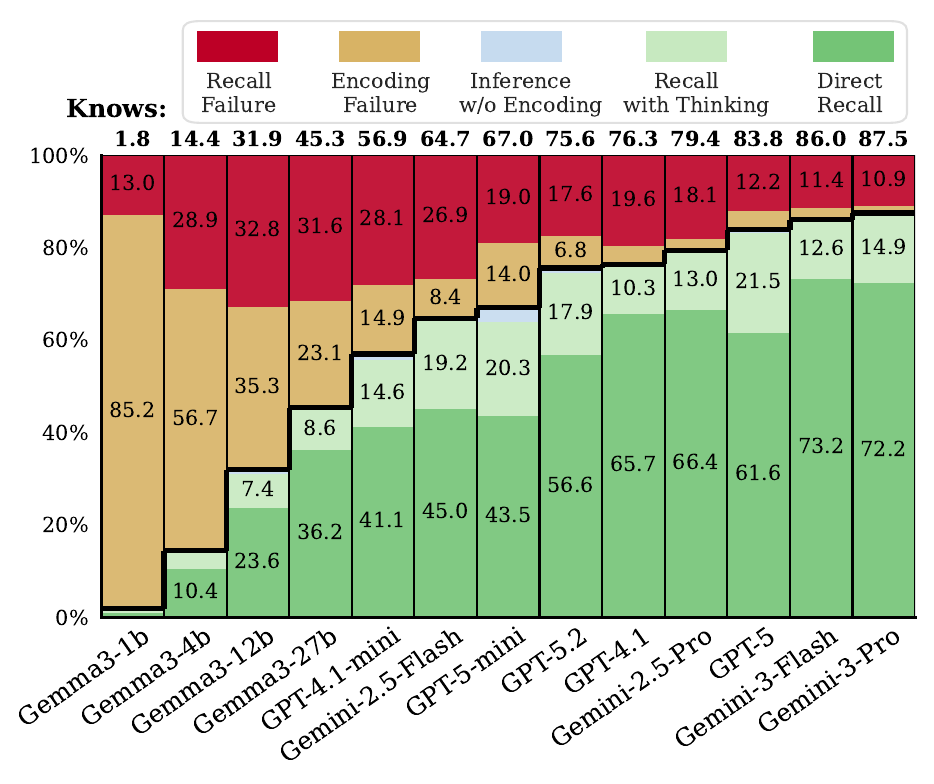}
    \caption{\textbf{Knowledge Profiles:} Distribution of the five profiles across 13 LLMs (percentages). The black line marks \emph{potential knowledge}: the fraction of facts known with or without thinking (`Direct Recall'+`Recall with Thinking'+`Inference without Encoding'). As shown, encoding failures decrease sharply with scale, while recall failures persist even in frontier models.}
  \label{fig:profiles_dist}
\vspace{-0.8em}
\end{figure}

\section{Results}
\label{sec:results}

Our results suggest that factual errors arise not from \emph{missing knowledge} (empty shelves), but from \emph{inaccessible knowledge} (lost keys). We first characterize this recall bottleneck by examining knowledge profiles across all 13 evaluated models (\S\ref{sub:bottleneck}). We then ask why recall fails, identifying two systematic patterns, fact popularity and question directionality, where encoded knowledge remains difficult to access (\S\ref{sub:recall_fails}). Finally, we examine thinking as a mechanism that compensates for these weaknesses, recovering facts that direct generation cannot surface (\S\ref{sub:thinking}). Due to the analysis scale, we occasionally report results for frontier LLMs only; complete results and details are provided in Appendix \ref{app_sec:additional_results}.

\subsection{The Recall Bottleneck}
\label{sub:bottleneck}

We begin by examining knowledge failures, namely, facts that are not known. These correspond to two profiles: \emph{Encoding Failures} and \emph{Recall Failures}. Figure~\ref{fig:profiles_dist} shows the distribution of all five knowledge profiles across 13 LLMs.

\smallskip\noindent\textbf{Encoding is saturated; recall is not.} For frontier LLMs such as Gemini-3-Pro and GPT-5, factual encoding is near saturation, with 95-98\% of facts encoded. Yet these models fail to directly recall 26--34\% of the facts, or 11--12\% even with thinking. Accordingly, recall failures account for more than 70\% of GPT-5.2's errors and a larger share in stronger models, suggesting recall is indeed a bottleneck. 

\smallskip\noindent\textbf{Scaling improves encoding more than recall.} The Gemma3 model family illustrates this clearly. Increasing model capacity and size from 1B to 27B parameters reduces encoding failures from 85\% to 23\%. However, as model scale increases, a growing fraction of errors shifts from encoding failures to recall failures, peaking at 33\% (with thinking) and 40\% (without). In this family, scaling increases what the model stores; it does not necessarily improve what it can access at inference time.

\subsection{Why Recall Fails}
\label{sub:recall_fails}

Having established that recall is the bottleneck, we now ask: what makes it difficult? To answer this, we next focus on encoded facts. We show that recall remains tied to the conditions under which facts were acquired. During pre-training, facts are encountered in specific contexts, phrasings, and orderings. When queries diverge from the training-time patterns, recall becomes harder. We examine two manifestations of this: fact popularity and question directionality.

\begin{figure}[h]
    \centering
    \includegraphics[width=0.48\textwidth]{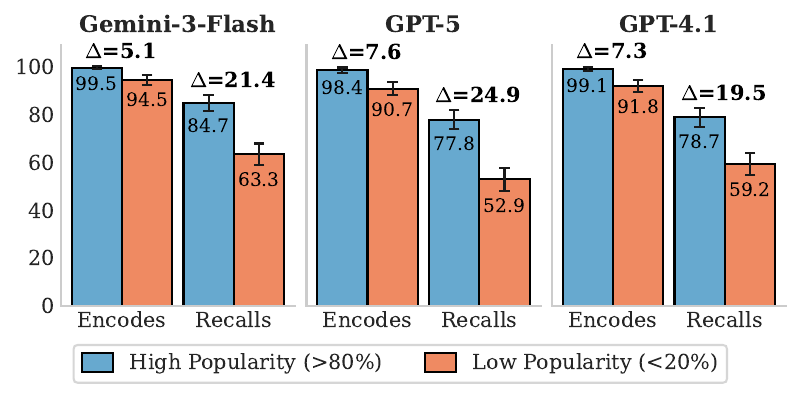}
    \caption{\textbf{Fact Popularity:} We compare two popularity tiers (bottom 20\% vs. top 20\%) in terms of encoding rates and direct recall rates (knowing encoded facts without thinking). The $\Delta$ indicates the gap between tiers. As shown, it is narrow for encoding but wide for recall. Figure~\ref{fig:popularity_appendix} presents all LLMs.}
    \label{fig:popularity}
\vspace{-0.8em}
\end{figure}

\smallskip\noindent\textbf{Rare facts are encoded but hard to recall.} We compare facts from low-popularity Wikipedia pages (bottom 20\% by page views, used as a proxy for rare facts) with those from high-popularity pages (top 20\%). Figure~\ref{fig:popularity} presents, for the two tiers, the percentage of encoded facts and, among encoded facts, the percentage that are directly recalled. Conditioning on encoded facts isolates recall from encoding, allowing us to assess whether rare facts are harder to recall even when they are encoded. First, we find that rare facts are encoded at high rates approaching those of popular facts. Second, while the \emph{encoding gap} between popular and rare facts is modest (e.g., for Gemini-3-Pro, 99.5\% vs. 94.5\%, $\Delta\!=\!5$), the \emph{recall gap} is far larger (84.6\% vs. 63.3\%, $\Delta\!=\!21.3$); this pattern holds for all frontier LLMs. For smaller models like Gemma3-4B, encoding gaps are much larger; this is consistent with our broader finding that scaling primarily improves encoding.

\smallskip\noindent\textbf{Reframing the long-tail problem.} Prior work established that LLMs struggle with long-tail (rare) facts, typically interpreted as a capacity limitation \citep{KandpalDRWR23, MallenAZDKH23, SunXZLD24}. Our results suggest a complementary picture: rare facts \emph{are} encoded; however, recalling them is the challenge. Currently, the bottleneck has shifted from knowledge acquisition to utilization. Consequently, improving factuality in frontier LLMs will likely require interventions that enhance recall, not only scale.

\begin{figure}[h]
    \centering
    \includegraphics[width=0.48\textwidth]{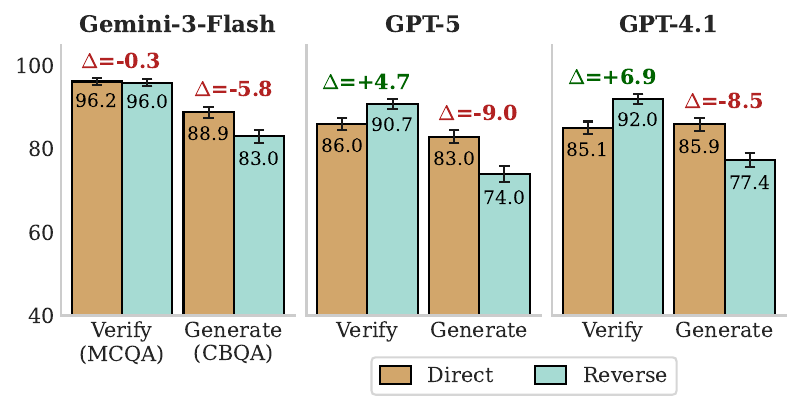}
    \caption{\textbf{The Reversal Curse:} We compare direct recall rates (knowing encoded facts without thinking) on direct and reverse questions across two tasks: verification (multiple-choice) and generation (closed-book). The $\Delta$ denotes the gap between the direct and reverse settings. We find that LLMs handle reverse questions effectively in verification but struggle in generation. Figure~\ref{fig:popularity_appendix} presents results for all LLMs.}
    \label{fig:dir_rev_mcqa}
\vspace{-0.5em}
\end{figure}

\smallskip\noindent\textbf{Reverse questions are verifiable but hard to recall.}
Figure~\ref{fig:dir_rev_mcqa} shows that all LLMs (except for Gemini-2.5-Flash) exhibit lower recall rates for reverse questions (e.g., GPT-5: 82.9\% direct vs. 74\% reverse). This asymmetry is consistent with the ``reversal curse'' documented by \citet{reversal}. Our results suggest a refinement of this view. If LLMs truly lack bidirectional knowledge, they should struggle with reverse questions regardless of format. To test this, we compare closed-book generation with multiple-choice questions, in which the correct answer is presented among distractors (verification). In verification, reverse questions are no harder than direct ones: for 9 models, reverse is actually \emph{easier}; for the remaining 4, performance is similar. This dissociation suggests that LLMs \emph{are aware} of the bidirectional association of the fact; they can recognize the correct answer when presented with it. Their failure lies in recalling it when the direction does not match the training data. In this view, the reversal curse is a recall phenomenon rather than a lack of bidirectional association.

\begin{figure*}[t]
    \centering
    \includegraphics[width=0.48\textwidth]{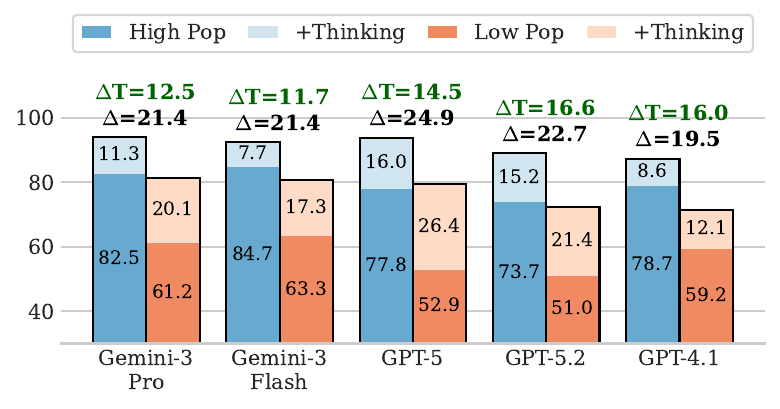}
    \includegraphics[width=0.48\textwidth]{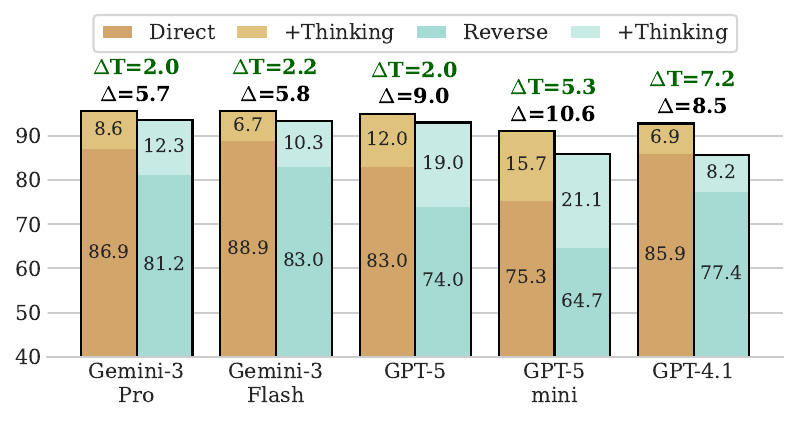}
    \caption{\textbf{Thinking and Recall:} 
    We examine the impact of thinking on recall (knowing encoded facts). On the \emph{left}, we compare two popularity tiers (bottom 20\% vs. top 20\%); On the \emph{right}, we compare direct and reverse questions. The popularity or directional gaps are denoted by $\Delta$ (no thinking) and \textcolor{OliveGreen}{$\Delta T$} (with thinking). As shown, thinking narrows the gaps (\textcolor{OliveGreen}{$\Delta T$}$< \Delta$). Figure~\ref{fig:dir_pop_thinking_appendix} presents all LLMs.}
    \label{fig:dir_pop_thinking}
\vspace{-0.5em}
\end{figure*}

\smallskip\noindent\textbf{Phrasing is not a factor.}
Our definition of \emph{knows} requires correctness across phrasings and directions. We verify that phrasing does not confound our results by conducting hypothesis tests comparing the \emph{distributions} of performance under high-verbatim and natural phrasings. Across 104 hypothesis tests (13 LLMs, with and without thinking, across four task pairs) and after FDR correction, we find no significant effects (see Appendix~\ref{app_sub:phrasing} for details).

\subsection{Thinking as a Recovery Mechanism}
\label{sub:thinking}

We now turn to the question of what enables the recovery of otherwise inaccessible knowledge. To this end, we examine the potential of \emph{thinking} to fill this role. Thinking-optimized LLMs such as Gemini-3, Gemini-2.5, and GPT-5 models allocate additional computation to thinking by default; for non-thinking models (Gemma3 and GPT-4.1), we elicit similar behavior through CoT prompting.

\smallskip\noindent\textbf{Thinking targets recall weaknesses.} Figure~\ref{fig:dir_pop_thinking} presents the improvement from thinking \textbf{on encoded facts} (i.e., recall gains), broken down by popularity and directionality. Thinking yields larger gains precisely where direct recall is weakest. For Gemini-3-Pro, thinking improves the recall of rare facts by 20.1 points, versus 11.3 for popular facts; for reverse questions, GPT-5 gains 19 points, versus 12 for direct questions. Consequently, thinking narrows both the popularity gap (from $\Delta\!=\!21.4$ to $12.5$) and the directionality gap (from $\Delta\!=\!9$ to $2$). 

\smallskip\noindent\textbf{Why does thinking help?}
Thinking gains are remarkable because our facts are single-hop; they require no complex multi-step reasoning. Why should thinking help recall a simple fact such as \emph{``Where did Oasis play their first show?''} We consider three candidate mechanisms (with additional details in Appendix~\ref{app_sub:mechanisms}) through which thinking improves performance:
(i)~\emph{Response diversity}: thinking increases output variance, improving the chance that at least one sample is correct (a sampling effect);
(ii)~\emph{Inference}: thinking enables multi-step derivation from other encoded facts;
(iii)~\emph{Recall facilitation}: thinking helps the model locate and retrieve a specific encoded fact that direct generation fails to access. 

\begin{figure}[b!]
    \centering
    \includegraphics[width=0.48\textwidth]{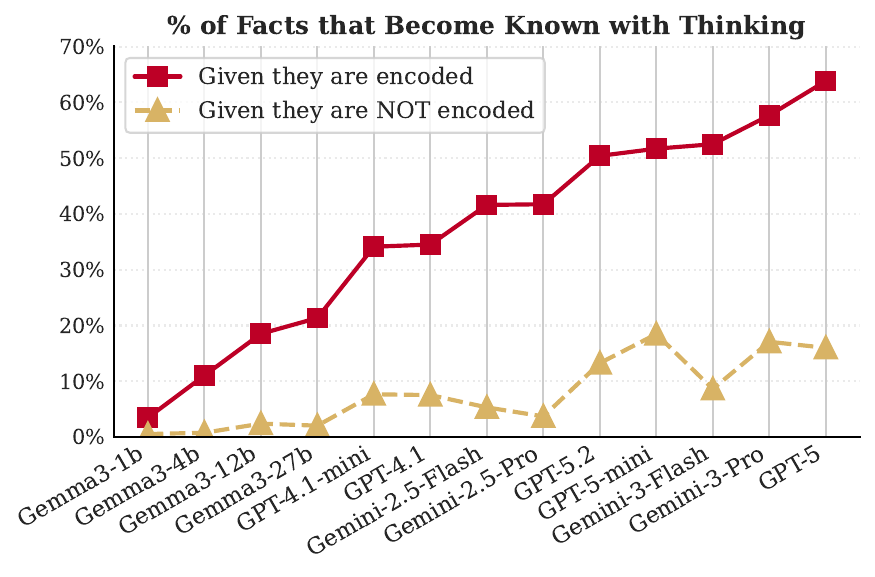}
    \vspace{-1.5em}
    \caption{\textbf{Recovering Depends on Encoding:} We report the percentage of not-known facts that become known with thinking, conditioned on whether the fact is encoded (red) or not (yellow). Thinking recovers 40--65\% of encoded facts in thinking-optimized LLMs, but only 5--15\% of non-encoded facts. This dependence on encoding is consistent with the recall facilitation mechanism.}
    \label{fig:known_given}
\vspace{-0.8em}
\end{figure}

Our analysis in Appendix~\ref{app_sub:mechanisms} rules out response diversity as the main mechanism that improves factuality for thinking-optimized LLMs. When thinking is enabled, models become more consistent in their correctness, producing a higher fraction of correct samples for the same question. This pattern is incompatible with increased output variance. We therefore believe that thinking primarily facilitates recall of encoded knowledge. 
Figure~\ref{fig:known_given} provides evidence that supports this: thinking recovers 40--65\% of encoded but not directly known facts in thinking-optimized LLMs, but only 5--20\% of non-encoded facts. This dependence on encoding status is expected under recall facilitation. It is also partially expected under inference, since non-encoded facts may lack the related encoded premises that inference requires. Nonetheless, our facts are single-hop, and complex deductive reasoning is unnecessary to derive the correct answer. For single-hop facts, recall facilitation is the more parsimonious explanation: the model already stores the answer and need not derive it from other knowledge. We therefore conclude that thinking primarily helps models access knowledge that they have already encoded. \citet{gekhman2026thinking} concurrently show that thinking expands parametric recall through computational buffering and priming.

\smallskip\noindent\textbf{The cost of thinking.}
While thinking improves recall, it is computationally expensive. Across frontier models, 10--20\% of facts are accessible \emph{only} via thinking. 
Determining exactly when to invoke thinking is 
fundamentally a metacognitive problem, as it requires the model to recognize when direct recall is insufficient before triggering more intensive computation \cite{yona2026hallucinations}. Improving direct recall could help side-step this issue until models develop improved metacognitive abilities.

\subsection{Connections to Human Cognition}
\label{sub_app:cognition}

The role of thinking in supporting recall resonates with well-studied phenomena in human memory. The \emph{tip-of-the-tongue} phenomenon describes states in which a person is confident they know something but cannot immediately produce it \citep{Brown1966, Schwartz2002}. Such states are often resolved through deliberate effort: thinking of related concepts or mentally retracing the context in which the information was encountered. The parallel to our findings is suggestive: encoded-but-not-known facts in LLMs may represent a functional analogue, in which the information is stored but the query is insufficient to trigger recall. Thinking provides the additional computation that, like deliberate effort in humans, bridges the gap.

Separately, our verification results (in multiple-choice settings, \S\ref{sub:recall_fails}) echo the \emph{feeling-of-knowing} phenomenon: people often predict they will recognize an answer even when they cannot recall it \citep{Hart1965, Nelson1990}. The dissociation, in which LLMs can verify (recognize) answers to reverse questions but cannot generate (recall) them, mirrors this asymmetry in human memory. While we do not claim that LLMs implement human-like cognitive processes, the functional parallels suggest that the gap between encoding and recall may be a general property of systems in which the conditions for storing information diverge from the conditions for recalling it.

\section{Related Work}
\label{sec:related}


\smallskip\noindent\textbf{Encoding and Memorization:}
Our behavioral definition of encoding connects to work on memorization in LLMs \citep{CarliniIJLTZ23, Huang2024Demys, LuLCD0Q24, MentaAA25, 0003AEAAZW25}. In particular, our encoding criterion aligns with the probabilistic framework of \citet{HayesSCYSNCLC25}. Memorization work typically focuses on privacy or data leakage concerns, whereas we use encoding-via-memorization for knowledge profiling.

\smallskip\noindent\textbf{Latent Knowledge:}
A separate line of work uses internal representations to argue that LLMs possess ``latent knowledge'' or ``hidden knowledge'': facts they encode but fail to express in generation \citep{Kadavath22, BurnsYKS23, GottesmanG24, Gekhman25Inside, OrgadTGRSKB25, Simhi25hack}. 
One possible reason is that the post-training phase may introduce new knowledge that undermines the use of the knowledge acquired during pre-training \citep{GekhmanYAEFRH24,kaplan2026fine}. Our results offer a complementary behavioral perspective: what appears to be latent knowledge may reflect context-dependent recall. Facts surface when queries match training conditions (as in our encoding tasks) but become harder to access when they diverge (as in our knowledge tasks). 

\smallskip\noindent\textbf{The Reversal Curse:}
\citet{reversal} documented that LLMs trained on ``A is B'' often fail to answer ``What is B?''.
Subsequent work has attributed this to the autoregressive objective \citep{Guo0G00Y24, KitouniNWRBI24, LvZ0TCWY24}, training dynamics \citep{ZhuHZJJTR24}, or data asymmetries \citep{ReverseTraining24, He2025pop}, with proposed solutions focusing on pre-training modifications \citep{ReverseTraining24, YuXDZQSZW25} or architectural changes \citep{Wang25jepa}. 
By comparing verification to generation, we show that LLMs can recognize reverse answers they cannot produce, suggesting the curse reflects a \emph{recall asymmetry} rather than a failure to learn bidirectional associations. Thinking can mitigate the curse, providing a path to improvement that prior work has not explored.

\smallskip\noindent\textbf{Thinking and Factuality:}
The effect of CoT prompting \citep{Wei0SBIXCLZ22, WangCWSL024} and reasoning-optimized models \citep{Marjanovic25} on factuality remains contested. Some studies report improved accuracy via increased diversity \citep{0002WSLCNCZ23, Yang25barrel}, while others find higher hallucination rates \citep{WangS0G24, Yao2025prone} or attribute gains to multi-hop reasoning \citep{Zhang25scaling}. Effects on consistency are similarly mixed, with some studies reporting improvements \citep{Kim2025medical, Wu25rlvr} and others reporting degradation \citep{Cheng2025cot, ChengLZW25}. A contemporaneous work by \cite{gekhman2026thinking} also demonstrated that thinking can expand the boundary of parametric knowledge recall, identifying two key mechanisms underlying this effect: a computational buffer and factual priming. Our work offers a unifying perspective: thinking functions as a recall mechanism, not just a reasoning mechanism. Its effect is most pronounced under challenging conditions, such as rare facts and reverse questions.

\section{Discussion}
\label{sec:discussion}

In this work, we propose a framework for knowledge profiling and apply it to profile 13 LLMs across a benchmark of naturally occurring facts from Wikipedia. Our results point to recall, not encoding, as the primary bottleneck for factuality in frontier LLMs. Across the models we evaluated, encoding is approaching saturation; the strongest LLMs encode over 95\% of facts in our benchmark. The challenge lies in accessing what has been learned. Recall failures are not random; they concentrate on rare facts and reverse questions, conditions where queries diverge from training-time contexts. This pattern suggests that LLMs store facts in ways that remain tied to their acquisition, making recall sensitive to surface-level query characteristics.

This shift from encoding to recall has implications beyond short-form factuality. Distinguishing between encoding and recall failures provides a framework that could extend to long-form generation, multi-step reasoning, and other domains where accessing learned knowledge is critical. In the age of retrieval-augmented generation (RAG) and tool-using agents, it is tempting to view parametric knowledge as secondary, something RAG can compensate for. But parametric knowledge is essential for fluency, speed, and integration across contexts. Improving how models access what they already encode is, therefore, a meaningful target.

The gap between encoding and recall might be bridged through intervention during \emph{pre-training}, for example, by augmenting data with self-generated question-answer pairs \cite{Lin2025active}, or during \emph{post-training}, as recent studies suggest that alignment teaches models how to better utilize knowledge acquired during pre-training \cite{zhou2023nips, GekhmanYAEFRH24, lin2024iclr}. Our findings also suggest that \emph{inference-time} techniques, such as thinking, can be effective in aiding parametric recall.

Our study has several limitations worth noting. WikiProfile is scoped to Wikipedia, an encyclopedic domain of relatively prominent facts, and our findings may not generalize to other corpora. Our operational definition of knowledge is likewise bounded: we measure recall along two axes, directionality and phrasing, for single-hop facts, and do not claim to exhaustively characterize all forms in which a fact can be queried. Finally, fully profiling a frontier model on WikiProfile costs $\pm$\$500, though this can be reduced substantially by omitting multiple-choice questions or using fewer samples per question.

\section*{Impact Statement}

This paper presents work aimed at understanding factual errors in LLMs. Our findings could inform deployment decisions by clarifying when models are likely to fail (e.g., with rare facts or reverse questions) and when inference-time computation may be beneficial. We do not foresee negative societal consequences specific to this work.


\bibliography{custom}

@article{Brown1966,
    title={The ``Tip of the Tongue'' Phenomenon},
    author={Brown, Roger and McNeill, David},
    journal={Journal of Verbal Learning and Verbal Behavior},
    year={1966},
    volume={5},
    number={4},
    pages={325--337},
    doi={10.1016/S0022-5371(66)80040-3}
  }

@article{yona2026hallucinations,
  title={Hallucinations Undermine Trust; Metacognition is a Way Forward},
  author={Yona, Gal and Geva, Mor and Matias, Yossi},
  journal={arXiv preprint arXiv:2605.01428},
  year={2026}
}

@article{gekhman2026thinking,
  title={Thinking to Recall: How Reasoning Unlocks Parametric Knowledge in LLMs},
  author={Gekhman, Zorik and Aharoni, Roee and Ofek, Eran and Geva, Mor and Reichart, Roi and Herzig, Jonathan},
  journal={arXiv preprint arXiv:2603.09906},
  year={2026}
}

@article{kaplan2026fine,
  title={Why Fine-Tuning Encourages Hallucinations and How to Fix It},
  author={Kaplan, Guy and Gekhman, Zorik and Zhu, Zhen and Rozner, Lotem and Reif, Yuval and Swayamdipta, Swabha and Hoiem, Derek and Schwartz, Roy},
  journal={arXiv preprint arXiv:2604.15574},
  year={2026}
}

@inproceedings{lin2024iclr,
  author       = {Bill Yuchen Lin and
                  Abhilasha Ravichander and
                  Ximing Lu and
                  Nouha Dziri and
                  Melanie Sclar and
                  Khyathi Raghavi Chandu and
                  Chandra Bhagavatula and
                  Yejin Choi},
  title        = {The Unlocking Spell on Base LLMs: Rethinking Alignment via In-Context
                  Learning},
  booktitle    = {The Twelfth International Conference on Learning Representations,
                  {ICLR} 2024, Vienna, Austria, May 7-11, 2024},
  publisher    = {OpenReview.net},
  year         = {2024},
  url          = {https://openreview.net/forum?id=wxJ0eXwwda}
}

@inproceedings{zhou2023nips,
  author       = {Chunting Zhou and
                  Pengfei Liu and
                  Puxin Xu and
                  Srinivasan Iyer and
                  Jiao Sun and
                  Yuning Mao and
                  Xuezhe Ma and
                  Avia Efrat and
                  Ping Yu and
                  Lili Yu and
                  Susan Zhang and
                  Gargi Ghosh and
                  Mike Lewis and
                  Luke Zettlemoyer and
                  Omer Levy},
  editor       = {Alice Oh and
                  Tristan Naumann and
                  Amir Globerson and
                  Kate Saenko and
                  Moritz Hardt and
                  Sergey Levine},
  title        = {{LIMA:} Less Is More for Alignment},
  booktitle    = {Advances in Neural Information Processing Systems 36: Annual Conference
                  on Neural Information Processing Systems 2023, NeurIPS 2023, New Orleans,
                  LA, USA, December 10 - 16, 2023},
  year         = {2023},
  url          = {http://papers.nips.cc/paper\_files/paper/2023/hash/ac662d74829e4407ce1d126477f4a03a-Abstract-Conference.html}
}

@article{Hart1965,
    title={Memory and the feeling-of-knowing experience},
    author={Hart, Joseph T.},
    journal={Journal of Educational Psychology},
    year={1965},
    volume={56},
    number={4},
    pages={208--216},
    doi={10.1037/h0022263}
  }

@book{Schwartz2002,
    title={Tip-of-the-Tongue States: Phenomenology, Mechanism, and Lexical Retrieval},
    author={Schwartz, Bennett L.},
    year={2002},
    publisher={Lawrence Erlbaum Associates},
    address={Mahwah, NJ}
  }

@inproceedings{Wei0SBIXCLZ22,
author       = {Jason Wei and Xuezhi Wang and Dale Schuurmans and Maarten Bosma and   
Brian Ichter and Fei Xia and Ed H. Chi and Quoc V. Le and Denny Zhou},                  
title        = {Chain-of-Thought Prompting Elicits Reasoning in Large Language    
Models},
booktitle    = {NeurIPS}, 
year         = {2022}
}

@inproceedings{SunXZLD24, author       = {Kai Sun and Yifan Ethan Xu and Hanwen Zha and Yue Liu and Xin Luna Dong}, title        = {Head-to-Tail: How Knowledgeable are Large Language Models (LLMs)? {A.K.A.} Will LLMs Replace Knowledge Graphs?}, booktitle    = {{NAACL}}, year         = {2024} }

@article{Kadavath22, author       = {Saurav Kadavath and Tom Conerly and Amanda Askell and Tom Henighan and Dawn Drain and Ethan Perez and Nicholas Schiefer and Zac Hatfield{-}Dodds and Nova DasSarma and Eli Tran{-}Johnson and Scott Johnston and Sheer El Showk and Andy Jones and Nelson Elhage and Tristan Hume and Anna Chen and Yuntao Bai and Sam Bowman and Stanislav Fort and Deep Ganguli and Danny Hernandez and Josh Jacobson and Jackson Kernion and Shauna Kravec and Liane Lovitt and Kamal Ndousse and Catherine Olsson and Sam Ringer and Dario Amodei and Tom Brown and Jack Clark and Nicholas Joseph and Ben Mann and Sam McCandlish and Chris Olah and Jared Kaplan}, title        = {Language Models (Mostly) Know What They Know}, journal      = {CoRR}, volume       = {abs/2207.05221}, year         = {2022} }

@inproceedings{CarliniIJLTZ23, author       = {Nicholas Carlini and Daphne Ippolito and Matthew Jagielski and Katherine Lee and Florian Tram{\`{e}}r and Chiyuan Zhang}, title        = {Quantifying Memorization Across Neural Language Models}, booktitle    = {{ICLR}}, year         = {2023} }

@inproceedings{BurnsYKS23, author       = {Collin Burns and Haotian Ye and Dan Klein and Jacob Steinhardt}, title        = {Discovering Latent Knowledge in Language Models Without Supervision}, booktitle    = {{ICLR}}, year         = {2023} }

@incollection{Nelson1990,
    title={Metamemory: A Theoretical Framework and New Findings},
    author={Nelson, Thomas O. and Narens, Louis},
    booktitle={The Psychology of Learning and Motivation},
    editor={Bower, Gordon H.},
    volume={26},
    pages={125--173},
    year={1990},
    publisher={Academic Press},
    address={New York}
  }

@inproceedings{MentaAA25,
  author       = {Tarun Ram Menta and
                  Susmit Agrawal and
                  Chirag Agarwal},
  editor       = {Luis Chiruzzo and
                  Alan Ritter and
                  Lu Wang},
  title        = {Analyzing Memorization in Large Language Models through the Lens of
                  Model Attribution},
  booktitle    = {Proceedings of the 2025 Conference of the Nations of the Americas
                  Chapter of the Association for Computational Linguistics: Human Language
                  Technologies, {NAACL} 2025 - Volume 1: Long Papers, Albuquerque, New
                  Mexico, USA, April 29 - May 4, 2025},
  pages        = {10661--10689},
  publisher    = {Association for Computational Linguistics},
  year         = {2025},
  url          = {https://doi.org/10.18653/v1/2025.naacl-long.535},
  doi          = {10.18653/V1/2025.NAACL-LONG.535},
  bibsource    = {dblp computer science bibliography, https://dblp.org}
}

@inproceedings{HayesSCYSNCLC25,
  author       = {Jamie Hayes and
                  Marika Swanberg and
                  Harsh Chaudhari and
                  Itay Yona and
                  Ilia Shumailov and
                  Milad Nasr and
                  Christopher A. Choquette{-}Choo and
                  Katherine Lee and
                  A. Feder Cooper},
  editor       = {Luis Chiruzzo and
                  Alan Ritter and
                  Lu Wang},
  title        = {Measuring memorization in language models via probabilistic extraction},
  booktitle    = {Proceedings of the 2025 Conference of the Nations of the Americas
                  Chapter of the Association for Computational Linguistics: Human Language
                  Technologies, {NAACL} 2025 - Volume 1: Long Papers, Albuquerque, New
                  Mexico, USA, April 29 - May 4, 2025},
  pages        = {9266--9291},
  publisher    = {Association for Computational Linguistics},
  year         = {2025},
  url          = {https://doi.org/10.18653/v1/2025.naacl-long.469},
  doi          = {10.18653/V1/2025.NAACL-LONG.469},
  bibsource    = {dblp computer science bibliography, https://dblp.org}
}

@inproceedings{OrgadTGRSKB25,
  author       = {Hadas Orgad and
                  Michael Toker and
                  Zorik Gekhman and
                  Roi Reichart and
                  Idan Szpektor and
                  Hadas Kotek and
                  Yonatan Belinkov},
  title        = {LLMs Know More Than They Show: On the Intrinsic Representation of
                  {LLM} Hallucinations},
  booktitle    = {The Thirteenth International Conference on Learning Representations,
                  {ICLR} 2025, Singapore, April 24-28, 2025},
  publisher    = {OpenReview.net},
  year         = {2025},
  url          = {https://openreview.net/forum?id=KRnsX5Em3W},
  bibsource    = {dblp computer science bibliography, https://dblp.org}
}

@inproceedings{WangWH00C0Y25,
  author       = {Yiqun Wang and
                  Chaoqun Wan and
                  Sile Hu and
                  Yonggang Zhang and
                  Xiang Tian and
                  Yaowu Chen and
                  Xu Shen and
                  Jieping Ye},
  editor       = {Wanxiang Che and
                  Joyce Nabende and
                  Ekaterina Shutova and
                  Mohammad Taher Pilehvar},
  title        = {Tracing and Dissecting How LLMs Recall Factual Knowledge for Real
                  World Questions},
  booktitle    = {Proceedings of the 63rd Annual Meeting of the Association for Computational
                  Linguistics (Volume 1: Long Papers), {ACL} 2025, Vienna, Austria,
                  July 27 - August 1, 2025},
  pages        = {23246--23271},
  publisher    = {Association for Computational Linguistics},
  year         = {2025},
  url          = {https://aclanthology.org/2025.acl-long.1133/},
  bibsource    = {dblp computer science bibliography, https://dblp.org}
}

@article{Zucchet2025dynamics,
  author       = {Nicolas Zucchet and
                  J{\"{o}}rg Bornschein and
                  Stephanie C. Y. Chan and
                  Andrew K. Lampinen and
                  Razvan Pascanu and
                  Soham De},
  title        = {How do language models learn facts? Dynamics, curricula and hallucinations},
  journal      = {CoRR},
  volume       = {abs/2503.21676},
  year         = {2025},
  url          = {https://doi.org/10.48550/arXiv.2503.21676},
  doi          = {10.48550/ARXIV.2503.21676},
  eprinttype    = {arXiv},
  eprint       = {2503.21676},
  bibsource    = {dblp computer science bibliography, https://dblp.org}
}

@inproceedings{0003AEAAZW25,
  author       = {Xinyi Wang and
                  Antonis Antoniades and
                  Yanai Elazar and
                  Alfonso Amayuelas and
                  Alon Albalak and
                  Kexun Zhang and
                  William Yang Wang},
  title        = {Generalization v.s. Memorization: Tracing Language Models' Capabilities
                  Back to Pretraining Data},
  booktitle    = {The Thirteenth International Conference on Learning Representations,
                  {ICLR} 2025, Singapore, April 24-28, 2025},
  publisher    = {OpenReview.net},
  year         = {2025},
  url          = {https://openreview.net/forum?id=IQxBDLmVpT},
  bibsource    = {dblp computer science bibliography, https://dblp.org}
}

@inproceedings{LuLCD0Q24,
  author       = {Xingyu Lu and
                  Xiaonan Li and
                  Qinyuan Cheng and
                  Kai Ding and
                  Xuanjing Huang and
                  Xipeng Qiu},
  editor       = {Yaser Al{-}Onaizan and
                  Mohit Bansal and
                  Yun{-}Nung Chen},
  title        = {Scaling Laws for Fact Memorization of Large Language Models},
  booktitle    = {Findings of the Association for Computational Linguistics: {EMNLP}
                  2024, Miami, Florida, USA, November 12-16, 2024},
  pages        = {11263--11282},
  publisher    = {Association for Computational Linguistics},
  year         = {2024},
  url          = {https://doi.org/10.18653/v1/2024.findings-emnlp.658},
  doi          = {10.18653/V1/2024.FINDINGS-EMNLP.658},
  bibsource    = {dblp computer science bibliography, https://dblp.org}
}

@inproceedings{ChenC00025,
  author       = {Yuheng Chen and
                  Pengfei Cao and
                  Yubo Chen and
                  Kang Liu and
                  Jun Zhao},
  title        = {Knowledge Localization: Mission Not Accomplished? Enter Query Localization!},
  booktitle    = {The Thirteenth International Conference on Learning Representations,
                  {ICLR} 2025, Singapore, April 24-28, 2025},
  publisher    = {OpenReview.net},
  year         = {2025},
  url          = {https://openreview.net/forum?id=tfyHbvFZ0K},
  bibsource    = {dblp computer science bibliography, https://dblp.org}
}

@inproceedings{Huang2024Demys,
  author       = {Jing Huang and
                  Diyi Yang and
                  Christopher Potts},
  editor       = {Yaser Al{-}Onaizan and
                  Mohit Bansal and
                  Yun{-}Nung Chen},
  title        = {Demystifying Verbatim Memorization in Large Language Models},
  booktitle    = {Proceedings of the 2024 Conference on Empirical Methods in Natural
                  Language Processing, {EMNLP} 2024, Miami, FL, USA, November 12-16,
                  2024},
  pages        = {10711--10732},
  publisher    = {Association for Computational Linguistics},
  year         = {2024},
  url          = {https://doi.org/10.18653/v1/2024.emnlp-main.598},
  doi          = {10.18653/V1/2024.EMNLP-MAIN.598},
  bibsource    = {dblp computer science bibliography, https://dblp.org}
}

@inproceedings{HaseBKG23,
  author       = {Peter Hase and
                  Mohit Bansal and
                  Been Kim and
                  Asma Ghandeharioun},
  editor       = {Alice Oh and
                  Tristan Naumann and
                  Amir Globerson and
                  Kate Saenko and
                  Moritz Hardt and
                  Sergey Levine},
  title        = {Does Localization Inform Editing? Surprising Differences in Causality-Based
                  Localization vs. Knowledge Editing in Language Models},
  booktitle    = {Advances in Neural Information Processing Systems 36: Annual Conference
                  on Neural Information Processing Systems 2023, NeurIPS 2023, New Orleans,
                  LA, USA, December 10 - 16, 2023},
  year         = {2023},
  url          = {http://papers.nips.cc/paper\_files/paper/2023/hash/3927bbdcf0e8d1fa8aa23c26f358a281-Abstract-Conference.html},
  bibsource    = {dblp computer science bibliography, https://dblp.org}
}

@article{Haller2025brittle,
  author       = {Patrick Haller and
                  Mark Ibrahim and
                  Polina Kirichenko and
                  Levent Sagun and
                  Samuel J. Bell},
  title        = {{LLM} Knowledge is Brittle: Truthfulness Representations Rely on Superficial
                  Resemblance},
  journal      = {CoRR},
  volume       = {abs/2510.11905},
  year         = {2025},
  url          = {https://doi.org/10.48550/arXiv.2510.11905},
  doi          = {10.48550/ARXIV.2510.11905},
  eprinttype    = {arXiv},
  eprint       = {2510.11905},
  bibsource    = {dblp computer science bibliography, https://dblp.org}
}

@article{Kalai25why,
  author       = {Adam Tauman Kalai and
                  Ofir Nachum and
                  Santosh S. Vempala and
                  Edwin Zhang},
  title        = {Why Language Models Hallucinate},
  journal      = {CoRR},
  volume       = {abs/2509.04664},
  year         = {2025},
  url          = {https://doi.org/10.48550/arXiv.2509.04664},
  doi          = {10.48550/ARXIV.2509.04664},
  eprinttype    = {arXiv},
  eprint       = {2509.04664},
  bibsource    = {dblp computer science bibliography, https://dblp.org}
}

@inproceedings{YonaAG24,
  author       = {Gal Yona and
                  Roee Aharoni and
                  Mor Geva},
  editor       = {Lun{-}Wei Ku and
                  Andre Martins and
                  Vivek Srikumar},
  title        = {Narrowing the Knowledge Evaluation Gap: Open-Domain Question Answering
                  with Multi-Granularity Answers},
  booktitle    = {Proceedings of the 62nd Annual Meeting of the Association for Computational
                  Linguistics (Volume 1: Long Papers), {ACL} 2024, Bangkok, Thailand,
                  August 11-16, 2024},
  pages        = {6737--6751},
  publisher    = {Association for Computational Linguistics},
  year         = {2024},
  url          = {https://doi.org/10.18653/v1/2024.acl-long.365},
  doi          = {10.18653/V1/2024.ACL-LONG.365},
  bibsource    = {dblp computer science bibliography, https://dblp.org}
}

@inproceedings{PetroniRRLBWM19,
  author       = {Fabio Petroni and
                  Tim Rockt{\"{a}}schel and
                  Sebastian Riedel and
                  Patrick Lewis and
                  Anton Bakhtin and
                  Yuxiang Wu and
                  Alexander H. Miller},
  editor       = {Kentaro Inui and
                  Jing Jiang and
                  Vincent Ng and
                  Xiaojun Wan},
  title        = {Language Models as Knowledge Bases?},
  booktitle    = {Proceedings of the 2019 Conference on Empirical Methods in Natural
                  Language Processing and the 9th International Joint Conference on
                  Natural Language Processing, {EMNLP-IJCNLP} 2019, Hong Kong, China,
                  November 3-7, 2019},
  pages        = {2463--2473},
  publisher    = {Association for Computational Linguistics},
  year         = {2019},
  url          = {https://doi.org/10.18653/v1/D19-1250},
  doi          = {10.18653/V1/D19-1250},
  bibsource    = {dblp computer science bibliography, https://dblp.org}
}

@article{gemini,
  author       = {Gheorghe Comanici and
                  Eric Bieber and
                  Mike Schaekermann and
                  Ice Pasupat and
                  Noveen Sachdeva and
                  Inderjit S. Dhillon and
                  Marcel Blistein and
                  Ori Ram and
                  Dan Zhang and
                  Evan Rosen and
                  Luke Marris and
                  Sam Petulla and
                  Colin Gaffney and
                  Asaf Aharoni and
                  Nathan Lintz and
                  Tiago Cardal Pais and
                  Henrik Jacobsson and
                  Idan Szpektor and
                  Nan{-}Jiang Jiang and
                  Krishna Haridasan and
                  Ahmed Omran and
                  Nikunj Saunshi and
                  Dara Bahri and
                  Gaurav Mishra and
                  Eric Chu and
                  Toby Boyd and
                  Brad Hekman and
                  Aaron Parisi and
                  Chaoyi Zhang and
                  Kornraphop Kawintiranon and
                  Tania Bedrax{-}Weiss and
                  Oliver Wang and
                  Ya Xu and
                  Ollie Purkiss and
                  Uri Mendlovic and
                  Ila{\"{\i}} Deutel and
                  Nam Nguyen and
                  Adam Langley and
                  Flip Korn and
                  Lucia Rossazza and
                  Alexandre Ram{\'{e}} and
                  Sagar Waghmare and
                  Helen Miller and
                  Nathan Byrd and
                  Ashrith Sheshan and
                  Raia Hadsell Sangnie Bhardwaj and
                  Pawel Janus and
                  Tero Rissa and
                  Dan Horgan and
                  Sharon Silver and
                  Ayzaan Wahid and
                  Sergey Brin and
                  Yves Raimond and
                  Klemen Kloboves and
                  Cindy Wang and
                  Nitesh Bharadwaj Gundavarapu and
                  Ilia Shumailov and
                  Bo Wang and
                  Mantas Pajarskas and
                  Joe Heyward and
                  Martin Nikoltchev and
                  Maciej Kula and
                  Hao Zhou and
                  Zachary Garrett and
                  Sushant Kafle and
                  Sercan Arik and
                  Ankita Goel and
                  Mingyao Yang and
                  Jiho Park and
                  Koji Kojima and
                  Parsa Mahmoudieh and
                  Koray Kavukcuoglu and
                  Grace Chen and
                  Doug Fritz and
                  Anton Bulyenov and
                  Sudeshna Roy and
                  Dimitris Paparas and
                  Hadar Shemtov and
                  Bo{-}Juen Chen and
                  Robin Strudel and
                  David Reitter and
                  Aurko Roy and
                  Andrey Vlasov and
                  Changwan Ryu and
                  Chas Leichner and
                  Haichuan Yang and
                  Zelda Mariet and
                  Denis Vnukov and
                  Tim Sohn and
                  Amy Stuart and
                  Wei Liang and
                  Minmin Chen and
                  Praynaa Rawlani and
                  Christy Koh and
                  JD Co{-}Reyes and
                  Guangda Lai and
                  Praseem Banzal and
                  Dimitrios Vytiniotis and
                  Jieru Mei and
                  Mu Cai},
  title        = {Gemini 2.5: Pushing the Frontier with Advanced Reasoning, Multimodality,
                  Long Context, and Next Generation Agentic Capabilities},
  journal      = {CoRR},
  volume       = {abs/2507.06261},
  year         = {2025},
  url          = {https://doi.org/10.48550/arXiv.2507.06261},
  doi          = {10.48550/ARXIV.2507.06261},
  eprinttype    = {arXiv},
  eprint       = {2507.06261},
  bibsource    = {dblp computer science bibliography, https://dblp.org}
}

@article{gemma3,
  author       = {Aishwarya Kamath and
                  Johan Ferret and
                  Shreya Pathak and
                  Nino Vieillard and
                  Ramona Merhej and
                  Sarah Perrin and
                  Tatiana Matejovicova and
                  Alexandre Ram{\'{e}} and
                  Morgane Rivi{\`{e}}re and
                  Louis Rouillard and
                  Thomas Mesnard and
                  Geoffrey Cideron and
                  Jean{-}Bastien Grill and
                  Sabela Ramos and
                  Edouard Yvinec and
                  Michelle Casbon and
                  Etienne Pot and
                  Ivo Penchev and
                  Ga{\"{e}}l Liu and
                  Francesco Visin and
                  Kathleen Kenealy and
                  Lucas Beyer and
                  Xiaohai Zhai and
                  Anton Tsitsulin and
                  R{\'{o}}bert Busa{-}Fekete and
                  Alex Feng and
                  Noveen Sachdeva and
                  Benjamin Coleman and
                  Yi Gao and
                  Basil Mustafa and
                  Iain Barr and
                  Emilio Parisotto and
                  David Tian and
                  Matan Eyal and
                  Colin Cherry and
                  Jan{-}Thorsten Peter and
                  Danila Sinopalnikov and
                  Surya Bhupatiraju and
                  Rishabh Agarwal and
                  Mehran Kazemi and
                  Dan Malkin and
                  Ravin Kumar and
                  David Vilar and
                  Idan Brusilovsky and
                  Jiaming Luo and
                  Andreas Steiner and
                  Abe Friesen and
                  Abhanshu Sharma and
                  Abheesht Sharma and
                  Adi Mayrav Gilady and
                  Adrian Goedeckemeyer and
                  Alaa Saade and
                  Alexander Kolesnikov and
                  Alexei Bendebury and
                  Alvin Abdagic and
                  Amit Vadi and
                  Andr{\'{a}}s Gy{\"{o}}rgy and
                  Andr{\'{e}} Susano Pinto and
                  Anil Das and
                  Ankur Bapna and
                  Antoine Miech and
                  Antoine Yang and
                  Antonia Paterson and
                  Ashish Shenoy and
                  Ayan Chakrabarti and
                  Bilal Piot and
                  Bo Wu and
                  Bobak Shahriari and
                  Bryce Petrini and
                  Charlie Chen and
                  Charline Le Lan and
                  Christopher A. Choquette{-}Choo and
                  CJ Carey and
                  Cormac Brick and
                  Daniel Deutsch and
                  Danielle Eisenbud and
                  Dee Cattle and
                  Derek Cheng and
                  Dimitris Paparas and
                  Divyashree Shivakumar Sreepathihalli and
                  Doug Reid and
                  Dustin Tran and
                  Dustin Zelle and
                  Eric Noland and
                  Erwin Huizenga and
                  Eugene Kharitonov and
                  Frederick Liu and
                  Gagik Amirkhanyan and
                  Glenn Cameron and
                  Hadi Hashemi and
                  Hanna Klimczak{-}Plucinska and
                  Harman Singh and
                  Harsh Mehta and
                  Harshal Tushar Lehri and
                  Hussein Hazimeh and
                  Ian Ballantyne and
                  Idan Szpektor and
                  Ivan Nardini and
                  Jean Pouget{-}Abadie and
                  Jetha Chan and
                  Joe Stanton and
                  John Wieting and
                  Jonathan Lai and
                  Jordi Orbay and
                  Joseph Fernandez and
                  Josh Newlan and
                  Ju{-}yeong Ji and
                  Jyotinder Singh and
                  Kat Black and
                  Kathy Yu and
                  Kevin Hui and
                  Kiran Vodrahalli and
                  Klaus Greff and
                  Linhai Qiu and
                  Marcella Valentine and
                  Marina Coelho and
                  Marvin Ritter and
                  Matt Hoffman and
                  Matthew Watson and
                  Mayank Chaturvedi and
                  Michael Moynihan and
                  Min Ma and
                  Nabila Babar and
                  Natasha Noy and
                  Nathan Byrd and
                  Nick Roy and
                  Nikola Momchev and
                  Nilay Chauhan and
                  Oskar Bunyan and
                  Pankil Botarda and
                  Paul Caron and
                  Paul Kishan Rubenstein and
                  Phil Culliton and
                  Philipp Schmid and
                  Pier Giuseppe Sessa and
                  Pingmei Xu and
                  Piotr Stanczyk and
                  Pouya Tafti and
                  Rakesh Shivanna and
                  Renjie Wu and
                  Renke Pan and
                  Reza Rokni and
                  Rob Willoughby and
                  Rohith Vallu and
                  Ryan Mullins and
                  Sammy Jerome and
                  Sara Smoot and
                  Sertan Girgin and
                  Shariq Iqbal and
                  Shashir Reddy and
                  Shruti Sheth and
                  Siim P{\~{o}}der and
                  Sijal Bhatnagar and
                  Sindhu Raghuram Panyam and
                  Sivan Eiger and
                  Susan Zhang and
                  Tianqi Liu and
                  Trevor Yacovone and
                  Tyler Liechty and
                  Uday Kalra and
                  Utku Evci and
                  Vedant Misra and
                  Vincent Roseberry and
                  Vlad Feinberg and
                  Vlad Kolesnikov and
                  Woohyun Han and
                  Woosuk Kwon and
                  Xi Chen and
                  Yinlam Chow and
                  Yuvein Zhu and
                  Zichuan Wei and
                  Zoltan Egyed and
                  Victor Cotruta and
                  Minh Giang and
                  Phoebe Kirk and
                  Anand Rao and
                  Jessica Lo and
                  Erica Moreira and
                  Luiz Gustavo Martins and
                  Omar Sanseviero and
                  Lucas Gonzalez and
                  Zach Gleicher and
                  Tris Warkentin and
                  Vahab Mirrokni and
                  Evan Senter and
                  Eli Collins and
                  Joelle K. Barral and
                  Zoubin Ghahramani and
                  Raia Hadsell and
                  Yossi Matias and
                  D. Sculley and
                  Slav Petrov and
                  Noah Fiedel and
                  Noam Shazeer and
                  Oriol Vinyals and
                  Jeff Dean and
                  Demis Hassabis and
                  Koray Kavukcuoglu and
                  Cl{\'{e}}ment Farabet and
                  Elena Buchatskaya and
                  Jean{-}Baptiste Alayrac and
                  Rohan Anil and
                  Dmitry (Dima) Lepikhin and
                  Sebastian Borgeaud and
                  Olivier Bachem and
                  Armand Joulin and
                  Alek Andreev and
                  Cassidy Hardin and
                  Robert Dadashi and
                  L{\'{e}}onard Hussenot},
  title        = {Gemma 3 Technical Report},
  journal      = {CoRR},
  volume       = {abs/2503.19786},
  year         = {2025},
  url          = {https://doi.org/10.48550/arXiv.2503.19786},
  doi          = {10.48550/ARXIV.2503.19786},
  eprinttype    = {arXiv},
  eprint       = {2503.19786},
  bibsource    = {dblp computer science bibliography, https://dblp.org}
}

@article{simpleqa_ver,
  author       = {Lukas Haas and
                  Gal Yona and
                  Giovanni D'Antonio and
                  Sasha Goldshtein and
                  Dipanjan Das},
  title        = {SimpleQA Verified: {A} Reliable Factuality Benchmark to Measure Parametric
                  Knowledge},
  journal      = {CoRR},
  volume       = {abs/2509.07968},
  year         = {2025},
  url          = {https://doi.org/10.48550/arXiv.2509.07968},
  doi          = {10.48550/ARXIV.2509.07968},
  eprinttype    = {arXiv},
  eprint       = {2509.07968},
  bibsource    = {dblp computer science bibliography, https://dblp.org}
}

@article{simpleqa,
  author       = {Jason Wei and
                  Nguyen Karina and
                  Hyung Won Chung and
                  Yunxin Joy Jiao and
                  Spencer Papay and
                  Amelia Glaese and
                  John Schulman and
                  William Fedus},
  title        = {Measuring short-form factuality in large language models},
  journal      = {CoRR},
  volume       = {abs/2411.04368},
  year         = {2024},
  url          = {https://doi.org/10.48550/arXiv.2411.04368},
  doi          = {10.48550/ARXIV.2411.04368},
  eprinttype    = {arXiv},
  eprint       = {2411.04368},
  bibsource    = {dblp computer science bibliography, https://dblp.org}
}

@article{Yuan2024Holistic,
  author       = {Jiaqing Yuan and
                  Lin Pan and
                  Chung{-}Wei Hang and
                  Jiang Guo and
                  Jiarong Jiang and
                  Bonan Min and
                  Patrick Ng and
                  Zhiguo Wang},
  title        = {Towards a Holistic Evaluation of LLMs on Factual Knowledge Recall},
  journal      = {CoRR},
  volume       = {abs/2404.16164},
  year         = {2024},
  url          = {https://doi.org/10.48550/arXiv.2404.16164},
  doi          = {10.48550/ARXIV.2404.16164},
  eprinttype    = {arXiv},
  eprint       = {2404.16164},
  bibsource    = {dblp computer science bibliography, https://dblp.org}
}

@inproceedings{WangCWSL024,
  author       = {Yifei Wang and
                  Yuheng Chen and
                  Wanting Wen and
                  Yu Sheng and
                  Linjing Li and
                  Daniel Zeng},
  editor       = {Yaser Al{-}Onaizan and
                  Mohit Bansal and
                  Yun{-}Nung Chen},
  title        = {Unveiling Factual Recall Behaviors of Large Language Models through
                  Knowledge Neurons},
  booktitle    = {Proceedings of the 2024 Conference on Empirical Methods in Natural
                  Language Processing, {EMNLP} 2024, Miami, FL, USA, November 12-16,
                  2024},
  pages        = {7388--7402},
  publisher    = {Association for Computational Linguistics},
  year         = {2024},
  url          = {https://doi.org/10.18653/v1/2024.emnlp-main.420},
  doi          = {10.18653/V1/2024.EMNLP-MAIN.420},
  bibsource    = {dblp computer science bibliography, https://dblp.org}
}

@inproceedings{MallenAZDKH23,
  author       = {Alex Mallen and
                  Akari Asai and
                  Victor Zhong and
                  Rajarshi Das and
                  Daniel Khashabi and
                  Hannaneh Hajishirzi},
  editor       = {Anna Rogers and
                  Jordan L. Boyd{-}Graber and
                  Naoaki Okazaki},
  title        = {When Not to Trust Language Models: Investigating Effectiveness of
                  Parametric and Non-Parametric Memories},
  booktitle    = {Proceedings of the 61st Annual Meeting of the Association for Computational
                  Linguistics (Volume 1: Long Papers), {ACL} 2023, Toronto, Canada,
                  July 9-14, 2023},
  pages        = {9802--9822},
  publisher    = {Association for Computational Linguistics},
  year         = {2023},
  url          = {https://doi.org/10.18653/v1/2023.acl-long.546},
  doi          = {10.18653/V1/2023.ACL-LONG.546},
  bibsource    = {dblp computer science bibliography, https://dblp.org}
}

@inproceedings{LuoVPH23,
  author       = {Linhao Luo and
                  Thuy{-}Trang Vu and
                  Dinh Q. Phung and
                  Gholamreza Haffari},
  editor       = {Houda Bouamor and
                  Juan Pino and
                  Kalika Bali},
  title        = {Systematic Assessment of Factual Knowledge in Large Language Models},
  booktitle    = {Findings of the Association for Computational Linguistics: {EMNLP}
                  2023, Singapore, December 6-10, 2023},
  pages        = {13272--13286},
  publisher    = {Association for Computational Linguistics},
  year         = {2023},
  url          = {https://doi.org/10.18653/v1/2023.findings-emnlp.885},
  doi          = {10.18653/V1/2023.FINDINGS-EMNLP.885},
  bibsource    = {dblp computer science bibliography, https://dblp.org}
}

@article{Kumar2024Automatic,
  author       = {Rohan Kumar and
                  Youngmin Kim and
                  Sunitha Ravi and
                  Haitian Sun and
                  Christos Faloutsos and
                  Ruslan Salakhutdinov and
                  Minji Yoon},
  title        = {Automatic Question-Answer Generation for Long-Tail Knowledge},
  journal      = {CoRR},
  volume       = {abs/2403.01382},
  year         = {2024},
  url          = {https://doi.org/10.48550/arXiv.2403.01382},
  doi          = {10.48550/ARXIV.2403.01382},
  eprinttype    = {arXiv},
  eprint       = {2403.01382},
  bibsource    = {dblp computer science bibliography, https://dblp.org}
}

@inproceedings{WuPXZ0MD0LW25,
  author       = {Xiaobao Wu and
                  Liangming Pan and
                  Yuxi Xie and
                  Ruiwen Zhou and
                  Shuai Zhao and
                  Yubo Ma and
                  Mingzhe Du and
                  Rui Mao and
                  Anh Tuan Luu and
                  William Yang Wang},
  editor       = {Wanxiang Che and
                  Joyce Nabende and
                  Ekaterina Shutova and
                  Mohammad Taher Pilehvar},
  title        = {AntiLeakBench: Preventing Data Contamination by Automatically Constructing
                  Benchmarks with Updated Real-World Knowledge},
  booktitle    = {Proceedings of the 63rd Annual Meeting of the Association for Computational
                  Linguistics (Volume 1: Long Papers), {ACL} 2025, Vienna, Austria,
                  July 27 - August 1, 2025},
  pages        = {18403--18419},
  publisher    = {Association for Computational Linguistics},
  year         = {2025},
  url          = {https://aclanthology.org/2025.acl-long.901/},
  bibsource    = {dblp computer science bibliography, https://dblp.org}
}

@inproceedings{KitouniNWRBI24,
  author       = {Ouail Kitouni and
                  Niklas Nolte and
                  Adina Williams and
                  Michael Rabbat and
                  Diane Bouchacourt and
                  Mark Ibrahim},
  editor       = {Amir Globersons and
                  Lester Mackey and
                  Danielle Belgrave and
                  Angela Fan and
                  Ulrich Paquet and
                  Jakub M. Tomczak and
                  Cheng Zhang},
  title        = {The Factorization Curse: Which Tokens You Predict Underlie the Reversal
                  Curse and More},
  booktitle    = {Advances in Neural Information Processing Systems 38: Annual Conference
                  on Neural Information Processing Systems 2024, NeurIPS 2024, Vancouver,
                  BC, Canada, December 10 - 15, 2024},
  year         = {2024},
  url          = {http://papers.nips.cc/paper\_files/paper/2024/hash/cbcce87f745072c819204529be843d16-Abstract-Conference.html},
  bibsource    = {dblp computer science bibliography, https://dblp.org}
}

@inproceedings{RavichanderGW025,
  author       = {Abhilasha Ravichander and
                  Shrusti Ghela and
                  David Wadden and
                  Yejin Choi},
  editor       = {Wanxiang Che and
                  Joyce Nabende and
                  Ekaterina Shutova and
                  Mohammad Taher Pilehvar},
  title        = {HALoGEN: Fantastic {LLM} Hallucinations and Where to Find Them},
  booktitle    = {Proceedings of the 63rd Annual Meeting of the Association for Computational
                  Linguistics (Volume 1: Long Papers), {ACL} 2025, Vienna, Austria,
                  July 27 - August 1, 2025},
  pages        = {1402--1425},
  publisher    = {Association for Computational Linguistics},
  year         = {2025},
  url          = {https://aclanthology.org/2025.acl-long.71/},
  bibsource    = {dblp computer science bibliography, https://dblp.org}
}

@article{Nahum2025llms,
  author       = {Omer Nahum and
                  Nitay Calderon and
                  Orgad Keller and
                  Idan Szpektor and
                  Roi Reichart},
  title        = {Are LLMs Better than Reported? Detecting Label Errors and Mitigating
                  Their Effect on Model Performance},
  journal      = {CoRR},
  volume       = {abs/2410.18889},
  year         = {2024},
  url          = {https://doi.org/10.48550/arXiv.2410.18889},
  doi          = {10.48550/ARXIV.2410.18889},
  eprinttype    = {arXiv},
  eprint       = {2410.18889},
  bibsource    = {dblp computer science bibliography, https://dblp.org}
}

@inproceedings{Guo0G00Y24,
  author       = {Qingyan Guo and
                  Rui Wang and
                  Junliang Guo and
                  Xu Tan and
                  Jiang Bian and
                  Yujiu Yang},
  editor       = {Lun{-}Wei Ku and
                  Andre Martins and
                  Vivek Srikumar},
  title        = {Mitigating Reversal Curse in Large Language Models via Semantic-aware
                  Permutation Training},
  booktitle    = {Findings of the Association for Computational Linguistics, {ACL} 2024,
                  Bangkok, Thailand and virtual meeting, August 11-16, 2024},
  pages        = {11453--11464},
  publisher    = {Association for Computational Linguistics},
  year         = {2024},
  url          = {https://doi.org/10.18653/v1/2024.findings-acl.680},
  doi          = {10.18653/V1/2024.FINDINGS-ACL.680},
  bibsource    = {dblp computer science bibliography, https://dblp.org}
}

@inproceedings{reversal,
  author       = {Lukas Berglund and
                  Meg Tong and
                  Maximilian Kaufmann and
                  Mikita Balesni and
                  Asa Cooper Stickland and
                  Tomasz Korbak and
                  Owain Evans},
  title        = {The Reversal Curse: LLMs trained on "A is B" fail to learn "B is A"},
  booktitle    = {The Twelfth International Conference on Learning Representations,
                  {ICLR} 2024, Vienna, Austria, May 7-11, 2024},
  publisher    = {OpenReview.net},
  year         = {2024},
  url          = {https://openreview.net/forum?id=GPKTIktA0k},
  bibsource    = {dblp computer science bibliography, https://dblp.org}
}

@inproceedings{YuXDZQSZW25,
  author       = {Sicheng Yu and
                  Yuanchen Xu and
                  Cunxiao Du and
                  Yanying Zhou and
                  Minghui Qiu and
                  Qianru Sun and
                  Hao Zhang and
                  Jiawei Wu},
  editor       = {Luis Chiruzzo and
                  Alan Ritter and
                  Lu Wang},
  title        = {Reverse Modeling in Large Language Models},
  booktitle    = {Proceedings of the 2025 Conference of the Nations of the Americas
                  Chapter of the Association for Computational Linguistics: Human Language
                  Technologies, {NAACL} 2025 - Volume 2: Short Papers, Albuquerque,
                  New Mexico, April 29 - May 4, 2025},
  pages        = {306--320},
  publisher    = {Association for Computational Linguistics},
  year         = {2025},
  url          = {https://doi.org/10.18653/v1/2025.naacl-short.27},
  doi          = {10.18653/V1/2025.NAACL-SHORT.27},
  bibsource    = {dblp computer science bibliography, https://dblp.org}
}

@inproceedings{LvZ0TCWY24,
  author       = {Ang Lv and
                  Kaiyi Zhang and
                  Shufang Xie and
                  Quan Tu and
                  Yuhan Chen and
                  Ji{-}Rong Wen and
                  Rui Yan},
  editor       = {Yaser Al{-}Onaizan and
                  Mohit Bansal and
                  Yun{-}Nung Chen},
  title        = {An Analysis and Mitigation of the Reversal Curse},
  booktitle    = {Proceedings of the 2024 Conference on Empirical Methods in Natural
                  Language Processing, {EMNLP} 2024, Miami, FL, USA, November 12-16,
                  2024},
  pages        = {13603--13615},
  publisher    = {Association for Computational Linguistics},
  year         = {2024},
  url          = {https://doi.org/10.18653/v1/2024.emnlp-main.754},
  doi          = {10.18653/V1/2024.EMNLP-MAIN.754},
  bibsource    = {dblp computer science bibliography, https://dblp.org}
}

@inproceedings{WangS0G24,
  author       = {Jianing Wang and
                  Qiushi Sun and
                  Xiang Li and
                  Ming Gao},
  editor       = {Lun{-}Wei Ku and
                  Andre Martins and
                  Vivek Srikumar},
  title        = {Boosting Language Models Reasoning with Chain-of-Knowledge Prompting},
  booktitle    = {Proceedings of the 62nd Annual Meeting of the Association for Computational
                  Linguistics (Volume 1: Long Papers), {ACL} 2024, Bangkok, Thailand,
                  August 11-16, 2024},
  pages        = {4958--4981},
  publisher    = {Association for Computational Linguistics},
  year         = {2024},
  url          = {https://doi.org/10.18653/v1/2024.acl-long.271},
  doi          = {10.18653/V1/2024.ACL-LONG.271},
  bibsource    = {dblp computer science bibliography, https://dblp.org}
}

@inproceedings{0002WSLCNCZ23,
  author       = {Xuezhi Wang and
                  Jason Wei and
                  Dale Schuurmans and
                  Quoc V. Le and
                  Ed H. Chi and
                  Sharan Narang and
                  Aakanksha Chowdhery and
                  Denny Zhou},
  title        = {Self-Consistency Improves Chain of Thought Reasoning in Language Models},
  booktitle    = {The Eleventh International Conference on Learning Representations,
                  {ICLR} 2023, Kigali, Rwanda, May 1-5, 2023},
  publisher    = {OpenReview.net},
  year         = {2023},
  url          = {https://openreview.net/forum?id=1PL1NIMMrw},
  bibsource    = {dblp computer science bibliography, https://dblp.org}
}

@article{Cheng2025cot,
  author       = {Jiahao Cheng and
                  Tiancheng Su and
                  Jia Yuan and
                  Guoxiu He and
                  Jiawei Liu and
                  Xinqi Tao and
                  Jingwen Xie and
                  Huaxia Li},
  title        = {Chain-of-Thought Prompting Obscures Hallucination Cues in Large Language
                  Models: An Empirical Evaluation},
  journal      = {CoRR},
  volume       = {abs/2506.17088},
  year         = {2025},
  url          = {https://doi.org/10.48550/arXiv.2506.17088},
  doi          = {10.48550/ARXIV.2506.17088},
  eprinttype    = {arXiv},
  eprint       = {2506.17088},
  bibsource    = {dblp computer science bibliography, https://dblp.org}
}

@inproceedings{ChengLZW25,
  author       = {Xiaoxue Cheng and
                  Junyi Li and
                  Xin Zhao and
                  Ji{-}Rong Wen},
  editor       = {Wanxiang Che and
                  Joyce Nabende and
                  Ekaterina Shutova and
                  Mohammad Taher Pilehvar},
  title        = {Think More, Hallucinate Less: Mitigating Hallucinations via Dual Process
                  of Fast and Slow Thinking},
  booktitle    = {Findings of the Association for Computational Linguistics, {ACL} 2025,
                  Vienna, Austria, July 27 - August 1, 2025},
  pages        = {7979--7990},
  publisher    = {Association for Computational Linguistics},
  year         = {2025},
  url          = {https://aclanthology.org/2025.findings-acl.417/},
  bibsource    = {dblp computer science bibliography, https://dblp.org}
}

@article{Marjanovic25,
  author       = {Sara Vera Marjanovic and
                  Arkil Patel and
                  Vaibhav Adlakha and
                  Milad Aghajohari and
                  Parishad BehnamGhader and
                  Mehar Bhatia and
                  Aditi Khandelwal and
                  Austin Kraft and
                  Benno Krojer and
                  Xing Han L{\`{u}} and
                  Nicholas Meade and
                  Dongchan Shin and
                  Amirhossein Kazemnejad and
                  Gaurav Kamath and
                  Marius Mosbach and
                  Karolina Stanczak and
                  Siva Reddy},
  title        = {DeepSeek-R1 Thoughtology: Let's about {LLM} Reasoning},
  journal      = {CoRR},
  volume       = {abs/2504.07128},
  year         = {2025},
  url          = {https://doi.org/10.48550/arXiv.2504.07128},
  doi          = {10.48550/ARXIV.2504.07128},
  eprinttype    = {arXiv},
  eprint       = {2504.07128},
  bibsource    = {dblp computer science bibliography, https://dblp.org}
}

@article{Kim2025medical,
  author       = {Yubin Kim and
                  Hyewon Jeong and
                  Shan Chen and
                  Shuyue Stella Li and
                  Mingyu Lu and
                  Kumail Alhamoud and
                  Jimin Mun and
                  Cristina Grau and
                  Minseok Jung and
                  Rodrigo Gameiro and
                  Lizhou Fan and
                  Eugene Park and
                  Tristan Lin and
                  Joonsik Yoon and
                  Wonjin Yoon and
                  Maarten Sap and
                  Yulia Tsvetkov and
                  Paul Liang and
                  Xuhai Xu and
                  Xin Liu and
                  Daniel McDuff and
                  Hyeonhoon Lee and
                  Hae Won Park and
                  Samir Tulebaev and
                  Cynthia Breazeal},
  title        = {Medical Hallucinations in Foundation Models and Their Impact on Healthcare},
  journal      = {CoRR},
  volume       = {abs/2503.05777},
  year         = {2025},
  url          = {https://doi.org/10.48550/arXiv.2503.05777},
  doi          = {10.48550/ARXIV.2503.05777},
  eprinttype    = {arXiv},
  eprint       = {2503.05777},
  bibsource    = {dblp computer science bibliography, https://dblp.org}
}

@article{Zhang25scaling,
  author       = {Mike Zhang and
                  Johannes Bjerva and
                  Russa Biswas},
  title        = {Follow the Path: Reasoning over Knowledge Graph Paths to Improve LLM Factuality},
  journal      = {CoRR},
  volume       = {abs/2505.11140},
  year         = {2025},
  url          = {https://doi.org/10.48550/arXiv.2505.11140},
  doi          = {10.48550/ARXIV.2505.11140},
  eprinttype    = {arXiv},
  eprint       = {2505.11140},
  bibsource    = {dblp computer science bibliography, https://dblp.org}
}

@article{Yang25barrel,
  author       = {Junxiao Yang and
                  Jinzhe Tu and
                  Haoran Liu and
                  Xiaoce Wang and
                  Chujie Zheng and
                  Zhexin Zhang and
                  Shiyao Cui and
                  Caishun Chen and
                  Tiantian He and
                  Hongning Wang and
                  Yew{-}Soon Ong and
                  Minlie Huang},
  title        = {{BARREL:} Boundary-Aware Reasoning for Factual and Reliable LRMs},
  journal      = {CoRR},
  volume       = {abs/2505.13529},
  year         = {2025},
  url          = {https://doi.org/10.48550/arXiv.2505.13529},
  doi          = {10.48550/ARXIV.2505.13529},
  eprinttype    = {arXiv},
  eprint       = {2505.13529},
  bibsource    = {dblp computer science bibliography, https://dblp.org}
}

@article{Holtzman25prompting,
  author       = {Ari Holtzman and
                  Chenhao Tan},
  title        = {Prompting as Scientific Inquiry},
  journal      = {CoRR},
  volume       = {abs/2507.00163},
  year         = {2025},
  url          = {https://doi.org/10.48550/arXiv.2507.00163},
  doi          = {10.48550/ARXIV.2507.00163},
  eprinttype    = {arXiv},
  eprint       = {2507.00163},
  bibsource    = {dblp computer science bibliography, https://dblp.org}
}

@article{Wu25rlvr,
  author       = {Fang Wu and
                  Weihao Xuan and
                  Ximing Lu and
                  Za{\"{\i}}d Harchaoui and
                  Yejin Choi},
  title        = {The Invisible Leash: Why {RLVR} May Not Escape Its Origin},
  journal      = {CoRR},
  volume       = {abs/2507.14843},
  year         = {2025},
  url          = {https://doi.org/10.48550/arXiv.2507.14843},
  doi          = {10.48550/ARXIV.2507.14843},
  eprinttype    = {arXiv},
  eprint       = {2507.14843},
  bibsource    = {dblp computer science bibliography, https://dblp.org}
}

@article{Yao2025prone,
  author       = {Zijun Yao and
                  Yantao Liu and
                  Yanxu Chen and
                  Jianhui Chen and
                  Junfeng Fang and
                  Lei Hou and
                  Juanzi Li and
                  Tat{-}Seng Chua},
  title        = {Are Reasoning Models More Prone to Hallucination?},
  journal      = {CoRR},
  volume       = {abs/2505.23646},
  year         = {2025},
  url          = {https://doi.org/10.48550/arXiv.2505.23646},
  doi          = {10.48550/ARXIV.2505.23646},
  eprinttype    = {arXiv},
  eprint       = {2505.23646},
  bibsource    = {dblp computer science bibliography, https://dblp.org}
}

@article{Wang25jepa,
  author       = {Boshi Wang and
                  Huan Sun},
  title        = {Is the Reversal Curse a Binding Problem? Uncovering Limitations of
                  Transformers from a Basic Generalization Failure},
  journal      = {CoRR},
  volume       = {abs/2504.01928},
  year         = {2025},
  url          = {https://doi.org/10.48550/arXiv.2504.01928},
  doi          = {10.48550/ARXIV.2504.01928},
  eprinttype    = {arXiv},
  eprint       = {2504.01928},
  bibsource    = {dblp computer science bibliography, https://dblp.org}
}

@article{ReverseTraining24,
  author       = {Olga Golovneva and
                  Zeyuan Allen{-}Zhu and
                  Jason Weston and
                  Sainbayar Sukhbaatar},
  title        = {Reverse Training to Nurse the Reversal Curse},
  journal      = {CoRR},
  volume       = {abs/2403.13799},
  year         = {2024},
  url          = {https://doi.org/10.48550/arXiv.2403.13799},
  doi          = {10.48550/ARXIV.2403.13799},
  eprinttype    = {arXiv},
  eprint       = {2403.13799},
  bibsource    = {dblp computer science bibliography, https://dblp.org}
}

@misc{gpt41,
  author       = {OpenAI},
  title        = {{GPT-4.1} Model Series},
  howpublished = {Online document},
  institution  = {OpenAI},
  year         = {2025},
  url          = {https://openai.com/index/gpt-4-1/}
}

@misc{gpt5,
  author       = {OpenAI},
  title        = {{GPT‐5 System Card}},
  institution  = {OpenAI},
  year         = {2025},
  url          = {https://openai.com/index/gpt-5-system-card/}
}

@article{Lior2025reliable,
  author       = {Gili Lior and
                  Eliya Habba and
                  Shahar Levy and
                  Avi Caciularu and
                  Gabriel Stanovsky},
  title        = {ReliableEval: {A} Recipe for Stochastic {LLM} Evaluation via Method
                  of Moments},
  journal      = {CoRR},
  volume       = {abs/2505.22169},
  year         = {2025},
  url          = {https://doi.org/10.48550/arXiv.2505.22169},
  doi          = {10.48550/ARXIV.2505.22169},
  eprinttype    = {arXiv},
  eprint       = {2505.22169},
  bibsource    = {dblp computer science bibliography, https://dblp.org}
}

@inproceedings{LinGOXLY024,
  author       = {Sheng{-}Chieh Lin and
                  Luyu Gao and
                  Barlas Oguz and
                  Wenhan Xiong and
                  Jimmy Lin and
                  Scott Yih and
                  Xilun Chen},
  editor       = {Amir Globersons and
                  Lester Mackey and
                  Danielle Belgrave and
                  Angela Fan and
                  Ulrich Paquet and
                  Jakub M. Tomczak and
                  Cheng Zhang},
  title        = {{FLAME} : Factuality-Aware Alignment for Large Language Models},
  booktitle    = {Advances in Neural Information Processing Systems 38: Annual Conference
                  on Neural Information Processing Systems 2024, NeurIPS 2024, Vancouver,
                  BC, Canada, December 10 - 15, 2024},
  year         = {2024},
  url          = {http://papers.nips.cc/paper\_files/paper/2024/hash/d16152d53088ad779ffa634e7bf66166-Abstract-Conference.html},
  bibsource    = {dblp computer science bibliography, https://dblp.org}
}

@article{Ma2024bird,
  author       = {Jun{-}Yu Ma and
                  Jia{-}Chen Gu and
                  Zhen{-}Hua Ling and
                  Quan Liu and
                  Cong Liu},
  title        = {Untying the Reversal Curse via Bidirectional Language Model Editing},
  journal      = {CoRR},
  volume       = {abs/2310.10322},
  year         = {2023},
  url          = {https://doi.org/10.48550/arXiv.2310.10322},
  doi          = {10.48550/ARXIV.2310.10322},
  eprinttype    = {arXiv},
  eprint       = {2310.10322},
  bibsource    = {dblp computer science bibliography, https://dblp.org}
}

@inproceedings{GottesmanG24,
  author       = {Daniela Gottesman and
                  Mor Geva},
  editor       = {Yaser Al{-}Onaizan and
                  Mohit Bansal and
                  Yun{-}Nung Chen},
  title        = {Estimating Knowledge in Large Language Models Without Generating a
                  Single Token},
  booktitle    = {Proceedings of the 2024 Conference on Empirical Methods in Natural
                  Language Processing, {EMNLP} 2024, Miami, FL, USA, November 12-16,
                  2024},
  pages        = {3994--4019},
  publisher    = {Association for Computational Linguistics},
  year         = {2024},
  url          = {https://doi.org/10.18653/v1/2024.emnlp-main.232},
  doi          = {10.18653/V1/2024.EMNLP-MAIN.232},
  bibsource    = {dblp computer science bibliography, https://dblp.org}
}

@inproceedings{WeiYWMZ0024,
  author       = {Yifan Wei and
                  Xiaoyan Yu and
                  Yixuan Weng and
                  Huanhuan Ma and
                  Yuanzhe Zhang and
                  Jun Zhao and
                  Kang Liu},
  editor       = {Edoardo Serra and
                  Francesca Spezzano},
  title        = {Does Knowledge Localization Hold True? Surprising Differences Between
                  Entity and Relation Perspectives in Language Models},
  booktitle    = {Proceedings of the 33rd {ACM} International Conference on Information
                  and Knowledge Management, {CIKM} 2024, Boise, ID, USA, October 21-25,
                  2024},
  pages        = {4118--4122},
  publisher    = {{ACM}},
  year         = {2024},
  url          = {https://doi.org/10.1145/3627673.3679900},
  doi          = {10.1145/3627673.3679900},
  bibsource    = {dblp computer science bibliography, https://dblp.org}
}

@article{Deng2023Rephrase,
  author       = {Yihe Deng and
                  Weitong Zhang and
                  Zixiang Chen and
                  Quanquan Gu},
  title        = {Rephrase and Respond: Let Large Language Models Ask Better Questions
                  for Themselves},
  journal      = {CoRR},
  volume       = {abs/2311.04205},
  year         = {2023},
  url          = {https://doi.org/10.48550/arXiv.2311.04205},
  doi          = {10.48550/ARXIV.2311.04205},
  eprinttype    = {arXiv},
  eprint       = {2311.04205},
  bibsource    = {dblp computer science bibliography, https://dblp.org}
}

@article{benjamini2001control,
  title={The control of the false discovery rate in multiple testing under dependency},
  author={Benjamini, Yoav and Yekutieli, Daniel},
  journal={Annals of statistics},
  pages={1165--1188},
  year={2001},
  publisher={JSTOR}
}

@inproceedings{KandpalDRWR23,
  author       = {Nikhil Kandpal and
                  Haikang Deng and
                  Adam Roberts and
                  Eric Wallace and
                  Colin Raffel},
  editor       = {Andreas Krause and
                  Emma Brunskill and
                  Kyunghyun Cho and
                  Barbara Engelhardt and
                  Sivan Sabato and
                  Jonathan Scarlett},
  title        = {Large Language Models Struggle to Learn Long-Tail Knowledge},
  booktitle    = {International Conference on Machine Learning, {ICML} 2023, 23-29 July
                  2023, Honolulu, Hawaii, {USA}},
  series       = {Proceedings of Machine Learning Research},
  volume       = {202},
  pages        = {15696--15707},
  publisher    = {{PMLR}},
  year         = {2023},
  url          = {https://proceedings.mlr.press/v202/kandpal23a.html},
  bibsource    = {dblp computer science bibliography, https://dblp.org}
}

@article{Lin2025active,
  author       = {Jessy Lin and
                  Vincent{-}Pierre Berges and
                  Xilun Chen and
                  Wen{-}Tau Yih and
                  Gargi Ghosh and
                  Barlas Oguz},
  title        = {Learning Facts at Scale with Active Reading},
  journal      = {CoRR},
  volume       = {abs/2508.09494},
  year         = {2025},
  url          = {https://doi.org/10.48550/arXiv.2508.09494},
  doi          = {10.48550/ARXIV.2508.09494},
  eprinttype    = {arXiv},
  eprint       = {2508.09494},
  bibsource    = {dblp computer science bibliography, https://dblp.org}
}

@article{ElazarKRRHSG21,
  author       = {Yanai Elazar and
                  Nora Kassner and
                  Shauli Ravfogel and
                  Abhilasha Ravichander and
                  Eduard H. Hovy and
                  Hinrich Sch{\"{u}}tze and
                  Yoav Goldberg},
  title        = {Measuring and Improving Consistency in Pretrained Language Models},
  journal      = {Trans. Assoc. Comput. Linguistics},
  volume       = {9},
  pages        = {1012--1031},
  year         = {2021},
  url          = {https://doi.org/10.1162/tacl\_a\_00410},
  doi          = {10.1162/TACL\_A\_00410},
  bibsource    = {dblp computer science bibliography, https://dblp.org}
}

@article{JiangXAN20,
  author       = {Zhengbao Jiang and
                  Frank F. Xu and
                  Jun Araki and
                  Graham Neubig},
  title        = {How Can We Know What Language Models Know},
  journal      = {Trans. Assoc. Comput. Linguistics},
  volume       = {8},
  pages        = {423--438},
  year         = {2020},
  url          = {https://doi.org/10.1162/tacl\_a\_00324},
  doi          = {10.1162/TACL\_A\_00324},
  bibsource    = {dblp computer science bibliography, https://dblp.org}
}

@inproceedings{triviaqa,
  author       = {Mandar Joshi and
                  Eunsol Choi and
                  Daniel S. Weld and
                  Luke Zettlemoyer},
  editor       = {Regina Barzilay and
                  Min{-}Yen Kan},
  title        = {TriviaQA: {A} Large Scale Distantly Supervised Challenge Dataset for
                  Reading Comprehension},
  booktitle    = {Proceedings of the 55th Annual Meeting of the Association for Computational
                  Linguistics, {ACL} 2017, Vancouver, Canada, July 30 - August 4, Volume
                  1: Long Papers},
  pages        = {1601--1611},
  publisher    = {Association for Computational Linguistics},
  year         = {2017},
  url          = {https://doi.org/10.18653/v1/P17-1147},
  doi          = {10.18653/V1/P17-1147},
  bibsource    = {dblp computer science bibliography, https://dblp.org}
}

@inproceedings{GekhmanYAEFRH24,
  author       = {Zorik Gekhman and
                  Gal Yona and
                  Roee Aharoni and
                  Matan Eyal and
                  Amir Feder and
                  Roi Reichart and
                  Jonathan Herzig},
  editor       = {Yaser Al{-}Onaizan and
                  Mohit Bansal and
                  Yun{-}Nung Chen},
  title        = {Does Fine-Tuning LLMs on New Knowledge Encourage Hallucinations?},
  booktitle    = {Proceedings of the 2024 Conference on Empirical Methods in Natural
                  Language Processing, {EMNLP} 2024, Miami, FL, USA, November 12-16,
                  2024},
  pages        = {7765--7784},
  publisher    = {Association for Computational Linguistics},
  year         = {2024},
  url          = {https://doi.org/10.18653/v1/2024.emnlp-main.444},
  doi          = {10.18653/V1/2024.EMNLP-MAIN.444},
  bibsource    = {dblp computer science bibliography, https://dblp.org}
}

@inproceedings{MengBAB22,
  author       = {Kevin Meng and
                  David Bau and
                  Alex Andonian and
                  Yonatan Belinkov},
  editor       = {Sanmi Koyejo and
                  S. Mohamed and
                  A. Agarwal and
                  Danielle Belgrave and
                  K. Cho and
                  A. Oh},
  title        = {Locating and Editing Factual Associations in {GPT}},
  booktitle    = {Advances in Neural Information Processing Systems 35: Annual Conference
                  on Neural Information Processing Systems 2022, NeurIPS 2022, New Orleans,
                  LA, USA, November 28 - December 9, 2022},
  year         = {2022},
  url          = {https://papers.nips.cc/paper_files/paper/2022/hash/6f1d43d5a82a37e89b0665b33bf3a182-Abstract-Conference.html},
  bibsource    = {dblp computer science bibliography, https://dblp.org}
}

@article{Gekhman25Inside,
  author       = {Zorik Gekhman and
                  Eyal Ben{-}David and
                  Hadas Orgad and
                  Eran Ofek and
                  Yonatan Belinkov and
                  Idan Szpektor and
                  Jonathan Herzig and
                  Roi Reichart},
  title        = {Inside-Out: Hidden Factual Knowledge in LLMs},
  journal      = {CoRR},
  volume       = {abs/2503.15299},
  year         = {2025},
  url          = {https://doi.org/10.48550/arXiv.2503.15299},
  doi          = {10.48550/ARXIV.2503.15299},
  eprinttype    = {arXiv},
  eprint       = {2503.15299},
  bibsource    = {dblp computer science bibliography, https://dblp.org}
}

@inproceedings{HabbaAIPBCSS25,
  author       = {Eliya Habba and
                  Ofir Arviv and
                  Itay Itzhak and
                  Yotam Perlitz and
                  Elron Bandel and
                  Leshem Choshen and
                  Michal Shmueli{-}Scheuer and
                  Gabriel Stanovsky},
  editor       = {Wanxiang Che and
                  Joyce Nabende and
                  Ekaterina Shutova and
                  Mohammad Taher Pilehvar},
  title        = {{DOVE:} {A} Large-Scale Multi-Dimensional Predictions Dataset Towards
                  Meaningful {LLM} Evaluation},
  booktitle    = {Findings of the Association for Computational Linguistics, {ACL} 2025,
                  Vienna, Austria, July 27 - August 1, 2025},
  pages        = {11744--11763},
  publisher    = {Association for Computational Linguistics},
  year         = {2025},
  url          = {https://aclanthology.org/2025.findings-acl.611/},
  bibsource    = {dblp computer science bibliography, https://dblp.org}
}

@article{MizrahiKMDSS24,
  author       = {Moran Mizrahi and
                  Guy Kaplan and
                  Dan Malkin and
                  Rotem Dror and
                  Dafna Shahaf and
                  Gabriel Stanovsky},
  title        = {State of What Art? {A} Call for Multi-Prompt {LLM} Evaluation},
  journal      = {Trans. Assoc. Comput. Linguistics},
  volume       = {12},
  pages        = {933--949},
  year         = {2024},
  url          = {https://doi.org/10.1162/tacl\_a\_00681},
  doi          = {10.1162/TACL\_A\_00681},
  bibsource    = {dblp computer science bibliography, https://dblp.org}
}

@article{KwiatkowskiPRCP19,
  author       = {Tom Kwiatkowski and
                  Jennimaria Palomaki and
                  Olivia Redfield and
                  Michael Collins and
                  Ankur P. Parikh and
                  Chris Alberti and
                  Danielle Epstein and
                  Illia Polosukhin and
                  Jacob Devlin and
                  Kenton Lee and
                  Kristina Toutanova and
                  Llion Jones and
                  Matthew Kelcey and
                  Ming{-}Wei Chang and
                  Andrew M. Dai and
                  Jakob Uszkoreit and
                  Quoc Le and
                  Slav Petrov},
  title        = {Natural Questions: a Benchmark for Question Answering Research},
  journal      = {Trans. Assoc. Comput. Linguistics},
  volume       = {7},
  pages        = {452--466},
  year         = {2019},
  url          = {https://doi.org/10.1162/tacl\_a\_00276},
  doi          = {10.1162/TACL\_A\_00276},
  bibsource    = {dblp computer science bibliography, https://dblp.org}
}

@article{He2025pop,
  author       = {Yuan He and
                  Bailan He and
                  Zifeng Ding and
                  Alisia Maria Lupidi and
                  Yuqicheng Zhu and
                  Shuo Chen and
                  Caiqi Zhang and
                  Jiaoyan Chen and
                  Yunpu Ma and
                  Volker Tresp and
                  Ian Horrocks},
  title        = {Supposedly Equivalent Facts That Aren't? Entity Frequency in
                  Pre-training Induces Asymmetry in LLMs},
  journal      = {CoRR},
  volume       = {abs/2503.22362},
  year         = {2025},
  url          = {https://doi.org/10.48550/arXiv.2503.22362},
  doi          = {10.48550/ARXIV.2503.22362},
  eprinttype    = {arXiv},
  eprint       = {2503.22362},
  bibsource    = {dblp computer science bibliography, https://dblp.org}
}

@inproceedings{LinFL0L00WY24,
  author       = {Zhengkai Lin and
                  Zhihang Fu and
                  Kai Liu and
                  Liang Xie and
                  Binbin Lin and
                  Wenxiao Wang and
                  Deng Cai and
                  Yue Wu and
                  Jieping Ye},
  editor       = {Amir Globersons and
                  Lester Mackey and
                  Danielle Belgrave and
                  Angela Fan and
                  Ulrich Paquet and
                  Jakub M. Tomczak and
                  Cheng Zhang},
  title        = {Delving into the Reversal Curse: How Far Can Large Language Models
                  Generalize?},
  booktitle    = {Advances in Neural Information Processing Systems 38: Annual Conference
                  on Neural Information Processing Systems 2024, NeurIPS 2024, Vancouver,
                  BC, Canada, December 10 - 15, 2024},
  year         = {2024},
  url          = {http://papers.nips.cc/paper\_files/paper/2024/hash/36b6180f3dab4025ba763e853b044814-Abstract-Conference.html},
  bibsource    = {dblp computer science bibliography, https://dblp.org}
}

@inproceedings{ZhuHZJJTR24,
  author       = {Hanlin Zhu and
                  Baihe Huang and
                  Shaolun Zhang and
                  Michael I. Jordan and
                  Jiantao Jiao and
                  Yuandong Tian and
                  Stuart J. Russell},
  editor       = {Amir Globersons and
                  Lester Mackey and
                  Danielle Belgrave and
                  Angela Fan and
                  Ulrich Paquet and
                  Jakub M. Tomczak and
                  Cheng Zhang},
  title        = {Towards a Theoretical Understanding of the 'Reversal Curse' via Training
                  Dynamics},
  booktitle    = {Advances in Neural Information Processing Systems 38: Annual Conference
                  on Neural Information Processing Systems 2024, NeurIPS 2024, Vancouver,
                  BC, Canada, December 10 - 15, 2024},
  year         = {2024},
  url          = {http://papers.nips.cc/paper\_files/paper/2024/hash/a4b95476f673e6e538f80862f622ba2f-Abstract-Conference.html},
  bibsource    = {dblp computer science bibliography, https://dblp.org}
}

@article{Simhi25hack,
  author       = {Adi Simhi and
                  Jonathan Herzig and
                  Itay Itzhak and
                  Dana Arad and
                  Zorik Gekhman and
                  Roi Reichart and
                  Fazl Barez and
                  Gabriel Stanovsky and
                  Idan Szpektor and
                  Yonatan Belinkov},
  title        = {{HACK:} Hallucinations Along Certainty and Knowledge Axes},
  journal      = {CoRR},
  volume       = {abs/2510.24222},
  year         = {2025},
  url          = {https://doi.org/10.48550/arXiv.2510.24222},
  doi          = {10.48550/ARXIV.2510.24222},
  eprinttype   = {arXiv},
  eprint       = {2510.24222},
  bibsource    = {dblp computer science bibliography, https://dblp.org}
}


\clearpage
\appendix

\renewcommand\thepart{}
\renewcommand\partname{}
\mtcsettitle{parttoc}{}
\part{Appendix}
\parttoc

\section{WikiProfile -- Full Details}
\label{sec:dataset}

In this section, we describe WikiProfile and its construction in detail. WikiProfile comprises ten tasks defined for each fact extracted from a Wikipedia source text; their properties and examples are summarized in Table~\ref{tab:examples}. The tasks span three formats: text completion, closed-book question answering, and multiple-choice question answering. Two tasks measure encoding (proposition completion and a contextual question), and four tasks measure knowledge (direct and reverse questions, each with two phrasings). Each knowledge task additionally has a multiple-choice variant.

WikiProfile is constructed using a fully automated pipeline powered by a prompted LLM, Gemini-2.5-Pro with thinking enabled \citep{gemini}. Our goal is to provide a framework that can be readily applied to other corpora, including domain-specific ones, enabling researchers to identify knowledge profiles and characterize LLM behavior. Prompts were developed through manual optimization on a small subset of approximately 100 documents and facts, which is excluded from the final dataset. This process involved iterative error analysis and refinement.

\subsection{Tasks and Properties}
\label{sub:properties}

We begin by outlining the properties that guided the design of WikiProfile.

\paragraph{Required Properties}
All questions must satisfy the following properties to ensure methodological rigor and enable comparison against a single verifiable gold answer:
\begin{itemize}[nosep, leftmargin=*]
\item \textbf{Answerable:} A correct answer exists.
\item \textbf{Unambiguous:} The question is self-contained and precisely phrased.
\item \textbf{Specific:} The expected answer type and level of detail are clearly specified.
\item \textbf{Single-answer:} Exactly one answer is correct.
\item \textbf{Not time-sensitive:} The answer is stable over time.
\end{itemize}

\paragraph{Controlled Properties}
These properties define systematic variations across tasks. By manipulating them, we examine factors that affect factual knowledge.
\begin{itemize}[nosep, leftmargin=*]
\item \textbf{Format:} Completion, closed-book, or multiple-choice.
\item \textbf{Context:} The amount of auxiliary information provided, ranging from full source context (used to measure encoding) to minimal context required for disambiguation.
\item \textbf{Verbatim:} The degree of linguistic overlap with the source text, spanning exact phrasing (for completion tasks), high-verbatim questions that reuse source phrases and terms, and low-verbatim, natural user-like queries.
\item \textbf{Direction:} The relational direction of the question, defined by the ordering of the entities in the source text.
\end{itemize}

\begin{table}[t]
\caption{The properties of the ten tasks in WikiProfile, including examples. All tasks were generated from the Wikipedia page for `Love and Money'. The extracted fact is: \textit{`The band Love and Money was formed by three former members of Friends Again, along with the bassist Bobby Paterson'}.}
\label{tab:examples}
\centering
\Large
\begin{adjustbox}{width=0.48\textwidth}
\begin{tabular}{p{3.5cm}cccc}
\toprule
\textbf{Task} & \textbf{Format} & \textbf{Context} & \textbf{Verbatim} & \textbf{Direction} \\
\midrule
\textbf{Completion} & \cellcolor{cCompletion!40!white}Completion & \cellcolor{cFull!40!white}Full & \cellcolor{cExact!40!white}Exact & \cellcolor{cDirect!40!white}Direct\\
\multicolumn{5}{p{14cm}}{{\fontfamily{ppl}\selectfont\large ``Love and Money are a Scottish rock/soul/funk band, formed in 1985 in Glasgow, Scotland. The band was formed by three former members of Friends Again (singer-songwriter and guitarist James Grant, drummer Stuart Kerr and keyboardist Paul McGeechan) along with bassist ...''}} \\
\midrule

\textbf{Contextual} & \cellcolor{cClosedBook!40!white}Closed-book& \cellcolor{cFull!40!white}Full & \cellcolor{cExact!40!white}High & \cellcolor{cDirect!40!white}Direct\\
\multicolumn{5}{p{14cm}}{{\fontfamily{ppl}\selectfont\large ``Love and Money are a Scottish rock/soul/funk band, formed in 1985 in Glasgow, Scotland. The band was formed by three former members of Friends Again (singer-songwriter and guitarist James Grant, drummer Stuart Kerr and keyboardist Paul McGeechan). What is the name of the bassist who, along with them, formed the band Love and Money?''}} \\
\midrule

\textbf{Direct} & \cellcolor{cClosedBook!40!white}Closed-book& \cellcolor{cMinimal!40!white}Minimal& \cellcolor{cHigh!40!white}High & \cellcolor{cDirect!40!white}Direct\\
\multicolumn{5}{p{14cm}}{{\fontfamily{ppl}\selectfont\large ``What is the name of the bassist who, along with three former members of Friends Again, formed the band Love and Money?''}} \\
\textbf{MC Direct} & \cellcolor{cMultipleChoice!40!white}Multiple-choice& \cellcolor{cMinimal!40!white}Minimal& \cellcolor{cHigh!40!white}High & \cellcolor{cDirect!40!white}Direct\\
\multicolumn{5}{p{14cm}}{{\fontfamily{ppl}\selectfont\large A. Bobby Paterson B. James Grant C. Neil Cunningham D. Stuart Kerr}} \\
\midrule

\textbf{Direct Natural} & \cellcolor{cClosedBook!40!white}Closed-book& \cellcolor{cMinimal!40!white}Minimal& \cellcolor{cNatural!40!white}Natural & \cellcolor{cDirect!40!white}Direct \\
\multicolumn{5}{p{14cm}}{{\fontfamily{ppl}\selectfont\large ``Who was the bassist that formed the band Love and Money with the three guys from Friends Again?''}} \\
\textbf{MC Dir Nat} & \cellcolor{cMultipleChoice!40!white}Multiple-choice& \cellcolor{cMinimal!40!white}Minimal& \cellcolor{cNatural!40!white}Natural & \cellcolor{cDirect!40!white}Direct \\
\multicolumn{5}{p{14cm}}{{\fontfamily{ppl}\selectfont\large A. Bobby Paterson B. James Grant C. Neil Cunningham D. Stuart Kerr}} \\

\midrule
\textbf{Reverse} & \cellcolor{cClosedBook!40!white}Closed-book & \cellcolor{cMinimal!40!white}Minimal& \cellcolor{cHigh!40!white}High& \cellcolor{cReverse!40!white}Reverse \\
\multicolumn{5}{p{14cm}}{{\fontfamily{ppl}\selectfont\large ``What is the name of the band that was formed by bassist Bobby Paterson and three former members of Friends Again?''}} \\
\textbf{MC Reverse} & \cellcolor{cMultipleChoice!40!white}Multiple-choice & \cellcolor{cMinimal!40!white}Minimal & \cellcolor{cHigh!40!white}High & \cellcolor{cReverse!40!white}Reverse \\
\multicolumn{5}{p{14cm}}{{\fontfamily{ppl}\selectfont\large A. Love and Money B. Friends Again C. Set the Tone D. Deacon Blue}} \\

\midrule
\textbf{Rev Natural} & \cellcolor{cClosedBook!40!white}Closed-book & \cellcolor{cMinimal!40!white}Minimal& \cellcolor{cNatural!40!white}Natural& \cellcolor{cReverse!40!white}Reverse \\
\multicolumn{5}{p{14cm}}{{\fontfamily{ppl}\selectfont\large ``Which band was formed by Bobby Paterson with three guys from Friends Again?''}} \\
\textbf{MC Rev Nat} & \cellcolor{cMultipleChoice!40!white}Multiple-choice & \cellcolor{cMinimal!40!white}Minimal & \cellcolor{cNatural!40!white}Natural & \cellcolor{cReverse!40!white}Reverse \\
\multicolumn{5}{p{14cm}}{{\fontfamily{ppl}\selectfont\large A. Love and Money B. Friends Again C. Set the Tone D. Deacon Blue}} \\




\bottomrule
\end{tabular}
\end{adjustbox}
\vspace{-0.8em}
\end{table}
\begin{figure*}[t]
    \centering
    \includegraphics[width=0.75\textwidth]{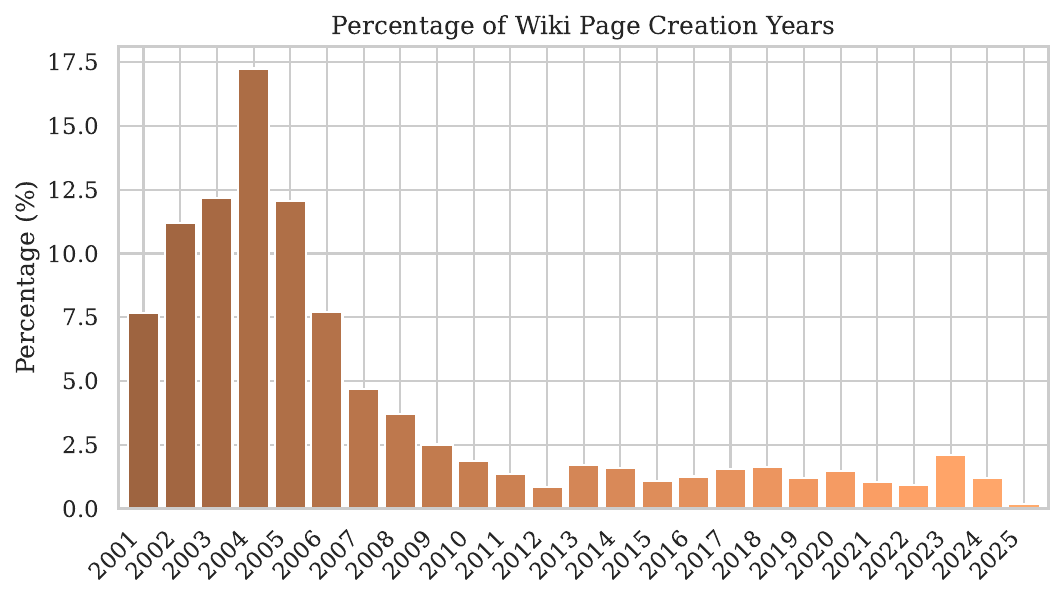}
    \caption{Distribution of Wikipedia pages according to their year of creation.}
    \label{fig:years}
\vspace{-0.8em}
\end{figure*}

\paragraph{Tasks}
Each fact is represented as a relation between a subject and an object entity, where the subject precedes the object in the source text. The \textit{left context} consists of all text preceding the object entity.

We use two tasks to measure encoding:
\smallskip
\begin{itemize}[nosep, leftmargin=*]
\item \textbf{Completion:} Proposition completion of the left context.
\item \textbf{Contextual:} A high-verbatim question posed over the same left context.
\end{itemize}
Completion can be ambiguous for post-trained LLMs, which may generate plausible continuations without stating the target fact. The contextual question explicitly elicits the fact while preserving the original context, mitigating this issue.

We use four tasks to measure knowledge:
\smallskip
\begin{itemize}[nosep, leftmargin=*]
\item \textbf{Direct:} A closed-book question, whose answer is the object entity.
\item \textbf{Direct Natural:} A conversational rephrasing of the Direct question.
\item \textbf{Reverse:} An inverted question where the object becomes the subject.
\item \textbf{Reverse Natural:} A conversational rephrasing of the Reverse question.
\end{itemize}

Finally, we reformulate each of the four closed-book questions as multiple-choice questions with four answer options. The distractors are designed to be plausible and realistic. These tasks evaluate sensitivity to task format and, crucially, the ability to verify facts. This shift from generation to verification probes a distinct form of factual knowledge.

\subsection{Fact Extraction Process}
\label{sub:extraction}

We aim to extract facts that are likely encoded by LLMs (to enable measurement of encoding even for small LLMs) yet remain non-trivial. Notably, we intentionally avoid extracting facts in a schema-based triplet format (subject, relation, object). Our goal is to generate non-trivial questions that are not limited to simple structural relations and may involve more complex ones (e.g., \textit{``What is the condition that caused Erika Nordby to spend two hours without a heartbeat before she was revived?''}).

\begin{figure}[!t]
    \centering
    \includegraphics[width=0.47\textwidth]{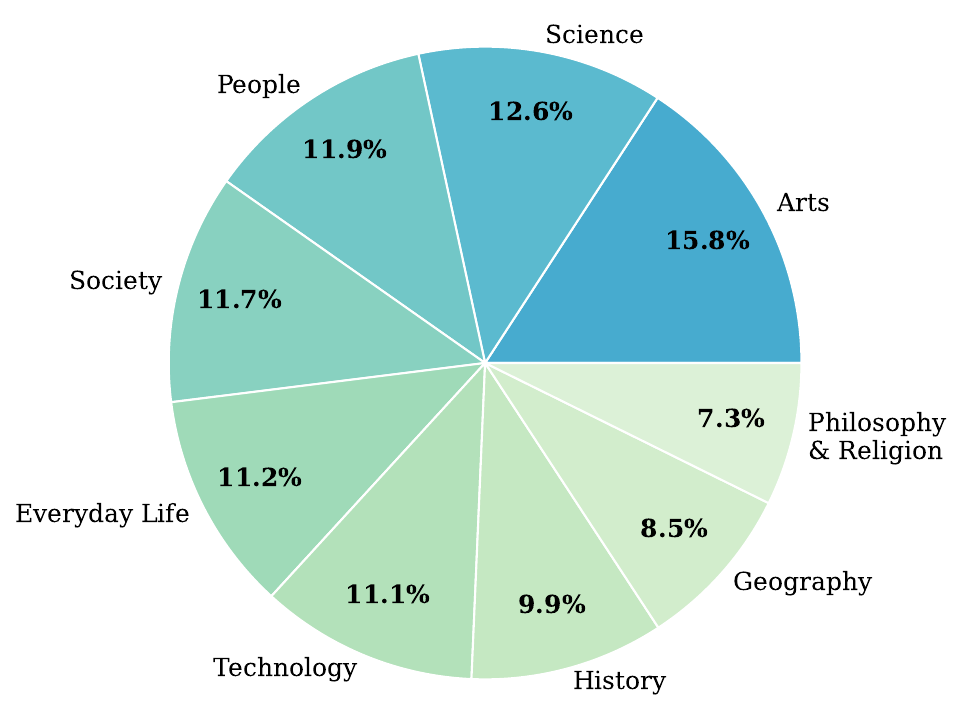}
    \caption{Distribution of Wikipedia page categories.}
    \label{fig:categories}
\vspace{-0.8em}
\end{figure}

We therefore extract facts from Wikipedia pages, which were likely a key component of LLM pre-training data. We sample 10,000 Wikipedia page summaries (first paragraphs), along with page popularity statistics.\footnote{Popularity is based on page views from 2023-02 to 2025-06, according to \url{https://analytics.wikimedia.org/published/datasets/country_project_page/00_README.html}} We retain only documents with at least 50 words. The distribution of page creation dates is shown in Figure~\ref{fig:years}. Pages are then categorized into nine Wikipedia Vital Article categories.\footnote{\url{https://en.wikipedia.org/wiki/Wikipedia:Vital_articles/Level/3}}  The distribution of the categories is presented in Figure~\ref{fig:categories}. 

\begin{figure}[!t]
    \centering
    \includegraphics[width=0.46\textwidth]{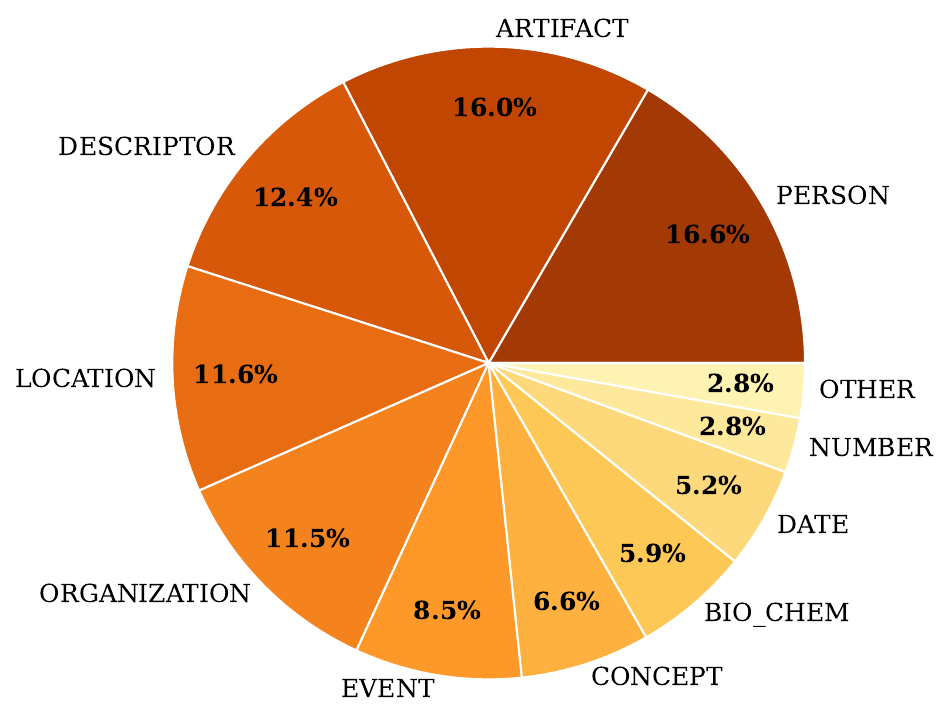}
    \caption{Distribution of object entity types. For this figure, we grouped together some types: ARTIFACT (WORK OF ART, PRODUCT), BIO\_CHEM (ORGANISM, CONDITION, SUBSTANCE), CONCEPT (IDEA, LANGUAGE, LAW), DESCRIPTOR (CLASSIFICATION, GROUP, IDENTIFIER, TITLE).}
    \label{fig:entities}
\vspace{-0.8em}
\end{figure}

Fact extraction proceeds in three stages. First, the LLM performs NER, identifying up to eight entities per document and assigning one of 19 predefined types (Box~\ref{box:prompt_ner}). The purpose of this stage is to generate a diverse set of entities that can serve as answers. The distribution of the entity types is presented in Figure~\ref{fig:entities}. 

Second, the LLM selects up to three entities suitable for proposition completion. Suitability is defined by whether the left context constrains the answer to a \textit{single, specific, and unique correct entity}, such that alternative answers would be factually inconsistent. In addition, the entity is neither time-sensitive nor trivial, making the proposition-completion task challenging. For additional details, including exact criteria and examples, see the prompt in Box~\ref{box:prompt_selection}. This selection step yields 33,709 candidate object entities. Third, candidates are verified using three prompts, and only those deemed suitable are retained (Boxes~\ref{box:prompt_verification1}, \ref{box:prompt_verification2},  \ref{box:prompt_difficulty}). This verification process results in 12,031 entities (35.6\%). Finally, we downsample the dataset to 5,000 entities by balancing fact categories and entity types.

\subsection{Question Generation Process}
\label{sub:generation}

We use three stages of generation, refinement, and filtering to ensure that questions satisfy the required properties. The refinement stage resolves ambiguity and enforces specificity, for example, by clarifying the required answer type and adding minimal disambiguating context when needed. The filtering stage is grounded in Google Search to recall plausible alternative answers and to discard questions that admit multiple answers.

We first apply these three stages to construct the \textit{Direct} question. Based on the Direct question, we generate the \textit{Reverse} question, which is then refined and filtered. From the Direct question, we also generate the \textit{Direct Natural} and \textit{Contextual} questions, which are subsequently filtered. Similarly, we generate and filter the \textit{Reverse Natural} questions. Finally, we generate the multiple-choice variants. Below, we describe the three-stage process used to construct the \textit{Direct} question, and then elaborate on the remaining types.

\paragraph{Generation}
The \textit{Direct} questions are generated by instructing the LLM to design a question based on the left context such that the only correct answer is the object entity. The question uses only the information necessary for disambiguation (\textit{minimal context}) while preserving exact phrasing from the source text (\textit{high verbatim}). We also specify a set of requirements in the prompt (Box~\ref{box:prompt_direct}), including ensuring that the question is self-contained and does not include pronouns. We then refine the Direct questions using two additional prompts: one to improve specificity and one to enforce minimalism.

\paragraph{Refinement}
The specification prompt (Box~\ref{box:prompt_specification}) instructs the LLM to evaluate a question's specificity: whether it clearly defines the required level of detail in the answer, and to revise it if necessary. The LLM should recommend rejecting a question if it cannot be fixed, either because revising it would leak a non-trivial clue that makes the question too easy or because the question is too vague and would require a full rewrite. This stage rejected 45 questions, leaving Direct questions for 4955 facts. The minimalism prompt (Box~\ref{box:prompt_minimalism}) instructs the LLM to revise a question if it contains non-essential information that can be removed while preserving disambiguation and specificity.

\paragraph{Filtering}
Some questions may still be vague or ambiguous, or admit multiple answers, according to sources outside the Wikipedia context. To filter these cases, we use Gemini-2.5-Pro with thinking grounded in Google Search. Each question is answered with search grounding, and a prompted LLM (Box~\ref{box:prompt_search}) evaluates whether the question is suitable. If the grounded answer requests clarification, provides multiple possible answers, or varies by interpretation, the question is rejected. This process removes 10.6\% of Direct questions. Filtering applies to all five question types: once a fact is discarded due to any of its questions, it is not reconsidered. The number of removed questions is reported in the relevant paragraphs below.

\paragraph{Reverse Questions}
For reverse questions, the object entity becomes the new subject, and the subject entity of the Direct question becomes the answer. Given the left context, the Direct question, and the target answer, we instruct the LLM to identify the subject entity and generate a reverse question that satisfies all required properties (Box~\ref{box:prompt_reverse}). In some cases, additional information from the left context is required to ensure that the reverse question is unambiguous and has a single correct answer. For example, consider the direct question: \textit{``At which event did the film `There's Still Tomorrow' premiere?''} with the answer \textit{`The 18th Rome Film Festival'}. The subject entity is the film \textit{`There's Still Tomorrow'}. A naive reverse question such as \textit{``Which film premiered at the 18th Rome Film Festival?''} is unsuitable, since multiple films could have premiered there. The LLM should therefore incorporate additional context, such as the director's name, yielding: \textit{``Which film directed by Paola Cortellesi premiered at the 18th Rome Film Festival?''}

The LLM may also reject generating a reverse question if it cannot do so without making it trivial (see the fourth example in Box~\ref{box:prompt_reverse}). This occurs in 2.2\% of cases. After generation, we apply the refinement stage, which improves specification and enforces minimalism, rejecting an additional 3.6\% of Reverse questions. We then proceed to filtering. Before applying Google Search–grounded filtering, we verify that the \textit{Direct} and \textit{Reverse} questions form a valid pair and are non-trivial, using the prompt in Box~\ref{box:prompt_reverse_verification}. This step removes 31.5\% of facts. Finally, the filtering stage removes an additional 13.1\% of facts.

\paragraph{Natural and Contextual Questions}
We instruct the LLM to rewrite each \textit{Direct} and \textit{Reverse} question as \textit{Direct Natural} and \textit{Reverse Natural}, following Box~\ref{box:prompt_natural}. The \textit{Contextual} question is constructed using Box~\ref{box:prompt_contextual}.\footnote{We do not generate contextual versions for reverse questions, as these tend to be too easy and the context often includes the answer, frequently the page title.} It combines the left context (excluding the final incomplete sentence) with the high-verbatim direct question, with minor edits for fluency and to avoid redundancy.

We then repeat the Google Search–based filtering for the \textit{Direct Natural}, \textit{Reverse Natural}, and \textit{Contextual} questions. This removes an additional 5.5\% of facts, leaving 2357 facts for which all five question types are available. Finally, we remove facts with lengthy questions (over 40 words) or short left contexts (below 30 words), yielding 2200 facts.

\subsection{Manual Quality Validation}
\label{app_sub:manual}

To ensure high data quality, we manually validate questions by adapting the protocol of \citet{Nahum2025llms}, which prioritizes instances in which an LLM ensemble fails. Specifically, we manually examine a fact and its ten associated questions if the four frontier LLMs (Gemini-3-Pro/Flash, GPT-5, and GPT-4.1; evaluated with and without thinking) unanimously fail to correctly answer at least one of the questions, i.e., $g(q)\leq0.5$. This process flagged 237 facts (10.7\%). Upon manual review, we identified 43 ($<$2\%) facts and their associated questions as low quality, and they were subsequently removed from the benchmark. To further validate our data, we examined a subset of 50 facts in which three of the four models provided incorrect answers. We found only 3 errors in this subset. Together, these results suggest that low-quality questions are uncommon in the remaining data and that the benchmark meets a high standard.
\section{Thinking and Recall}
\label{app_sec:thinking}

In \S\ref{sub:thinking}, we argued that thinking primarily facilitates recall of encoded knowledge. Here we present the full analysis, including formal definitions, detailed results across all models, and a discussion of limitations.

\subsection{Three Possible Mechanisms}
\label{app_sub:mechanisms}

We consider three mechanisms by which thinking may improve factual performance on our benchmark. Each mechanism makes distinct empirical predictions, which we evaluate in the subsequent sections.

\smallskip\noindent\textbf{(i) Response diversity.} Thinking may alter the output distribution by increasing its variance. With higher variance, the probability that at least one of $n$ samples is correct increases -- a sampling effect analogous to the relationship between response diversity and $\text{pass}@k$ \citep{0002WSLCNCZ23, Yang25barrel}. Importantly, this mechanism does not reflect improved access to knowledge; rather, it is a statistical consequence of broader exploration of the output space.

\emph{Example:} Without thinking, a model asked ``Where did Oasis play their first show?'' might produce the same incorrect answer in 8 out of 8 samples. Thinking trajectories, by incorporating additional sampling steps, may lead to additional answers, some correct, most not, yielding, say, 2 out of 8 correct samples. Availability of the correct answer has improved (from 0 to 2 correct), but robustness has not: the model does not reliably converge to the correct answer.

\emph{Prediction:} Response diversity increases \emph{availability} (the chance of at least one correct answer across samples) but cannot increase \emph{robustness} (consistent correctness), since higher variance cannot produce reliable convergence to the same correct output.

\smallskip\noindent\textbf{(ii) Inference.} Thinking may enable multi-step derivation from other encoded facts. By chaining together related knowledge, the LLM can produce an answer that it does not store directly. This mechanism relies on the availability of related encoded premises, not on the encoding status of the target fact itself.

\emph{Example:} A model may not encode that Oasis played their first gig at the Boardwalk. However, through thinking, it might chain together other related encoded facts: ``Oasis formed in Manchester'' $\rightarrow$ ``the Boardwalk was a famous music club in Manchester during the 1990s'' $\rightarrow$ ``the Boardwalk hosted early gigs by emerging bands.'' This inference produces the correct answer without direct recall.

\emph{Prediction:} Inference gains are regardless of whether the specific target fact is encoded, since it operates over \emph{other related} encoded facts. We therefore expect inference to produce correct answers even for non-encoded facts. 

\smallskip\noindent\textbf{(iii) Recall facilitation.} Thinking may help the model locate and recall a fact that is encoded but not accessible via direct generation. The fact is stored in the model's parameters, as verified by the encoding tasks, but the bare question fails to activate its recall. Thinking provides additional computation that bridges this gap.

\emph{Example:} An LLM can complete the relevant Wikipedia proposition about Oasis's first gig when given the left context. However, when asked ``Where did Oasis play their first show?'' without context, direct generation fails. With thinking, the model can generate intermediate context (e.g., recalling details about the band's early history in Manchester) and perform additional computations that prime recall of the stored answer.

\emph{Prediction:} Recall facilitation, by definition, requires the target fact to be encoded. It should therefore benefit only encoded facts. Unlike response diversity, it can improve both availability and robustness, since successful recall of a stored fact yields consistent correctness.

\subsection{Measuring Availability and Robustness} 
\label{app_sub:correct_p} 

To operationalize the distinction between availability and robustness, we propose $\text{correct}\!\geq\!p$, a metric that captures both aspects along a continuous spectrum. It is defined as the proportion of questions whose estimated probability of correctness across samples is at least $p$

For each question, we draw $n$ independent samples and observe $k$ correct responses, yielding an empirical correctness rate $\hat{p}=k/n$. We then estimate the probability that, in a repeated run of $n$ samples, at least a $p$ fraction of the answers would be correct. Let $m=\lceil pn\rceil$. Using a binomial model $K \sim \mathrm{Binomial}(n,\hat{p})$, then:
\[
\hat{P}\!\left(K \ge m\right)
\;=\;
\sum_{j=m}^{n} \binom{n}{j}\, \hat{p}^{\,j}\, (1-\hat{p})^{\,n-j}.
\]
Aggregating over questions, $\text{correct}\!\geq\!p$ measures the fraction of questions for which this estimated probability exceeds 0.5, i.e., $\hat{P}\!\left(K \ge \lceil pn\rceil\right)>0.5$. 

The metric makes the trade-off between availability and robustness clear. At low values of $p$ (e.g., $p=0.125$, corresponding to at least 1 correct out of 8), the metric asks whether the model can produce any correct answer, which is analogous to $\text{pass}@k$. At high values of $p$ (e.g., $p=0.875$), it asks whether the model is correct almost all of the time.\footnote{With $n\!=\!8$ samples, $\text{correct}\!\ge\!p$ is insensitive to changes at very low correctness probabilities (roughly $p\!\lesssim\!0.1$), since such cases frequently yield zero correct samples. Thus, availability gains in the extreme low-probability tail may be underestimated, while robustness at high $p$ remains well-estimated.}

\subsection{Ruling Out Response Diversity}
\label{app_sub:diversity}

\begin{figure*}[t]
    \centering
    \includegraphics[width=0.97\textwidth]{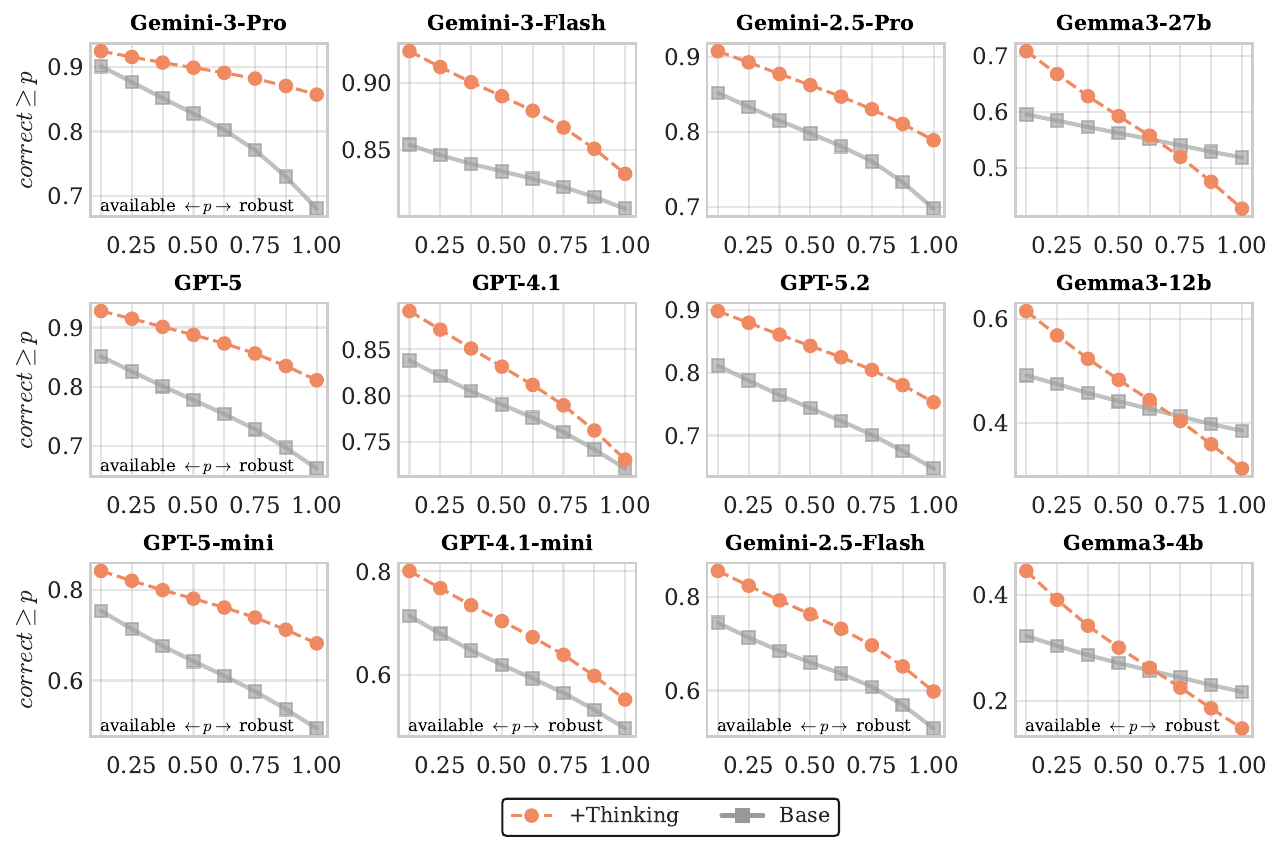}
    \caption{We report $correct \ge p$ for 12 LLMs, defined as the proportion of questions whose estimated probability of correctness across samples is at least $p$. Low values of $p$ reflect availability (the model produces at least one correct answer), while higher values reflect robustness (the model is correct consistently). Different LLMs exhibit distinct patterns; each column corresponds to a different pattern.}
    \label{fig:correctp_appendix}
\vspace{-0.8em}
\end{figure*}

Figure~\ref{fig:correctp_appendix} presents $\text{correct}\!\geq\!p$ for all 12 LLMs, comparing performance with and without thinking across the full range of thresholds.
At low thresholds ($p\!<\! 0.5$, corresponding to \textbf{Availability}), thinking improves $\text{correct}\!\geq\!p$ for all LLMs. This is consistent with all three mechanisms, including response diversity. At high thresholds ($p\!>\! 0.5$, corresponding to \textbf{Robustness}), the picture diverges across model families. For thinking-optimized LLMs (Gemini-3-Pro, Gemini-3-Flash, Gemini-2.5-Pro, GPT-5, GPT-5.2, GPT-5-mini), thinking also improves robustness: the $+$Thinking curve remains above the Base curve across the full range of $p$. This pattern is incompatible with response diversity as the sole explanation, since increased output variance cannot produce greater consistency in correctness.

For Gemma3 models (4B, 12B, 27B), the pattern reverses: thinking (elicited via CoT prompting) \emph{degrades} robustness, with the curves crossing at intermediate $p$. GPT-4.1 occupies a middle ground: CoT prompting yields modest gains in robustness. Our results suggest that, unlike thinking-optimized LLMs, CoT prompting mainly increases response diversity, and its effectiveness in enhancing inference or recall may depend on model capacity or training.

\subsection{Evidence for Recall Facilitation}
\label{app_sub:recall}

Having ruled out response diversity for thinking-optimized LLMs, we now ask whether the remaining gains reflect inference or recall facilitation. Figure~\ref{fig:known_given} reports, for each model, the fraction of facts that become known only with thinking, conditioned on encoding status.

The results show a strong dependence on encoding: for thinking-optimized LLMs, thinking recovers 40--65\% of of encoded but not directly known facts, compared to less than 20\% of non-encoded facts. This dependence is expected under recall facilitation; however, it does not, by itself, rule out inference. Nonetheless, the cumulative evidence favors recall facilitation as the primary mechanism by which thinking improves the factuality of thinking-optimized LLMs:
\begin{itemize}[nosep, leftmargin=*]
\item Thinking improves robustness, ruling out diversity (\S\ref{app_sub:diversity}).
\item Thinking gains concentrate mostly on encoded facts, consistent
with recall facilitation (Figure~\ref{fig:known_given}).
\item The facts in our benchmark are single-hop: complex deductive reasoning is unnecessary to derive the correct answer, making inference a less parsimonious explanation than recall of an encoded fact.
\end{itemize}

\subsection{The Role of Thinking-Specific Training}
\label{app_sub:training}

Our evaluation pools two distinct implementations of thinking: \emph{native thinking}, where the model was optimized during training to use thinking tokens (Gemini-3, Gemini-2.5, GPT-5 families), and \emph{CoT prompting}, where thinking is elicited at test time without dedicated training (Gemma3, GPT-4.1). The results suggest that these implementations yield qualitatively different effects. Native thinking consistently improves both availability and robustness, consistent with recall facilitation. CoT prompting produces mixed results: for Gemma3 models, it improves availability but degrades robustness, consistent with a response-diversity effect; for GPT-4.1, it yields modest gains in both.

This distinction has practical implications. The recall facilitation mechanism, in which thinking helps the model access encoded knowledge, may depend on thinking-specific training rather than merely on producing additional tokens at inference time. If so, the benefits of thinking for factuality are not a free lunch obtainable through prompting alone; they require dedicated training investment. We note, however, that our experimental design cannot fully disentangle the effects of thinking-specific training from other differences between model families (e.g., model size, data, post-training procedures). The pattern is suggestive but not conclusive.

\section{Additional Results}
\label{app_sec:additional_results}

In this section, we clarify how each analysis, figure, and table is computed, and report results for all evaluated LLMs to support the main-text trends.

\subsection{Profiles}
\label{app_sub:profiles}

\begin{figure}[t!]
    \centering
    \includegraphics[width=0.48\textwidth]{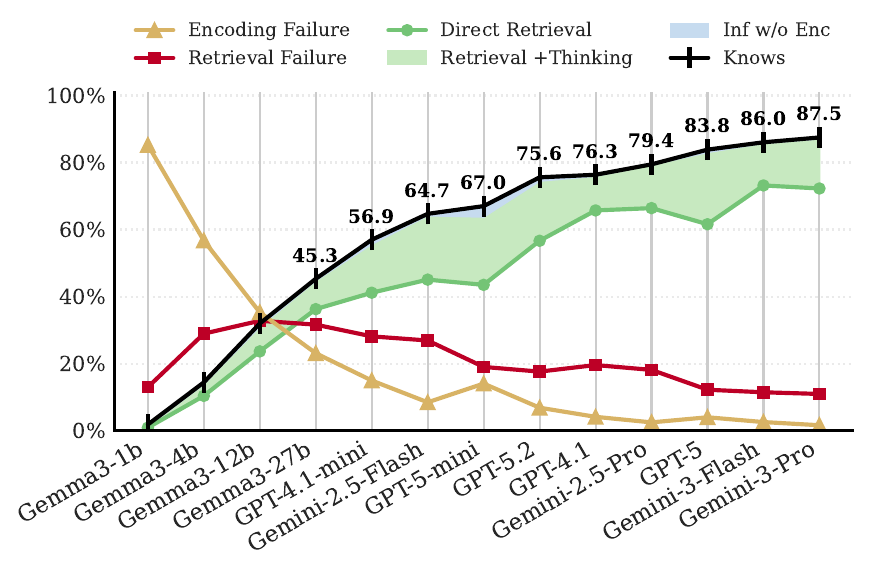}
    \vspace{-0.8em}
    \caption{\textbf{Trends Identified by Profiles:} `Knows' equal to the sum of `Direct Recall' (green line), `Recall with Thinking' (light green shade), and `Inference without Encoding' (light blue shade). LLMs are sorted according to `Knows'. Figure~\ref{fig:profiles_dist} presents distributions of the profiles.}
    \label{fig:profiles_line}
\vspace{-0.8em}
\end{figure}

Figure~\ref{fig:profiles_dist} presents the profile distributions for the 13 LLMs, and Figure~\ref{fig:profiles_line} provides a line plot that highlights trends. To estimate the distributions, we exclude facts with non-gradable questions (see Appendix~\ref{app_sub:nans}), then assign each fact a profile and compute the share of each profile.
The \emph{Knows} value sums the \emph{Direct recall}, \emph{recall with Thinking}, and \emph{Inference without Thinking} profiles. It therefore slightly overestimates the LLM's actual performance (with thinking enabled), since some facts become unknown when thinking is enabled. As such, \emph{Knows} can be viewed as an LLM's knowledge potential. 

\subsection{Fact Popularity}
\label{app_sub:fact_pop}

Popularity tiers (bottom 20\%, top 20\%) are defined based on Wikipedia page visit counts for the pages from which facts are extracted. This serves as a reasonable but imperfect proxy for fact incidence, and we expect the observed trends to be even stronger with a more direct measure. Figure~\ref{fig:popularity_appendix} compares the encoding and recall gaps between the two popularity tiers across all 13 LLMs. Encoding performance is measured as the proportion of encoded facts within each tier, while recall performance is measured as the proportion of known facts conditioned on being encoded. This conditioning isolates recall from encoding, allowing us to assess whether rare facts are harder to recall even when they are encoded. As shown, for most LLMs (10 out of 13; excluding Gemma3 12b, 4b, and 1b), the recall gap is substantially larger than the encoding gap.

\subsection{Phrasing}
\label{app_sub:phrasing}

A large body of work suggests that LLMs are sensitive to prompt phrasing \citep{Deng2023Rephrase, LinGOXLY024, MizrahiKMDSS24, HabbaAIPBCSS25, Lior2025reliable}. In contrast, in \S\ref{sub:recall_fails}, we find that LLMs are insensitive to question rephrasing in our setup. Here we elaborate on this analysis. We conduct hypothesis tests comparing the distributions of performance across phrasings, treating success as a binary variable (1 if the question grade exceeds 0.5, 0 otherwise). In total, we perform 104 tests, covering 13 LLMs (with and without thinking) and four question pairs (direct vs. direct natural, reverse vs. reverse natural, and their multiple-choice variants). After applying False Discovery Rate (FDR) correction using the Benjamini–Yekutieli procedure with $q=0.05$ \citep{benjamini2001control}, none of the tests are significant. This indicates that, in our setup, phrasing does not confound knowledge assessment.

\begin{figure*}[t]
    \centering
    \includegraphics[width=0.48\textwidth]{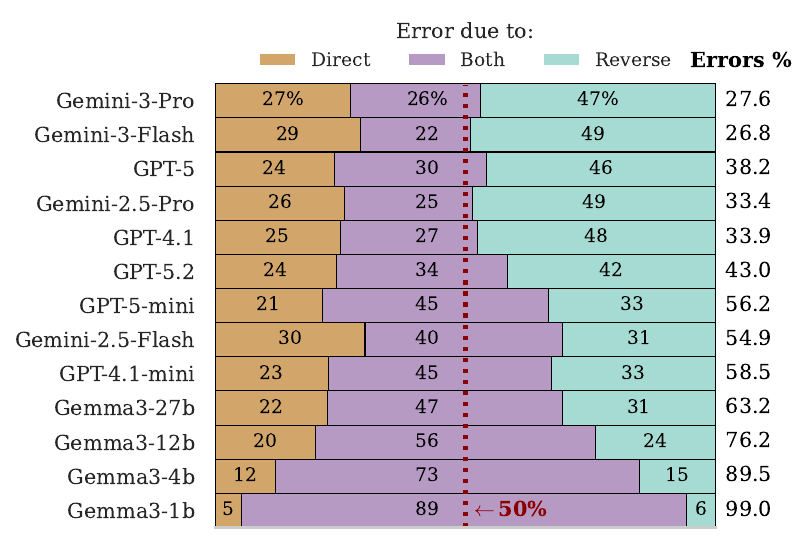}
    \includegraphics[width=0.48\textwidth]{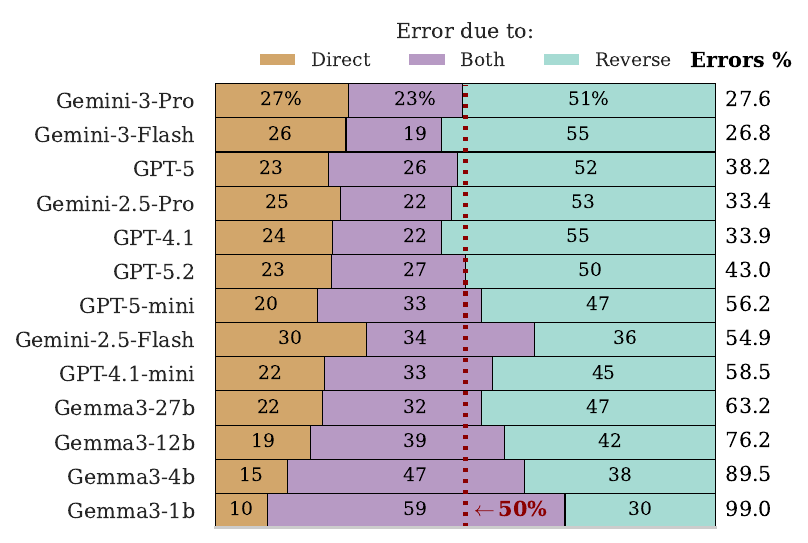}
    \caption{We decompose errors into three categories: failures on only direct questions, only reverse questions, or both. The dashed red line marks 50\%. The left bar plot shows results for LLMs without thinking on all facts, while the right bar plot shows results for encoded facts. As shown, errors from reverse questions exceed those from direct questions, regardless of encoding.}
    \label{fig:dir_rev_errors}
\vspace{-0.8em}
\end{figure*}

\subsection{Reverse Questions}
\label{app_sub:dir_vs_rev}

\begin{figure*}[t]
    \centering
    \includegraphics[width=0.8\textwidth]{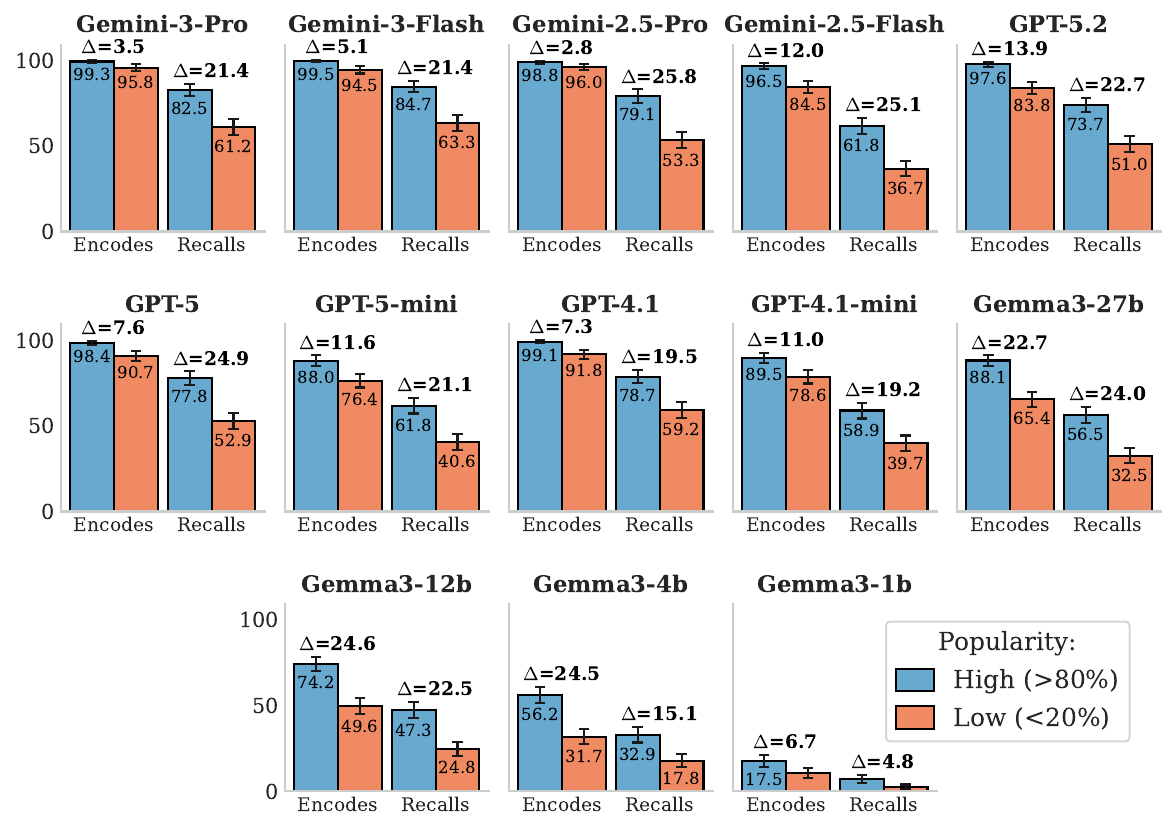}
    \includegraphics[width=0.8\textwidth]{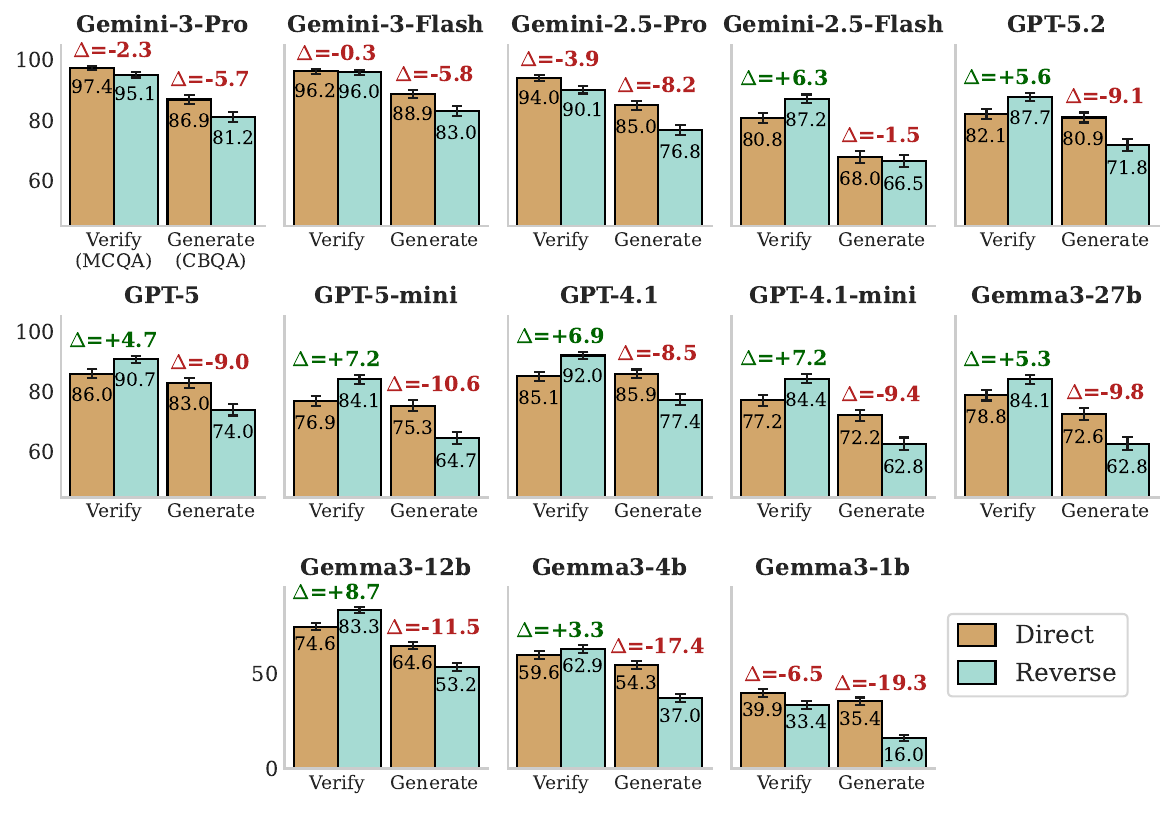}
    \caption{\emph{Top:} We compare two popularity tiers (bottom 20\% vs. top 20\%) in terms of encoding rates and direct recall rates (knowing encoded facts without thinking). The $\Delta$ indicates the gap between tiers. \emph{Bottom:} We compare direct recall rates on direct and reverse questions across two tasks: verification (multiple-choice) and generation (closed-book). The $\Delta$ denotes the gap between the directions.}
    \label{fig:popularity_appendix}
\vspace{-0.8em}
\end{figure*}

Figure~\ref{fig:popularity_appendix} reports recall performance conditioned on facts being encoded, for both multiple-choice and closed-book questions, under direct and reverse directions, across 13 LLMs. Each bar shows the fraction of encoded facts that are known under a given direction and task format. A fact is counted as known under the direct (respectively, reverse) condition only if the grades for both corresponding questions (direct and direct natural; reverse and reverse natural) exceed $g(q) > 0.5$. The same criterion is applied to the multiple-choice variants. Figure~\ref{fig:popularity_appendix} shows that, for all LLMs, the gap between direct and reverse questions is much smaller for verification than for generation. For 9 out of 13 LLMs, reverse verification is in fact easier.

Our analysis comparing direct and reverse questions focuses on encoded facts, as this allows us to isolate recall failures. One might argue that this conditioning introduces selection bias, since the encoding tasks share the same answer entity as direct questions. We therefore conduct several robustness analyses. First, all key results replicate when we do not condition on encoding: reverse generation remains substantially harder than direct generation, the gap is much smaller or absent for verification (with reverse verification often easier), and thinking mitigates the gap. Second, we control for answer type (e.g., person, location, etc) by computing performance separately for each type and then comparing reverse to direct questions; the same trends persist. 

Finally, we analyze error attribution over all not-known facts and over those that are encoded. We decompose the sources of error into three categories: failure on direct questions only, failure on reverse questions only, or failure on both. As shown in Figure~\ref{fig:dir_rev_errors}, across all LLMs, errors from reverse questions consistently exceed those from direct questions. For smaller LLMs, the dominant error source is failing `both' questions. As scale increases, the share of `both' decreases, and `only reverse' becomes the dominant failure mode (40-50\% are `only reverse' and an additional 15-30\% are `both').

\subsection{Thinking}
\label{app_sub:thinking}

\begin{figure*}[t]
    \centering
    \includegraphics[width=0.48\textwidth]{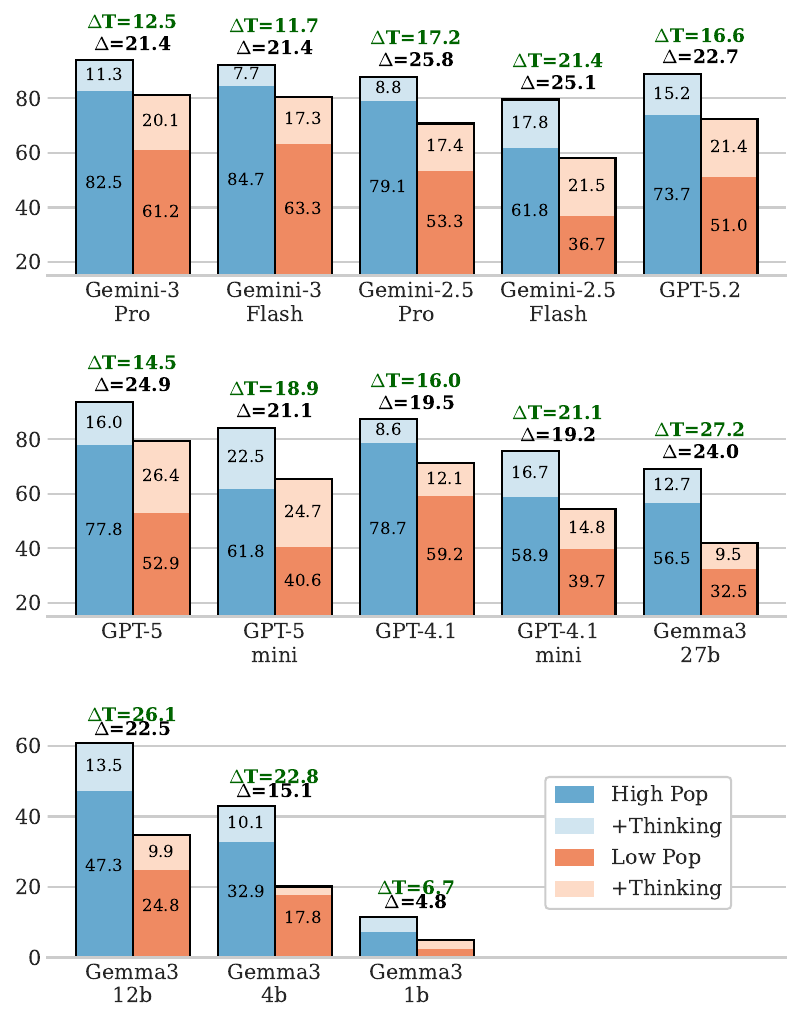}
    \includegraphics[width=0.48\textwidth]{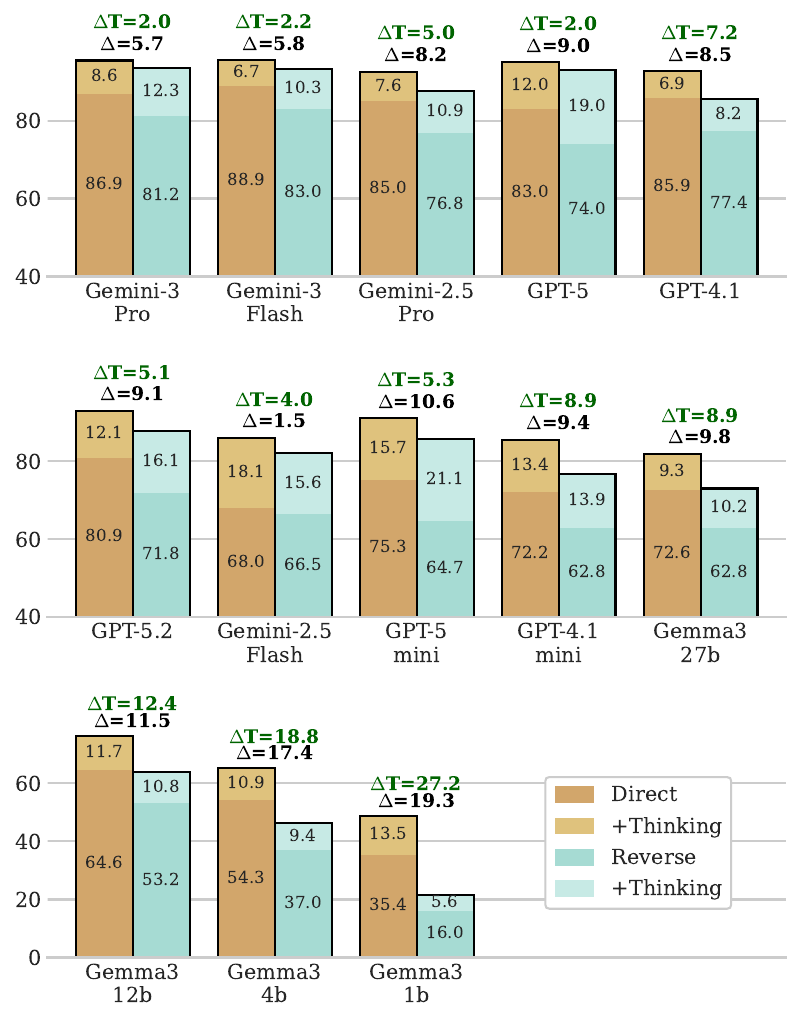}
    \caption{We examine the impact of thinking on retrieval (knowing encoded facts). On the \emph{left}, we compare two popularity tiers (bottom 20\% vs. top 20\%), while on the \emph{right}, we compare direct and reverse questions. The popularity or directional gaps are denoted by $\Delta$ (no thinking) and \textcolor{OliveGreen}{$\Delta T$} (with thinking).}
    \label{fig:dir_pop_thinking_appendix}
\vspace{-0.8em}
\end{figure*}

Figure~\ref{fig:dir_pop_thinking_appendix} examines the impact of inference-time thinking on recall across 13 LLMs, focusing on popularity and directionality, respectively. In both analyses, we condition on encoded facts and decompose recall into baseline performance (facts known without thinking, given they are encoded) and the additional share that becomes known only with thinking. As shown, thinking mitigates the popularity gap for 8 out of 13 LLMs (excluding GPT-4.1-mini and the Gemma3 models) and the directional gap for 9 out of 13 LLMs (excluding Gemini-2.5-Flash and Gemma3 models). In both cases, the mitigation is substantial for frontier LLMs, that is, the stronger proprietary models.

\section{Methodological and Design Choices}
\label{app_sec:methodological}

The goal of this section is to describe and justify methodological details and design choices that influence our evaluation and analysis.

\begin{table*}[t]
\centering
\caption{Profile distributions under three strategies for handling PARTIALLY and OTHER responses. The three rightmost columns report the share of `Excluded' facts, decomposed into `OTHER or PARTIALLY' responses and `Direct Inference' cases. The columns `Knows (+Think)' and `Knows' are computed with `Direct Inference'.}
\normalsize
\begin{adjustbox}{width=0.975\textwidth}
\begin{tabular}{l|rrr|rrrrr|rrr}
\toprule
\rowcolor{gray!20} \multicolumn{12}{c}{\textbf{PARTIALLY:} Assign NaN. \textbf{OTHER:} Assign NaN. \textbf{NaNs Strategy:} Ignore the entire fact.} \\
\midrule
\multirow{2}{*}{\textbf{LLM}} & \multirow{2}{*}{\textbf{Encodes}} & \multirow{2}{*}{\makecell{\textbf{Knows} \\ (+Think)}} & \multirow{2}{*}{\textbf{Knows}} & \multirow{2}{*}{\makecell{\textbf{Encoding} \\ \textbf{Failure}}} & \multirow{2}{*}{\makecell{\textbf{Recall} \\ \textbf{Failure}}} & \multirow{2}{*}{\makecell{\textbf{Direct} \\ \textbf{Recall}}} & \multirow{2}{*}{\makecell{\textbf{Recall w/} \\ \textbf{Thinking}}} & \multirow{2}{*}{\makecell{\textbf{Inference w/o} \\ \textbf{Encoding}}} & \multirow{2}{*}{\textbf{Excluded}} & \multirow{2}{*}{\makecell{\textbf{OTHER} or \\ \textbf{PARTIALLY}}} & \multirow{2}{*}{\makecell{\textbf{Direct} \\ \textbf{Inference}}} \\
 & & & & & & & & & & & \\
\midrule
Gemini-3-Pro & 98.3 & 87.4 & 72.4 & 1.5 & 11.2 & 72.3 & 14.9 & 0.1 & 5.2 & 5.1 & 0.1 \\
Gemini-3-Flash & 98.1 & 86.6 & 73.4 & 1.8 & 11.7 & 73.3 & 13.1 & 0.1 & 9.5 & 9.5 & 0.0 \\
Gemini-2.5-Pro & 98.1 & 79.7 & 66.1 & 1.9 & 18.5 & 66.0 & 13.5 & 0.1 & 7.4 & 7.3 & 0.1 \\
Gemini-2.5-Flash & 92.8 & 65.8 & 46.0 & 6.9 & 27.4 & 46.0 & 19.5 & 0.3 & 13.3 & 13.2 & 0.0 \\
GPT-5.2 & 92.8 & 75.7 & 56.8 & 6.3 & 18.1 & 56.6 & 18.0 & 0.9 & 8.2 & 8.0 & 0.2 \\
GPT-5 & 95.9 & 84.5 & 62.0 & 3.3 & 12.4 & 61.8 & 21.8 & 0.7 & 6.8 & 6.6 & 0.2 \\
GPT-5-mini & 83.5 & 67.9 & 44.2 & 13.4 & 19.0 & 44.0 & 20.6 & 3.1 & 7.2 & 7.0 & 0.2 \\
GPT-4.1 & 96.0 & 77.3 & 64.9 & 3.7 & 19.1 & 64.9 & 12.0 & 0.4 & 9.2 & 9.1 & 0.1 \\
GPT-4.1-mini & 85.7 & 58.1 & 41.7 & 13.4 & 28.7 & 41.5 & 15.4 & 0.9 & 11.1 & 10.9 & 0.2 \\
Gemma3-27b & 80.1 & 47.3 & 37.3 & 19.5 & 33.4 & 37.0 & 9.7 & 0.4 & 21.0 & 20.7 & 0.3 \\
Gemma3-12b & 68.0 & 33.4 & 24.6 & 31.1 & 35.6 & 24.5 & 7.9 & 1.0 & 27.2 & 27.1 & 0.1 \\
Gemma3-4b & 46.3 & 16.0 & 11.0 & 53.5 & 30.6 & 10.9 & 4.9 & 0.2 & 27.2 & 27.1 & 0.1 \\
Gemma3-1b & 17.0 & 1.8 & 1.1 & 82.7 & 15.5 & 1.1 & 0.4 & 0.3 & 45.8 & 45.8 & 0.0 \\
\bottomrule
\multicolumn{12}{c}{} \\
\multicolumn{12}{c}{} \\
\toprule
\rowcolor{Goldenrod!60} \multicolumn{12}{c}{**Selected** \textbf{PARTIALLY:} Assign NaN. \textbf{OTHER:} Assign NaN. \textbf{NaNs Strategy:} Ignore only if both tasks are NaNs. **Selected**} \\
\midrule
\multirow{2}{*}{\textbf{LLM}} & \multirow{2}{*}{\textbf{Encodes}} & \multirow{2}{*}{\makecell{\textbf{Knows} \\ (+Think)}} & \multirow{2}{*}{\textbf{Knows}} & \multirow{2}{*}{\makecell{\textbf{Encoding} \\ \textbf{Failure}}} & \multirow{2}{*}{\makecell{\textbf{Recall} \\ \textbf{Failure}}} & \multirow{2}{*}{\makecell{\textbf{Direct} \\ \textbf{Recall}}} & \multirow{2}{*}{\makecell{\textbf{Recall w/} \\ \textbf{Thinking}}} & \multirow{2}{*}{\makecell{\textbf{Inference w/o} \\ \textbf{Encoding}}} & \multirow{2}{*}{\textbf{Excluded}} & \multirow{2}{*}{\makecell{\textbf{OTHER} or \\ \textbf{PARTIALLY}}} & \multirow{2}{*}{\makecell{\textbf{Direct} \\ \textbf{Inference}}} \\
 & & & & & & & & & & & \\
\midrule
Gemini-3-Pro & 98.1 & 87.7 & 72.4 & 1.6 & 10.9 & 72.2 & 14.9 & 0.3 & 1.3 & 1.1 & 0.2 \\
Gemini-3-Flash & 97.2 & 86.1 & 73.2 & 2.5 & 11.4 & 73.2 & 12.6 & 0.2 & 3.2 & 3.1 & 0.1 \\
Gemini-2.5-Pro & 97.5 & 79.6 & 66.6 & 2.5 & 18.1 & 66.4 & 13.0 & 0.1 & 1.3 & 1.2 & 0.2 \\
Gemini-2.5-Flash & 91.1 & 64.8 & 45.1 & 8.4 & 26.9 & 45.0 & 19.2 & 0.5 & 1.9 & 1.8 & 0.1 \\
GPT-5.2 & 92.1 & 75.9 & 57.0 & 6.8 & 17.6 & 56.6 & 17.9 & 1.0 & 2.0 & 1.6 & 0.3 \\
GPT-5 & 95.3 & 84.1 & 61.8 & 4.0 & 12.2 & 61.6 & 21.5 & 0.8 & 1.6 & 1.3 & 0.2 \\
GPT-5-mini & 82.8 & 67.3 & 43.8 & 14.0 & 19.0 & 43.5 & 20.3 & 3.2 & 1.8 & 1.5 & 0.3 \\
GPT-4.1 & 95.6 & 76.7 & 66.1 & 4.1 & 19.6 & 65.7 & 10.3 & 0.3 & 2.5 & 2.1 & 0.4 \\
GPT-4.1-mini & 83.8 & 57.3 & 41.5 & 14.9 & 28.1 & 41.1 & 14.6 & 1.2 & 2.6 & 2.2 & 0.3 \\
Gemma3-27b & 76.4 & 45.8 & 36.8 & 23.1 & 31.6 & 36.2 & 8.6 & 0.5 & 3.5 & 3.0 & 0.5 \\
Gemma3-12b & 63.8 & 32.1 & 23.8 & 35.3 & 32.8 & 23.6 & 7.4 & 0.9 & 3.8 & 3.6 & 0.1 \\
Gemma3-4b & 42.9 & 14.6 & 10.5 & 56.7 & 28.9 & 10.4 & 3.6 & 0.4 & 5.3 & 5.1 & 0.2 \\
Gemma3-1b & 14.3 & 1.9 & 1.0 & 85.2 & 13.0 & 0.9 & 0.5 & 0.5 & 10.0 & 9.8 & 0.1 \\
\bottomrule
\multicolumn{12}{c}{} \\
\multicolumn{12}{c}{} \\
\toprule
\rowcolor{gray!20} \multicolumn{12}{c}{\textbf{PARTIALLY:} 0.51 Weight. \textbf{OTHER:} Assign NaN. \textbf{NaNs Strategy:} Ignore only if both tasks are NaNs.} \\
\midrule
\multirow{2}{*}{\textbf{LLM}} & \multirow{2}{*}{\textbf{Encodes}} & \multirow{2}{*}{\makecell{\textbf{Knows} \\ (+Think)}} & \multirow{2}{*}{\textbf{Knows}} & \multirow{2}{*}{\makecell{\textbf{Encoding} \\ \textbf{Failure}}} & \multirow{2}{*}{\makecell{\textbf{Recall} \\ \textbf{Failure}}} & \multirow{2}{*}{\makecell{\textbf{Direct} \\ \textbf{Recall}}} & \multirow{2}{*}{\makecell{\textbf{Recall w/} \\ \textbf{Thinking}}} & \multirow{2}{*}{\makecell{\textbf{Inference w/o} \\ \textbf{Encoding}}} & \multirow{2}{*}{\textbf{Excluded}} & \multirow{2}{*}{\makecell{\textbf{OTHER} or \\ \textbf{PARTIALLY}}} & \multirow{2}{*}{\makecell{\textbf{Direct} \\ \textbf{Inference}}} \\
 & & & & & & & & & & & \\
\midrule
Gemini-3-Pro & 98.2 & 88.7 & 74.3 & 1.5 & 9.8 & 74.2 & 14.2 & 0.2 & 0.2 & 0.2 & 0.0 \\
Gemini-3-Flash & 97.5 & 87.0 & 75.4 & 2.3 & 10.9 & 75.2 & 11.4 & 0.2 & 0.5 & 0.3 & 0.2 \\
Gemini-2.5-Pro & 97.7 & 81.0 & 68.5 & 2.2 & 16.9 & 68.3 & 12.4 & 0.1 & 0.1 & 0.0 & 0.1 \\
Gemini-2.5-Flash & 92.3 & 67.1 & 47.2 & 7.3 & 25.8 & 47.0 & 19.5 & 0.5 & 0.8 & 0.6 & 0.2 \\
GPT-5.2 & 92.6 & 76.3 & 57.8 & 6.4 & 17.5 & 57.6 & 17.5 & 1.0 & 0.7 & 0.6 & 0.2 \\
GPT-5 & 95.7 & 85.3 & 63.4 & 3.5 & 11.3 & 63.2 & 21.1 & 0.7 & 0.4 & 0.1 & 0.2 \\
GPT-5-mini & 84.1 & 68.7 & 45.1 & 12.9 & 18.8 & 44.7 & 20.6 & 3.0 & 0.8 & 0.4 & 0.4 \\
GPT-4.1 & 96.1 & 77.5 & 67.2 & 3.7 & 19.1 & 66.9 & 10.0 & 0.2 & 0.5 & 0.2 & 0.3 \\
GPT-4.1-mini & 85.2 & 58.4 & 42.7 & 13.5 & 28.5 & 42.4 & 14.4 & 1.3 & 1.0 & 0.6 & 0.4 \\
Gemma3-27b & 78.8 & 47.2 & 38.6 & 20.8 & 32.7 & 37.9 & 8.1 & 0.4 & 1.8 & 1.2 & 0.7 \\
Gemma3-12b & 66.5 & 33.5 & 25.7 & 32.7 & 34.1 & 25.4 & 7.1 & 0.8 & 1.7 & 1.4 & 0.3 \\
Gemma3-4b & 48.3 & 15.6 & 11.4 & 51.5 & 33.2 & 11.1 & 4.0 & 0.2 & 3.2 & 2.8 & 0.3 \\
Gemma3-1b & 21.6 & 2.5 & 1.4 & 78.0 & 19.7 & 1.3 & 0.6 & 0.4 & 7.9 & 7.7 & 0.2 \\
\bottomrule
\end{tabular}
\end{adjustbox}
\label{tab:nan_strategy}
\vspace{-0.8em}
\end{table*}

\subsection{Handling Partially and Other Grades}
\label{app_sub:nans}

The question grade is computed as accuracy over CORRECT and INCORRECT responses only, excluding responses labeled PARTIALLY or OTHER. For example, if a question receives 3 CORRECT, 1 INCORRECT, 2 PARTIALLY, and 2 OTHER responses, its grade is $\tfrac{3}{3+1}=0.75$. We exclude PARTIALLY and OTHER from grading because assigning them weights is subjective \citep{Kalai25why}.

We next describe how we handle PARTIALLY and OTHER responses at the fact level. We evaluated three strategies; Table~\ref{tab:nan_strategy} reports the resulting profile distributions and the fraction of excluded facts.

The first strategy excludes a fact if any of its six profiling questions is non-gradable, meaning all responses are PARTIALLY or OTHER. This approach removes a substantial number of facts, around 10\% for frontier LLMs and over 20\% for Gemma3 models.

The second strategy groups questions into task pairs (encoding, direct, and reverse) and excludes a fact only if an entire pair is non-gradable. This substantially reduces exclusions to 1–4\% for nearly all LLMs, while producing profile distributions nearly identical to the first strategy.

The third strategy extends the second by assigning PARTIALLY responses a weight of 0.51, allowing facts to be considered known even when all responses are PARTIALLY. While this further reduces exclusions to below 1\% and yields results very similar to the second strategy, it introduces subjective judgment that is harder to justify. We therefore adopt the second strategy throughout the paper.

Finally, in \S\ref{sub:profiles}, we do not define a profile for facts that are not encoded but are known without thinking. Such cases are rare (less than 0.5\% across all models; see the ``Direct Inference'' column in Table~\ref{tab:nan_strategy}) and are likely due to sampling noise or grader error; excluding this profile does not change the results.

\subsection{Threshold Selection}
\label{app_sub:thresholds}

Recall that both our definitions of \emph{encodes} and \emph{knows} rely on a threshold $g(q) > \tau$. A natural choice is $\tau = 0.5$, which corresponds to correctness being more likely than incorrectness. However, the threshold is a design choice, and different values can yield different profile distributions.

\begin{figure*}[t!]
    \centering
    \includegraphics[width=0.98\textwidth]{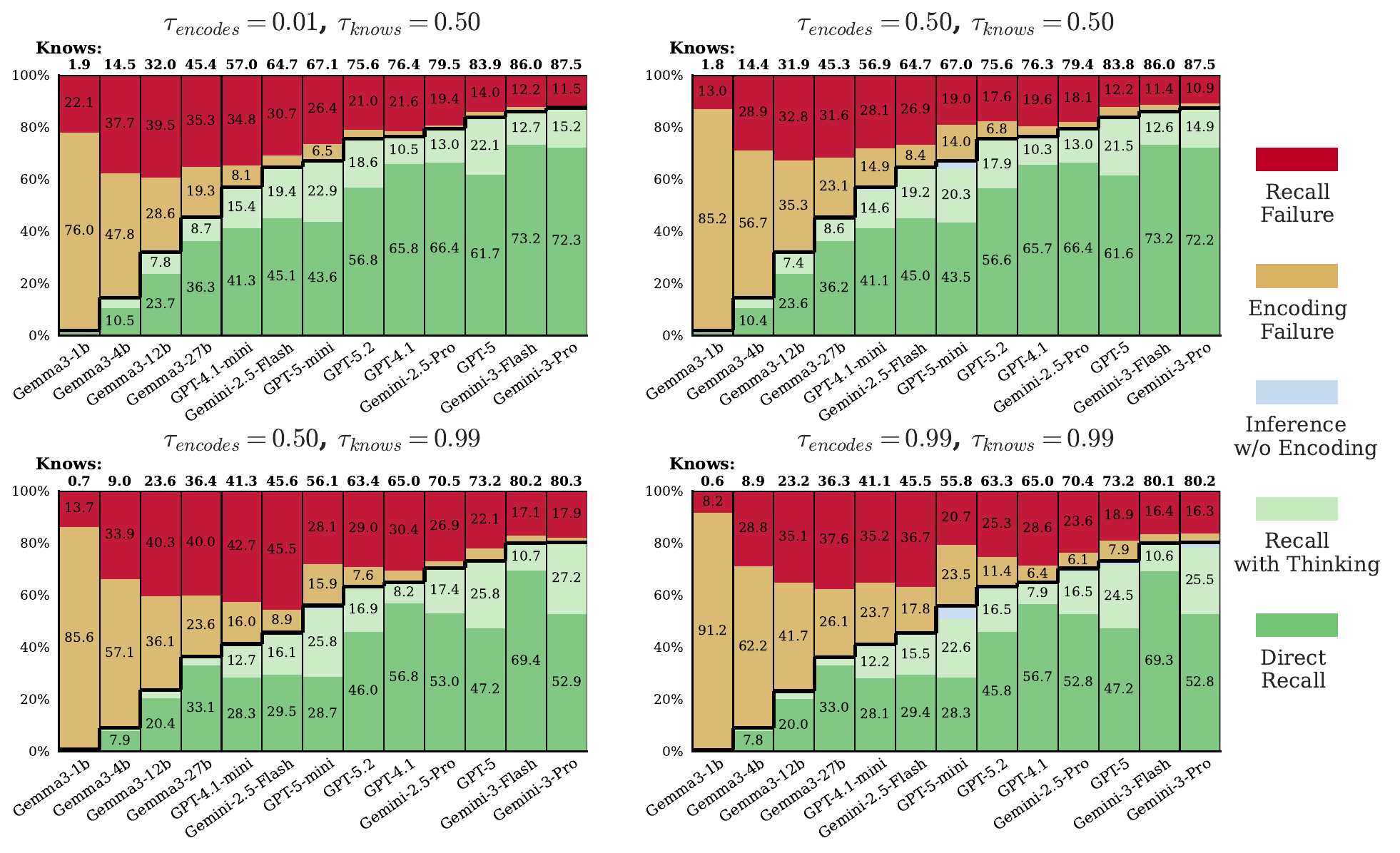}
    \vspace{-0.8em}
    \caption{Distributions (percentages) of knowledge profiles across 13 LLMs under different $g(q) > \tau$ thresholds for encoding and knowledge. While the exact profile percentages vary with $\tau$, our main findings are robust to the choice of threshold.}
    \label{fig:thresholds}
\vspace{-0.8em}
\end{figure*}

To assess sensitivity to this choice, Figure~\ref{fig:thresholds} presents profile distributions for five additional reasonable values of $\tau$. While the exact proportions of profiles vary with $\tau$, our main conclusions are robust. Specifically, (1) encoding is nearly saturated for frontier LLMs, even under a conservative threshold of $\tau = 0.99$; (2) recall failures account for a substantial fraction of errors, and their relative impact increases under stricter thresholds; and (3) thinking consistently acts as a recovery mechanism, with over 10\% of facts known only with thinking across all threshold choices.

\begin{figure}[!t]
    \centering
    \includegraphics[width=0.4\textwidth]{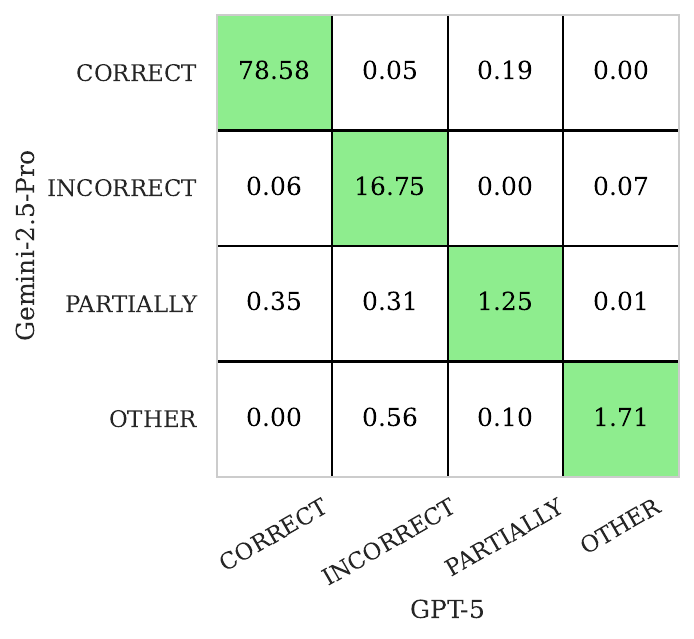}
    \caption{Confusion matrix between graders based on Gemini-2.5-Pro and GPT-5. Entries are percentages. Overall agreement is 98.2\%, with most disagreements occurring when Gemini-2.5-Pro predicts OTHER or PARTIALLY.}
    \label{fig:grader_cmax}
\vspace{-0.8em}
\end{figure}

\subsection{Grader Evaluation}
\label{app_sub:grader}

Our evaluation relies on a prompted LLM grader, a common practice in short-form factuality evaluation \citep{simpleqa, simpleqa_ver}. We use Gemini-2.5-Pro with thinking as the primary grader. The prompt is adapted from SimpleQA \citep{simpleqa} and refined through manual error analysis with Gemini-2.5-Pro. In this subsection, we assess robustness to the choice of the grader's underlying LLM by comparing our grader to a GPT-5-based grader with thinking (medium effort). We uniformly sample 4,160 responses across six tasks and all evaluated LLMs. The two graders agree on 98.2\% of responses.

Figure~\ref{fig:grader_cmax} shows the confusion matrix between the two graders. Most disagreements occur when Gemini-2.5-Pro assigns OTHER or PARTIALLY, while GPT-5 assigns a definitive label. To better understand these disagreements, we manually examine them; representative examples are shown in Box~\ref{box:disagreements_analysis}.
In most cases, Gemini-2.5-Pro yields the more accurate label, which is unsurprising given that the prompt is optimized for it. In addition, it occasionally draws on external knowledge. 

Overall, since the graders agree on the vast majority of responses and the remaining disagreements primarily involve OTHER or PARTIALLY labels, which are excluded from our analysis, we conclude that our results are not sensitive to the choice of the grader's underlying LLM.

\subsection{Multiple Response Generation}
\label{app_sub:n_sensitivity}

For each task, we generate $n\!=\!8$ responses per question for three reasons. First, our goal is to assess factual knowledge at the level of facts; a single response is a noisy estimator and may misclassify a fact as not encoded or not known due to sampling variance. Second, because our questions are not single-token predictions, probability-based estimates are not well defined, and token-level likelihoods are inaccessible for most proprietary LLMs, making repeated sampling a natural alternative. Finally, multiple responses are required to distinguish availability (at least one correct response) from robustness (all responses correct), which is part of our analysis of inference-time thinking.

\begin{figure*}[t!]
    \centering
    \includegraphics[width=0.95\textwidth]{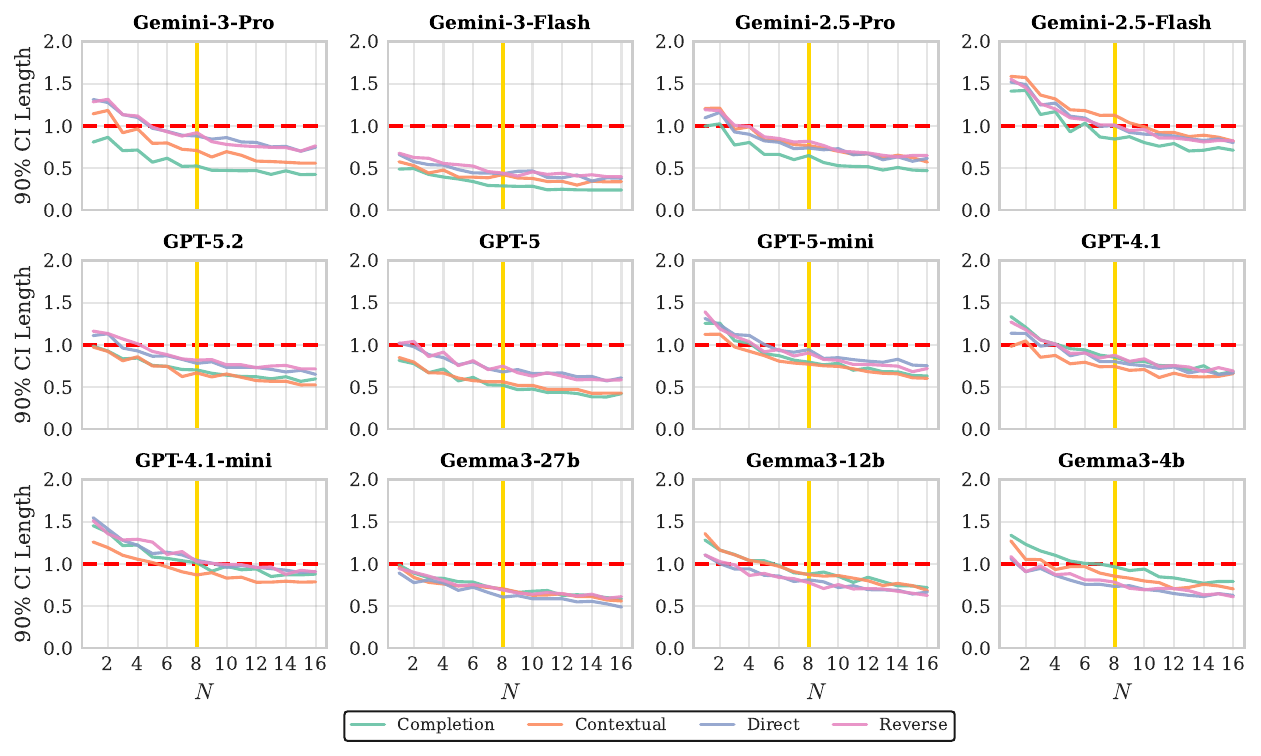}
    \vspace{-0.8em}
    \caption{Sensitivity of the fraction of facts with $g(q)\!>\!0.5$ to the number of sampled responses $n$, shown as the width of the 90\% empirical bootstrap confidence interval (95th–5th percentiles) across eight LLMs. As shown, the interval width drops below 1\% at $n\!=\!8$.}
    \label{fig:n_sensitivity}
\vspace{-0.8em}
\end{figure*}

To assess the effect of multiple sampling and justify our choice of $n\!=\!8$, we analyze how the fraction of facts with $g(q)\!>\!0.5$ varies with $n$ across four tasks: two encoding tasks (completion and contextual) and two knowledge tasks (direct and reverse). For each task and model, we perform 1,000 bootstrap resamples, each time subsampling $n$ responses per question and recomputing the fraction of facts with $g(q)\!>\!0.5$. Figure~\ref{fig:n_sensitivity} shows the width of the 90\% empirical bootstrap confidence interval (95th--5th percentiles) as a function of $n$. Across the evaluated models, the interval width falls below 1\% for all tasks at $n\!=\!8$, indicating that the estimated fraction is stable to within $\pm$0.5\%. We therefore use $n\!=\!8$ throughout the paper.
\onecolumn
\begin{prompt}[label={box:disagreements_analysis}]{Gray}{Analysis of Disagreements Between Graders}

\textbf{Direct Question:} Who developed the circuit for the Cockcroft–Walton generator in 1919? \\
\textbf{Gold Answer:} Heinrich Greinacher \\
\textbf{Prediction of Gemini-3-Pro:} The circuit itself was essentially discovered by Heinrich Greinacher, a Swiss physicist, in 1919. \\
\textbf{Better Grader:} Gemini (CORRECT vs GPT-5 INCORRECT) \\
\textbf{Rationale:} The prediction correctly identifies Greinacher; GPT-5 likely penalized the shift from ``developed" to ``discovered''.

\tcbline

\textbf{Direct Natural Question:} About how many iPhone 11s had sold as of March 2022? \\
\textbf{Gold Answer:} 159.2 million \\
\textbf{Prediction of Gemini-2.5-Pro+Thinking:} Approximately 160 million units. \\
\textbf{Better Grader:} Gemini (CORRECT vs GPT-5 INCORRECT) \\
\textbf{Rationale:} Rounding 159.2 to ``Approximately 160'' is a permissible and align with the guidelines.

\tcbline

\textbf{Reverse Question:} What district within the Lahore Division of Punjab, Pakistan has a total area of 3,995 square kilometres? \\
\textbf{Gold Answer:} Kasur District \\
\textbf{Prediction of Gemini-2.5-Pro+Thinking:} Kasur \\
\textbf{Better Grader:} Gemini (CORRECT vs GPT-5 PARTIALLY) \\
\textbf{Rationale:} Dropping ``District'' is explicitly allowed by the guidelines.

\tcbline

\textbf{Reverse Question:} What is the name of the female version of Taskmaster who appears in Black Widow? \\
\textbf{Gold Answer:} Antonia Dreykov \\
\textbf{Prediction of Gemini-2.5-Flash:} Taskmaster \\
\textbf{Better Grader:} Gemini (OTHER vs GPT-5 INCORRECT) \\
\textbf{Rationale:} The prediction does not contradict the question (she is Taskmaster).

\tcbline

\textbf{Reverse Question:} Whose activities did the Federal Bureau of Investigation (FBI) program COINTELPRO attempt to subvert by placing a counterintelligence operative in the local Panthers organization? \\
\textbf{Gold Answer:} Fredrick Allen Hampton \\
\textbf{Prediction of GPT-5-mini+Thinking:} The Black Panther Party \\
\textbf{Better Grader:} Gemini (PARTIALLY vs GPT-5 INCORRECT) \\
\textbf{Rationale:} Identifying the group (Black Panthers) instead of the leader (Hampton) is a ``more general version'' of the target.

\tcbline

\textbf{Reverse Natural Question:} Who watched 'The UFO Incident' on the night Mike Rogers admitted the job was seriously behind schedule? \\
\textbf{Gold Answer:} Walton and Rogers \\
\textbf{Prediction of GPT-4.1+Thinking:} Travis Walton and the other logging crew members (including Mike Rogers) watched 'The UFO Incident' together that night. \\
\textbf{Better Grader:} Gemini (CORRECT vs GPT-5 INCORRECT) \\
\textbf{Rationale:} Including the rest of the crew does not contradict the presence of Walton and Rogers.

\tcbline

\textbf{Direct Question:} What is the compound wall that surrounds the St. Antony's Syro-Malabar Catholic Forane Church at Ollur called? \\
\textbf{Gold Answer:} Elephant Wall \\
\textbf{Prediction of Gemma3-4b+Thinking:} Perimeter Wall \\
\textbf{Better Grader:} Gemini (OTHER vs GPT-5 INCORRECT) \\
\textbf{Rationale:} ``Perimeter Wall'' is missing the specific name, but it is not a factual contradiction (it \textit{is} a perimeter wall).

\tcbline

\textbf{Reverse Question:} What statue depicting Vairocana Buddha is located within the Fodushan Scenic Area? \\
\textbf{Gold Answer:} Spring Temple Buddha \\
\textbf{Prediction of Gemma3-1b+Thinking:} Fodushan Scenic Area, specifically the Temple of the White Horse, where the statue of Vairocana Buddha is located. \\
\textbf{Better Grader:} Gemini (INCORRECT vs GPT-5 OTHER) \\
\textbf{Rationale:} The prediction places the statue in a completely different temple (White Horse).

\tcbline

\textbf{Direct Question:} Who built the town, originally called Beodericsworth, on a grid pattern? \\
\textbf{Gold Answer:} Abbot Baldwin \\
\textbf{Prediction of Gemma3-1b+Thinking:} The people of Beodericsworth built the town. \\
\textbf{Better Grader:} Gemini (OTHER vs GPT-5 INCORRECT) \\
\textbf{Rationale:} ``The people built it'' is a vague truism that fails to answer the ``Who'' (Architect/Leader) but doesn't contradict the answer.

\tcbline

\textbf{Direct Natural Question:} Which institution was the fossil specimen "Apex" loaned to in December 2024? \\
\textbf{Gold Answer:} American Museum of Natural History \\
\textbf{Prediction of Gemini-2.5-Pro:} According to the provided information, the fossil specimen "Apex" was loaned in December 2022, not December 2024. Therefore, I cannot answer your question. \\
\textbf{Better Grader:} Gemini (INCORRECT vs GPT-5 OTHER) \\
\textbf{Rationale:} The prediction explicitly contradicts the question (claiming it is wrong).

\tcbline

\textbf{Direct Question:} What existing highway was US 1 mostly assigned to when the road system was laid out? \\
\textbf{Gold Answer:} Atlantic Highway \\
\textbf{Prediction of Gemma3-1b+Thinking:} US 1 was primarily assigned to transport goods. \\
\textbf{Better Grader:} Gemini (OTHER vs GPT-5 INCORRECT) \\
\textbf{Rationale:} Stating the purpose (``transport goods'') instead of the highway name is irrelevant (Other).

\tcbline

\textbf{Reverse Natural Question:} What's the name of the Hindu temple complex in Delhi that opened on November 6, 2005? \\
\textbf{Gold Answer:} Swaminarayan Akshardham \\
\textbf{Prediction of Gemini-2.5-Pro:} Akshardham \\
\textbf{Better Grader:} GPT-5 (CORRECT vs Gemini PARTIALLY) \\
\textbf{Rationale:} ``Swaminarayan'' is a modifier and dropping it is allowed.

\tcbline

\textbf{Direct Question:} What was the name of the exposition at which Westinghouse Electric showcased its AC system to illuminate the ``White City"? \\
\textbf{Gold Answer:} 1893 Columbian Exposition \\
\textbf{Prediction of Gemini-2.5-Flash:} The World's Columbian Exposition \\
\textbf{Better Grader:} GPT-5 (CORRECT vs Gemini PARTIALLY) \\
\textbf{Rationale:} ``World's Columbian Exposition'' is the formal name for the ``1893 Columbian Exposition''.

\tcbline

\textbf{Reverse Question:} On which album did bassist Danny Thompson have his first collaboration on record with John Martyn? \\
\textbf{Gold Answer:} The Road to Ruin \\
\textbf{Prediction of Gemma3-4b+Thinking:} Sure \\
\textbf{Better Grader:} GPT-5 (OTHER vs Gemini INCORRECT) \\
\textbf{Rationale:} ``Sure'' is a refusal or nonsense filler.
\end{prompt}

\section{Prompts}
\label{app_sec:prompts}

\subsection{Facts Extraction Prompts}

\begin{prompt}[label={box:prompt_cat}]{RoyalBlue}{Prompt for Classifying Pages into Categories}
\begin{lstlisting}[basicstyle=\ttfamily, breaklines=true]
You will be provided with a document and its title.
Your task is to analyze the document and assign it to the single, most appropriate category from the predefined list below.

----

**Categories and Descriptions:**

*   **People:** Covers specific individuals, their biographies, life stories, achievements, and influence.
*   **History:** Covers past events and historical periods.
*   **Geography:** Covers Earth's physical features, locations, countries, and cities.
*   **Arts:** Covers all forms of creative expression, including literature, music, visual arts, film, and performance.
*   **Philosophy & Religion:** Covers fundamental principles, belief systems, ethics, religions, and mythological narratives.
*   **Everyday Life:** Covers the practical and routine aspects of human life, including food, clothing, family, sports, and hobbies.
*   **Society:** Covers the structures and systems that govern human communities, such as politics, law, economics, and language.
*   **Science:** Covers the study of the natural and physical world, including biology, physics, chemistry, and mathematics.
*   **Technology:** Covers the application of scientific knowledge, including engineering, computers, and inventions.
*   **Other:** Covers topics that serve as a catch-all and do not clearly fit into any of the other categories.

----

**Output Format:**

Your response must consist of **ONLY** one category name. Do not add any explanation, punctuation, or other text.

Title: {page_title.}
Content: {summary}
\end{lstlisting}
\end{prompt}

\begin{prompt}[label={box:prompt_ner}]{ProcessBlue}{Prompt for Performing Named Entity Recognition}
\begin{lstlisting}[basicstyle=\ttfamily, breaklines=true]
You will be provided with two parts of a paragraph.
Your task is to perform Named Entity Recognition on the *second* part of the paragraph.
Please follow these instructions carefully.

### **Instructions:**

1.  Read the two parts provided at the end of this prompt.
2.  From the second part, identify and select up to a maximum of **eight** entities.
3.  Do not select lengthy entities (more than five words) or entities that are not named entities.
4.  Select specific core entities and avoid overly descriptive phrases that combine multiple separate concepts (e.g., select "documentary photographer" and not "feminist social documentary photographer").
5.  For each entity you select, you must assign an entity type from the following specific list.
6.  You should return a new version of the second part of the paragraph. Copy-paste the original second part and replace the entities you chose with the format `{{{{entity_name; entity_type}}}}`

----

### **Entity Types:**

*   **PERSON:** An individual human, whether real or fictional.
*   **ORGANIZATION:** A structured company, institution, or collective with a common purpose.
*   **LOCATION:** A specific geographical place, region, or feature on Earth.
*   **EVENT:** A notable occurrence or incident that happens at a particular time and place.
*   **WORK_OF_ART:** A specific creative work, such as a book, movie, song, or painting.
*   **PRODUCT:** A commercially produced good, food, service, or piece of technology.
*   **ORGANISM:** A non-human biological life form, including animals, plants, and microbes.
*   **SUBSTANCE:** A material with a distinct physical or chemical composition.
*   **DATE:** A specific point in time, including full dates, years, or historical periods.
*   **NUMERIC_VALUE:** A number representing a specific quantity, measurement, or other quantifiable value.
*   **CONDITION:** A specific state of being, typically a medical, physical, or operational status.
*   **TITLE:** An official position, role, award, or honorific held by a person or group.
*   **LANGUAGE:** Natural languages, dialects, writing systems, including phrases.
*   **GROUP:** A collective of people defined by a shared nationality, ethnicity, or cultural affiliation.
*   **LAW:** A specific, formally enacted rule, regulation, or statute created by a governing body.
*   **IDEA:** An abstract concept, belief system, named phenomenon, or cultural practice.
*   **CLASSIFICATION:** A specific type, genre, format, class or list used to categorize other entities.
*   **IDENTIFIER:** A unique name, code, or alias used as a formal reference or label for another entity
*   **OTHER:** A catch-all category for named entities that do not fit into any of the other defined types.

----

### **Required Output Format:**

Copy-paste the second part of the paragraph. In its new version, replace the entities you chose with the format `{{{{entity_name; entity_type}}}}`, for example: {{{{tennis player; TITLE}}}}. The `entity_name` must be exactly as it appears in the original text.

----

**Example:**

*Part 1:*
...

*Part 2*
`Liam Tarquin Broady (BROH-dee; born 4 January 1994) is a British professional tennis player who competes mainly on the ATP Challenger Tour. He reached a career high ranking of world No. 93 on 25 September 2023.`

*Correct Output:*
{{{{Liam Tarquin Broady; PERSON}}}} (BROH-dee; born {{4 January 1994; DATE}}) is a {{{{British; GROUP}}}} professional {{{{tennis player; TITLE}}}} who competes mainly on the {{{{ATP Challenger Tour; EVENT}}}}. He reached a career high ranking of world No. {{{{93; NUMERIC_VALUE}}}} on 25 September 2023.

Explanation:
1.  **Liam Tarquin Broady**: This is the full name of an individual human, therefore, it as a **PERSON**.
2.  **4 January 1994**: This is a specific point in time, a full **DATE**.
3.  **British**: This word describes a nationality or cultural affiliation. According to the definitions, this is a **GROUP**.
4.  **tennis player**: This is a title describing a person's profession or role. The correct type is **TITLE**.
5.  **ATP Challenger Tour**: This is a named series of tennis tournaments, which is a type of recurring **EVENT**.
6.  **93**: This is a number representing a specific quantity (a world ranking), therefore, it is a **NUMERIC_VALUE**.

**Instruction End**

**Please process the following paragraph and identify up to eight entities:**
{paragraph}
\end{lstlisting}
\end{prompt}

\begin{prompt}[label={box:prompt_selection}]{BlueGreen}{Prompt for Selecting Object Entities}
\begin{lstlisting}[basicstyle=\ttfamily, breaklines=true]
You will be provided with a paragraph annotated with entities in the format of `{{{{entity_name}}}}`.
Your task is to select up to {n_entities} **OBJECT** entities for the pre-training completion task (causal language modeling).
Each OBJECT entity should complete a natural "fill-in-the-blank" based on the preceding text: the *left-context*.
Specifically, the left-context should clearly indicate **what kind of entity** (e.g., a person, country, or biological species) is expected at that position.
However, correctly identifying the **specific entity value** (e.g., "South Africa", "banana plants") should require recalling a world fact.
These selected entities serve as factual anchors for generating our suite of completion-based and QA-based robustness tests.

### **Instructions for selecting suitable OBJECT entities**

* Select OBJECT entities that involve recalling factual knowledge, and is not trivial to guess.
* The left-context (the text that precedes the entity) must suggest **what kind** of entity is expected (e.g., a date, a country, a person's name, a title, a profession, a species, an event, a number, etc).
* The left-context should contain enough information to strongly suggest a single, unique, unambiguous completion.
* OBJECT entities whose specific value can be predicted solely based on the "left-context" or are too easy to guess are considered poor choices.
  * *Example:* `Bang bang chicken (Chinese: ...) is a popular dish in {{{{Chinese cuisine}}}}` -> {{{{Chinese cuisine}}}} is a poor choice because the Chinese name makes the answer too obvious.
  * *Example:* `Airlink is based in Johannesburg, {{{{South Africa}}}}` -> {{{{South Africa}}}} is a poor choice because it can be easily guessed from the mention of Johannesburg, without requiring any knowledge about Airlink.
* The OBJECT should have a strong relationship with an entity mentioned earlier in the text.
  * *Example:* `{{{{Panama disease}}}}... infects {{{{banana plants}}}}` -> Knowing what Panama disease is makes "banana plants" a highly predictable object of the verb "infects".
* Do not select OBJECT entities from the start of a sentence. Ideally, the OBJECT entity should appear at the end of a sentence or complete clause.
* Ensure the left-context provides sufficient disambiguation to yield a unique and specific completion. If the blank could plausibly be filled by multiple entity types, it is not a suitable choice.
  * *Example:* The left-context "Michael Jordan was born on..." is ambiguous, as multiple people share this name (e.g., the basketball player and the ML researcher). In contrast, "Michael Jeffrey Jordan was born on..." and "Michael Jordan, the famous basketball player, was born on..." clearly indicate a single referent, enabling a unique factual prediction.
  * *Example:* `Liam Tarquin Broady is a...` -> This is a poor choice. The context is too broad and could be completed with multiple entities, such as a profession (tennis player), nationality (British), or a descriptor (left-handed).
* The immediate context preceding the entity should constrain the expected entity type, strongly signaling what kind of entity is likely to follow. The OBJECT entity can follow a preposition that narrows down the possibilities.
  * *Example:* `...born on {{{{4 January 1994}}}}` -> The phrase "born on" strongly suggests a date.
  * *Example:* `...caused by the fungus {{{{Fusarium oxysporum f. sp. cubense}}}}` -> The phrase "the fungus" clearly indicates that a specific fungal name should follow.
  * *Example:* `The Virginia Plan (also known as the {{{{Randolph Plan}}}})...` -> The phrase "also known as" implies an alternative name is expected.
* Do not consider any information that appears after the OBJECT entity when making your selection.
* If the uniqueness or the disambiguation of the entity type relies on information that appears after the entity, it is a poor OBJECT choice.
  * *Example:* `It is their first album to feature {{{{Matt Wachter}}}} on bass.` -> {{{{Matt Wachter}}}} is a poor choice. The phrase "It is their first album to feature..." does not constrain the prediction. The disambiguating detail "on bass" comes *after* the entity, which violates the instruction that the **OBJECT** must be inferable solely from the left-context.
* You must select up to {n_entities} suitable OBJECT entities and return them in a list, sorted from the most to the least suitable.

----

### **Examples**

### **Text 1**
{{{{Panama disease}}}} (or Fusarium wilt) is a plant {{{{disease}}}} that infects {{{{banana plants}}}} ({{{{Musa spp.}}}}). It is a wilting disease caused by the fungus {{{{Fusarium oxysporum f. sp. cubense}}}} (Foc). During the {{{{1950s}}}}, an outbreak of Panama disease almost wiped out commercial {{{{Gros Michel}}}} banana production.

*   {{{{Panama disease}}}} - It appears in the first clause of the first sentence and is therefore ineligible. This choice is **poor**.
*   {{{{disease}}}} - This entity does not require recalling a world fact; it can be predicted from the context clues "Fusarium wilt" and "infects." This choice is **poor**.
*   {{{{Musa spp.}}}} - While recalling the scientific name for banana plants is a good test of knowledge, the use of parentheses makes the answer less certain. The context could also imply other related species. This choice is **good**.
*   {{{{banana plants}}}} - The context "a plant disease that infects..." combined with the subject "Panama disease" creates a very strong expectation for this specific plant, a well-known fact. This choice is **excellent**.
*   {{{{Fusarium oxysporum f. sp. cubense}}}} - The preceding phrase "...caused by the fungus..." requires a specific, factual answer. Knowing the fungus that causes Panama disease is a clear test of world knowledge. This choice is **excellent**.
*   {{{{1950s}}}} - The phrase "During the..." is too vague. It could be completed with many different time periods, making the answer unpredictable. This choice is **poor**.
*   {{{{Gros Michel}}}} - The context "...wiped out commercial..." strongly points to a specific type of banana. However, it doesn't uniquely identify the variety without further knowledge, so it's not perfectly constrained. This choice is **good**.

### **Text 2**
Liam Tarquin Broady (BROH-dee; born {{{{4 January 1994}}}}) is a {{{{British}}}} professional {{{{tennis player}}}} who competes mainly on the {{{{ATP Challenger Tour}}}}. He reached a career high ranking of world No. {{{{93}}}} on 25 September 2023.

*   {{{{4 January 1994}}}} - The entity appears in the first clause of the first sentence, making it ineligible per the instructions. This choice is **poor**.
*   {{{{British}}}} - The left-context "...is a..." is too generic and does not sufficiently narrow down the possibilities. This choice is **poor**.
*   {{{{tennis player}}}} - The context "...is a British professional..." strongly implies a profession. Knowing who Liam Broady is leads to a single, specific prediction. This choice is **excellent**.
*   {{{{ATP Challenger Tour}}}} - The context "who competes mainly on the..." points to a tournament or tour. However, the word "mainly" introduces uncertainty, meaning the answer is not guaranteed to be unique. This choice is **good**.
*   {{{{93}}}} - The preceding phrase "...career high ranking of world No..." creates a very specific slot for a number. Recalling a player's exact ranking is a great test of factual knowledge. This choice is **excellent**.

### **Text 3**
Happiness for Beginners is a {{{{2023}}}} American {{{{romantic comedy}}}} film starring {{{{Ellie Kemper}}}} and {{{{Luke Grimes}}}}, an adaptation of the {{{{Katherine Center}}}} 2015 novel of the same name. The novel was adapted for the screen and directed by {{{{Vicky Wight}}}}.

*   {{{{2023}}}} - This entity appears in the first clause, making it ineligible. Additionally, the context "...is a..." is too general to uniquely predict a year. This choice is **poor**.
*   {{{{Ellie Kemper}}}} - The phrase "film starring" clearly points to an actor. However, since a film can have multiple stars, the specific actor named first is not uniquely predictable. This choice is **good**.
*   {{{{Luke Grimes}}}} - The context "...starring Ellie Kemper and..." creates a strong expectation for a co-star, a common structure in this genre. This requires specific knowledge about the film. This choice is **excellent**.
*   {{{{Katherine Center}}}} - The crucial context clue, "novel," appears *after* the entity. Therefore, the answer cannot be predicted from the left-context alone. This choice is **poor**.
*   {{{{Vicky Wight}}}} - The left-context "...adapted for the screen and directed by..." clearly and uniquely signals that a director's name is required. This choice is **excellent**.

----

### **Required Output Format**

Return a Python list of selected **OBJECT** entities, sorted from most suitable to least. Each entity must be written in the format `{{{{entity_name}}}}`, exactly as it appears in the original text.
Example:
["{{{{entity1_name}}}}", "{{{{entity2_name}}}}", "{{{{entity3_name}}}}"]

**Instruction End**

**Please select up to three OBJECT entities from the following annotated paragraph, and return them in a list sorted by the most suitable OBJECT to the least:**
{tagged_paragraph}
\end{lstlisting}
\end{prompt}

\begin{prompt}[label={box:prompt_verification1}]{SeaGreen}{Prompt for Verifying Object Entity Suitability -- with Gold Target}
\begin{lstlisting}[basicstyle=\ttfamily, breaklines=true]
You will be provided with a *left-context* (a non-completed sentence or paragraph) and a proposed **OBJECT** entity that completes it.
Your task is to verify whether this OBJECT entity is a **suitable factual completion**, meaning:

1. The left-context provides enough semantic cues to **strongly constrain** what type of entity should come next (e.g., a date, person, award, location, number, title, etc.).
2. There is a **unique and specific** gold entity completion which is the OBJECT entity:

  * There are **no multiple plausible alternative completions** that refer to different entities.
  * If a language model predicts a different entity, it would **clearly contradict** the information conveyed by the *left-context* and the OBJECT.
  * The OBJECT entity is not underdetermined, vague, or overly general.

----

### **Guidelines for Determining Suitability**

A suitable OBJECT entity must meet **all** of the following:

* The left-context clearly constrains the **type** of entity expected next (e.g., a food, a date, a chemical stabilizer, a person's name, a specific event).
* The OBJECT entity fits seamlessly into the left-context to form a coherent, factual statement.
* The left-context likely **narrows the answer to a unique entity** and any alternative completion would conflict with the facts conveyed so far.

The OBJECT is **not suitable** if:

* The left-context is too vague or underspecified to narrow the completion to a single, specific entity.
* There are multiple plausible completions of the same type (e.g., many possible awards, people, or ingredients).
* The constraints on the OBJECT rely on **right-context** (i.e., text that appears after the entity).
* The OBJECT appears to fit syntactically or thematically, but the left-context provides **no strong evidence** that it is the only valid answer.

*Clarification on Uniqueness:* uniqueness refers to the **semantic identity** of the OBJECT entity and not to surface forms (the string used to represent an entity). The OBJECT is still considered unique even if it has other forms that:

* Differ in capitalization, punctuation, or grammatical form (e.g., pluralization, articles).
* Omit non-essential modifiers (e.g., "Pulitzer Prize for Fiction" vs. "2003 Pulitzer Prize for Fiction").
* Use a common variant or alias (e.g., "United States" vs. "U.S.").

However, if multiple **distinct** entities (e.g., different stabilizers, places, awards, or people) could reasonably complete the left-context, then the OBJECT is **not suitable**.

----

### **NOT SUITABLE Examples**

**Left-context:**
Garfield is portrayed as a lazy, fat, cynical, and self-absorbed orange tabby Persian cat. He is noted for his love of lasagna and *
**OBJECT:** pizza
**Explanation:** The left-context does not constrain the answer to be a food. Other plausible completions include "naps", "Mondays", or "sleep". The OBJECT is not uniquely inferable.


**Left-context:**
Glass noodles, sometimes called cellophane noodles, are a type of transparent noodle made from starch (such as mung bean starch, potato starch, sweet potato starch, tapioca, or canna starch) and water. They originated in China. A stabilizer such as chitosan or *
**OBJECT:** alum
**Explanation:** The left-context signals that a stabilizer is expected, but it does not sufficiently constrain which one. Multiple valid completions exist, so the OBJECT is not unique.


**Left-context:**
Rayagada railway station is located at the heart of the city and now it has the longest platform in Odisha (Platform No :- 3 measuring 910m), the 6th in the country and the
**OBJECT:** 7th
**Explanation:** The left-context does not clarify what the "7th" refers to. The comparison class is underspecified (7th in what? globally? regionally?), so the answer is ambiguous.

----

### **SUITABLE Examples**

**Left-context:**
Adrian Darnell Griffin Sr. (born July 4, 1974) is an American professional basketball coach and former player who most recently served as the head coach for the Milwaukee Bucks of the National Basketball Association (NBA). He played in the NBA as a shooting guard and small forward from 1999 to 2008. Griffin grew up in *
**OBJECT:** Wichita
**Explanation:** The phrase "grew up in" combined with known biographical details provides strong context for a specific city. Other completions would clearly contradict the known fact.


**Left-context:**
Andrea Barrett (born November 16, 1954) is an American novelist and short story writer. Her collection Ship Fever won the 1996 U.S. National Book Award for Fiction, and she received a MacArthur Fellowship in 2001. Her book Servants of the Map was a finalist for the 2003 *
**OBJECT:** Pulitzer Prize for Fiction
**Explanation:** While there are many awards, the combination of book title, year, and prior accolades strongly constrains the answer. The context likely narrows the answer to a unique award.


**Left-context:**
Bering Air Flight 445 was a scheduled domestic flight from Unalakleet Airport to Nome Airport which crashed on February 6, 2025 while flying over the Norton Sound. The flight was aboard a *
**OBJECT:** Cessna 208B Grand Caravan EX
**Explanation:** The left-context refers to a known historical event. There is only one aircraft associated with this flight, making the answer uniquely identifiable.


**Left-context:**
George Mark Collie (born January 18, 1956) is an American country music singer, songwriter, musician, actor, record producer, and fundraiser for Type 1 diabetes study. He has won awards and acclaim for his music, his acting, and his philanthropy. His singing career has included five major-label albums: four for MCA Nashville and one for Giant Records. Sixteen of his singles have charted on *
**OBJECT:** Hot Country Songs
**Explanation:** The specific number of charting singles and genre context constrain the answer to a unique Billboard chart. Alternate completions would contradict the left-context.

----

### **Required Output Format**

Respond only with "SUITABLE" or "NO".

**Instruction End**

**Please process the following left-context and the OBJECT entity:**

**Left-context:** {left_context}
**OBJECT:** {object_entity}
\end{lstlisting}
\end{prompt}

\begin{prompt}[label={box:prompt_verification2}]{YellowGreen}{Prompt for Verifying Object Entity Suitability -- without Gold Target}
\begin{lstlisting}[basicstyle=\ttfamily, breaklines=true]
You will be provided with a *left-context* (a non-completed sentence or paragraph). Your task is to verify whether this left-context is a **suitable factual completion task**, meaning:

1. The left-context provides enough semantic cues to **strongly constrain** what type of entity should come next (e.g., a date, person, award, location, number, title, etc.).
2. There is a **unique and specific** gold entity completion:

  * There are **no multiple plausible alternative completions** that refer to different entities.
  * If a language model predicts a different entity than the gold entity, it would **clearly contradict** the information conveyed by the *left-context* and the gold entity.

----

### **Guidelines for Determining Suitability**

A suitable left-context must meet **all** of the following:

* The left-context clearly constrains the **type** of entity expected next (e.g., a food, a date, a chemical stabilizer, a person's name, a specific event).
* The left-context likely **narrows the answer to a unique entity** and any alternative completion would conflict with the facts conveyed so far.

The left-context is **not suitable** if:

* The left-context is too vague or underspecified to narrow the completion to a single, specific entity.
* There are multiple plausible completions of the same type (e.g., many possible awards, people, or ingredients).

*Clarification on Uniqueness:* uniqueness refers to the **semantic identity** of the completion and not to surface forms (the string used to represent an entity). The completion is still considered unique even if it has other forms that:

* Differ in capitalization, punctuation, or grammatical form (e.g., pluralization, articles).
* Omit non-essential modifiers (e.g., "Pulitzer Prize for Fiction" vs. "2003 Pulitzer Prize for Fiction").
* Use a common variant or alias (e.g., "United States" vs. "U.S.").

However, if multiple **distinct** entities (e.g., different stabilizers, places, awards, or people) could reasonably complete the left-context, then it is **not suitable**.

----

### **NOT SUITABLE Examples**

**Left-context:**
Garfield is portrayed as a lazy, fat, cynical, and self-absorbed orange tabby Persian cat. He is noted for his love of lasagna and *
**Explanation:** The left-context does not constrain the answer to a single entity type. There are many plausible completions such as "pizza", "naps", "Mondays", or "sleep".


**Left-context:**
Glass noodles, sometimes called cellophane noodles, are a type of transparent noodle made from starch (such as mung bean starch, potato starch, sweet potato starch, tapioca, or canna starch) and water. They originated in China. A stabilizer such as chitosan or *
**Explanation:** The left-context signals that a stabilizer is expected, but it does not sufficiently constrain which one. Multiple valid completions exist, so the completion is not unique.


**Left-context:**
Rayagada railway station is located at the heart of the city and now it has the longest platform in Odisha (Platform No :- 3 measuring 910m), the 6th in the country and the *
**Explanation:** The left-context does not clarify what should come after "...and the". Even if a number is likely ("...and the Xth..."), the comparison class is underspecified (Xth in what? globally? regionally?), so the answer is ambiguous.

----

### **SUITABLE Examples**

**Left-context:**
Adrian Darnell Griffin Sr. (born July 4, 1974) is an American professional basketball coach and former player who most recently served as the head coach for the Milwaukee Bucks of the National Basketball Association (NBA). He played in the NBA as a shooting guard and small forward from 1999 to 2008. Griffin grew up in *
**Explanation:** The phrase "grew up in" combined with known biographical details provides strong context for a specific city. Other completions would clearly contradict the known fact.


**Left-context:**
Andrea Barrett (born November 16, 1954) is an American novelist and short story writer. Her collection Ship Fever won the 1996 U.S. National Book Award for Fiction, and she received a MacArthur Fellowship in 2001. Her book Servants of the Map was a finalist for the 2003 *
**Explanation:** While there are many awards, the combination of book title, prior accolades and the year, strongly constrains the answer. The left-context likely narrows the answer to a unique award.


**Left-context:**
Bering Air Flight 445 was a scheduled domestic flight from Unalakleet Airport to Nome Airport which crashed on February 6, 2025 while flying over the Norton Sound. The flight was aboard a *
**Explanation:** The left-context refers to a known historical event. There is only one aircraft associated with this flight, making the answer unique.


**Left-context:**
George Mark Collie (born January 18, 1956) is an American country music singer, songwriter, musician, actor, record producer, and fundraiser for Type 1 diabetes study. He has won awards and acclaim for his music, his acting, and his philanthropy. His singing career has included five major-label albums: four for MCA Nashville and one for Giant Records. Sixteen of his singles have charted on *
**Explanation:** The specific number of charting singles and genre context constrain the answer to a unique Billboard chart. Alternate completions would contradict the left-context.

----

### **Required Output Format**

Respond only with "SUITABLE" or "NO".

**Instruction End**

**Please process the following left-context:**

**Left-context:** {left_context}
\end{lstlisting}
\end{prompt}

\begin{prompt}[label={box:prompt_difficulty}]{YellowGreen}{Prompt for Verifying Completion Difficulty}
\begin{lstlisting}[basicstyle=\ttfamily, breaklines=true]
You will be provided with a *left-context* (a non-completed sentence or paragraph) and a proposed **OBJECT** entity that completes it.
Your task is to verify whether this OBJECT entity is a **suitable factual completion**.

The OBJECT is **suitable** if:

1. The completion is challenging and requires specific factual knowledge.
2. It is not trivially easy to guess or logically infer from the left-context alone.
3. The left-context does not contain the answer or any strong hints that directly disclose the OBJECT entity.
4. The OBJECT entity is not time-sensitive and will not change over time.

The OBJECT is **not suitable** if:

1. The OBJECT entity can be predicted using general common sense or an educated guess.
2. The OBJECT entity is trivially inferable using general common knowledge.
3. The OBJECT entity, as a completion of the left-context, is subject to change over time (it is not static).

----

### **NOT SUITABLE Examples**

**Left-context:**
The Strait of Hormuz (Persian: ... Tangeh-ye Hormoz) is situated between *
**OBJECT:** Iran
**Explanation:** Since the left-context provides the name in Persian script, the answer "Iran" can be easily guessed. Therefore, the completion is not challenging and is considered not suitable.


**Left-context:**
The Nobel Prize in Literature is a Swedish literature prize that is awarded annually to an author from any country who has, in the words of Alfred Nobel, "in the field of literature, produced the most outstanding work in an idealistic direction". It is one of the five *
**OBJECT:** Nobel Prizes
**Explanation:** The completion is trivial because the subject ("The Nobel Prize in Literature") already contains the target OBJECT entity.


**Left-context:**
Income in the United States is measured by the various federal agencies including the Internal Revenue Service, Bureau of Labor Statistics, US Department of Commerce, and the US Census Bureau. Overall, including all households/individuals regardless of employment status, the median household income was *
**OBJECT:** $67,521
**Explanation:** The OBJECT entity is time-sensitive because the specific year is not specified in the context. Since median income changes over time, this completion is not suitable.

----

### **SUITABLE Examples**

**Left-context:**
Redbus India Private Limited, doing business as redBus, is an Indian multinational online bus-ticketing platform. In 2018, the company achieved a GMV of *
**OBJECT:** 50 billion Indian Rupee
**Explanation:** The OBJECT entity is not time-sensitive because the context restricts the fact to a specific year (2018), making the value static.


**Left-context:**
"Thriller" is a song by the American singer Michael Jackson. It includes a spoken-word sequence performed by the horror actor *
**OBJECT:** Vincent Price
**Explanation:** This is a challenging completion that requires the recall of specific factual knowledge; therefore, it is suitable.


**Left-context:**
The Panathenaic Stadium is a multi-purpose stadium in Athens, Greece. After the rise of Christianity in the 4th century it was largely abandoned. After being refurbished, it hosted the opening and closing ceremonies of the *
**OBJECT:** first modern Olympics
**Explanation:** Although the stadium's location in Greece hints that the OBJECT is related to the Olympic Games, identifying the specific OBJECT requires knowing it hosted the *first modern* Olympics. This makes the completion non-trivial and suitable.

----

### **Required Output Format**

Respond only with "SUITABLE" or "NO".

**Instruction End**

**Please process the following left-context:**

**Left-context:** {left_context}
**OBJECT:** {object_entity}
\end{lstlisting}
\end{prompt}

\subsection{Question Generation Prompts}

\begin{prompt}[label={box:prompt_direct}]{Mahogany}{Prompt for Generating the \textit{Direct} Task}
\begin{lstlisting}[basicstyle=\ttfamily, breaklines=true]
You will be provided with a *left-context* (an incomplete sentence or paragraph) and an *OBJECT* entity that completes it. Based only on the left-context, your task is to design a unambiguous question for which the only correct and specific answer is the OBJECT. The question should only use information necessary for disambiguation and to ensure the answer's specificity, using the exact phrasing from the left-context.

**Requirements:**

*   Only the OBJECT should be the correct answer.
*   The question must be self-contained: it should not use pronouns or references that require outside context.
*   Use only the information available in the left-context.
*   The question must be unambiguous: its phrasing should be clear and avoid multiple interpretations.
*   The question must be specific and precise: it should specify the level of detail and type of answer required (e.g., "What year...", "What is the name of the condition...", "Which city...").
*   The question must be minimal: include only the essential details from the left-context needed to make the question unambiguous and ensure the OBJECT is the only possible answer.
*   Use the same terms, phrases, and words that appear in the left-context. Do not rephrase or use synonyms, unless essential for grammaticality or clarity.

----

### **Required Output Format**

Return only the question without any additional text.

----

### **Examples**

Here are a few examples, including explanations for how the questions were formulated:

1.
**Left-context:** Erika Nordby (born February 2000), also known as Baby Erika, Miracle Baby and Canada's Miracle Child, is a Canadian originally from Edmonton, Alberta known for having been revived after spending two hours without a heartbeat due to
**OBJECT:** hypothermia
**Output:** What was the condition that caused Erika Nordby to spend two hours without a heartbeat before she was revived?
*Explanation:*
*   We ask "What was the condition..." because hypothermia is a medical condition. This phrasing demands a specific type of answer.
*   We include "Erika Nordby" to identify the person the question is about, making it self-contained.
*   We include "two hours without a heartbeat before she was revived" because these are the key, unique details from the context that pinpoint the specific cause being asked about.
*   We do not include the birth date or the nicknames to make the question minimal. The combination of the name and the incident of 'two hours without a heartbeat' is enough for disambiguation.
*   All terms are taken directly from the left-context.

2.
**Left-context:** When I Look in Your Eyes is the fifth studio album by singer Diana Krall, released on
**OBJECT:** June 8, 1999
**Output:** What was the release date of When I Look in Your Eyes by Diana Krall?
*Explanation:*
*   The phrasing "What was the release date..." is precise and directly corresponds to the left-context phrase "released on."
*   The album title could be ambiguous on its own, so we add "by Diana Krall" for disambiguation.
*   To keep the question minimal, we do not include the fact that it was her fifth studio album. The combination of the album title and the artist's name is sufficient for disambiguation.

3.
**Left-context:** Celebrimbor is a fictional character in J. R. R. Tolkien's legendarium. In Tolkien's stories, Celebrimbor was an elven-smith who was manipulated into forging the Rings of Power by the Dark Lord Sauron, in a fair disguise and named
**OBJECT:** Annatar
**Output:** What name did Sauron use in his fair disguise to manipulate Celebrimbor?
*Explanation:*
*   "What name..." sets the clear expectation for the type of answer required.
*   We must include "in his fair disguise" and "to manipulate Celebrimbor" because these are the specific circumstances under which Sauron used the name Annatar. Simply asking "What name did Sauron use?" would be too broad.
*   Further information is not required to make the question unique and unambiguous. For example, mentioning Celebrimbor and Sauron is enough to understand that they are characters from Tolkien's legendarium.
*   All terms are taken directly from the left-context.

4.
**Left-context:** The horse-collar tackle is a gridiron football maneuver in which a defender tackles another player by grabbing the back collar or the back-inside of an opponent's shoulder pads and pulling the ball carrier directly downward violently in order to pull his feet from underneath him. The technique is most closely associated with Pro Bowl safety Roy Williams. After being blamed for a series of major injuries in the 2004 season , the horse-collar tackle was banned from the NFL during the 2005 off-season. The rule forbidding it is often referred to in the press as
**OBJECT:** The Roy Williams Rule
**Output:** What is the name the press often uses to refer to the rule forbidding the horse-collar tackle from the NFL?
*Explanation:*
*   To make the question self-contained, we must describe the rule being referenced: "the rule forbidding the horse-collar tackle from the *NFL*".
*   The left-context specifies the rule is "referred to in the press as..." This detail is critical. Therefore, the question includes "What is the name the press often uses to refer to..." to remain faithful to the source text.
*   The years, the description of the maneuver, and the injuries are unnecessary for understanding the question and can be excluded for minimalism.
*   All terms are taken directly from the left-context.

**Instruction End**

**Please process the following left-context and OBJECT:**

**Left-context:** {left_context}
**OBJECT:** {object_entity}
\end{lstlisting}
\end{prompt}

\begin{prompt}[label={box:prompt_specification}]{Red}{Prompt for Refining Specification}
\begin{lstlisting}[basicstyle=\ttfamily, breaklines=true]
You will be provided with a question, a gold answer, and its corresponding entity type. Your task is to evaluate the question's specificity and precision, with respect to the gold answer, and revise the question if necessary.
A specific and precise question should clearly define the level of detail required in the answer.
Your evaluation process should consider the following stages: **Verification**, **Revision** and **Rejection**.

### **Verification**

Determine if the question is specific and precise. Consider the following when making your decision:

*   Would a person with the necessary knowledge, who is familiar with the gold answer, be led to provide that specific gold answer to the question?
*   Does the question specify the required level of detail and type for the answer?
*   Is it easy to infer the required entity type of the answer from the question?
*   If relevant, does the question specify the required granularity of the answer?
    *   For example: "What year...", "What year and month...", "What date...", "What is the name of the condition...", "Which city...", "Which country...", "Which province...", "What is the scientific name...".

If the question is specific and precise, you can stop here and return it exactly as it is. Note that most of your inputs will be considered specific and precise.
If the question is not specific or precise, you should try to revise it.

### **Revision**

A question should be revised if it is not specific but can be fixed with **minimal changes**.
A valid revision must result in a question that perfectly and uniquely targets the gold answer without leaking non-trivial information.
Common revision techniques involve adding a short, clarifying phrase or changing a single word, for example:

*   **Specify the answer's entity type.** This is often done by changing the question word to be more precise, e.g., `Who` -> `Which company`.
*   **Specify the answer's required granularity or format.** This is often done by clarifying the necessary level of detail. e.g., `When` -> `In what month and year`, `Where` -> `In which city and country`,
 `What name` -> `What is the scientific name`.
*   **Specify the units or precision for numerical data.** This removes ambiguity from the expected answer. e.g., Adding `in Celsius`, `in US dollars`, or `in millions`.
*   **Add a qualifier to narrow the scope.** When a question is ambiguous and could have multiple correct answers, add a qualifier to focus the question on the single intended answer.  Note that you should not disclose non-trivial information. e.g., Revise `What ecosystem...`? to `What ecosystem **of mountains**...`? to distinguish from other valid answers (e.g., a forest).

### **Rejection**

If a question cannot be fixed with a valid minimal revision, it must be rejected.
A question is considered unfixable and should be rejected in the following cases:
*   The revision would leak information: The only way to fix the question would be to add a non-trivial clue that is part of the gold answer itself, making the question significantly easier.
*   The revision would require a major change: The question is too vague or fundamentally mismatched with the answer, and fixing it would require a complete rewrite rather than a simple, minimal clarification.

----

### **Required Output Format**

You must return *only* a string, which can be the original question, the revised question, or "REJECT".
If the question is specific and precise as it is, return the question. If the question needs revision, revise it and return the revised question. If you conclude that the question should be rejected, return "REJECT".
Remember, return only the original question, the revised question, or "REJECT", without any additional text.

----

### **Examples**

1.
**Question:** What is the condition that caused Erika Nordby to spend two hours without a heartbeat before she was revived?
**Gold answer:** hypothermia
**Explanation:** The question is specific and precise.
**Output:** What is the condition that caused Erika Nordby to spend two hours without a heartbeat before she was revived?

2.
**Question:** What name did Sauron use in his fair disguise to manipulate Celebrimbor?
**Gold answer:** Annatar
**Explanation:** The question is specific and precise.
**Output:** What name did Sauron use in his fair disguise to manipulate Celebrimbor?

3.
**Question:** Who produced the 2020 German romantic comedy film Isi & Ossi?
**Gold answer:** Netflix
**Explanation:** The question word "Who" is ambiguous, as it typically refers to a person. The revision to "Which company" precisely targets the entity type of the answer.
**Output:** Which company produced the 2020 German romantic comedy film Isi & Ossi?

4.
**Question:** What biodiverse ecosystem surrounds Lake Burton?
**Gold answer:** Blue Ridge Mountains
**Explanation:** The term "ecosystem" is too broad, allowing for multiple correct answers (e.g., a forest). The qualifier "of mountains" narrows the scope to the intended answer without leaking information.
**Output:** What biodiverse ecosystem of mountains surrounds Lake Burton?

**Instruction End**

**Please process the following question, gold answer, and entity type:**

**Question:** {question}
**Gold answer:** {gold_answer}
**Entity type:** {entity_type}
\end{lstlisting}
\end{prompt}

\begin{prompt}[label={box:prompt_minimalism}]{RedOrange}{Prompt for Refining Minimalism}
\begin{lstlisting}[basicstyle=\ttfamily, breaklines=true]
You will be provided with a question and its gold answer. Your task is to evaluate the question's minimalism and revise it by removing any non-essential information.
The question must be unambiguous and lead to a single, unique answer, while avoiding non-essential details that are not required for the disambiguation of the core entities in the question or the answer.
The primary goal is not simply to make the question shorter, but to remove information that acts as an unnecessary hint (which helps answer the question) but does not aid in disambiguation.
Your evaluation process should consider the following stages: **Verification** and **Revision**.

### **Verification**

Determine if the question is already minimal. A question is considered minimal if it uniquely leads to the provided gold answer without containing unnecessary details. Consider the following:

*   Are all entities in the question (people, works, events) identified with just enough detail to be unambiguous for the purpose of the question? For example, adding "by Diana Krall" is essential if the song title "When I Look in Your Eyes" might not be unique on its own, as there may be another song with the same title.
*   Does the question contain any superfluous words or phrases (e.g., adjectives, specific years, genres) that could be removed without affecting the question's ability to point uniquely to the required type of gold answer?
*   Does the question avoid providing unnecessary hints or clues? A hint is a piece of information that makes the answer easier but is not needed for the identification of the entities or the intended answer from other possibilities.
*   Note: Most questions are already minimal. Minimalism is not about shorter questions, but about not including details that provide a hint without being essential for the question's disambiguation.

If the question is already minimal, you can stop here and return it exactly as it is. Note that many questions, even if long, will be considered minimal if all parts are essential.

### **Revision**

A question should be revised if it contains non-essential information that can be removed while preserving the question's disambiguity, entity identification, and answer specification.
A valid revision consists **only of deleting** words or phrases. You must not add, rephrase, or reorder any part of the question. The original wording of the remaining parts must be preserved.
You can perform a minimal change to fix the grammar.

After deleting a word or phrase, you must confirm that the revised question still:
1.  Makes sense grammatically and logically.
2.  Uniquely and unambiguously points to the provided gold answer.

Common details to remove for minimalism include:

*   **Redundant Specifiers:** Details like a specific year or genre, *if* another, more unique identifier in the question already makes them unnecessary.
    *   Example: "Which **1958 crime** drama by Orson Welles featured..." might be reducible if another part of the question is sufficiently unique.
*   **Superfluous Descriptors:** Words that describe an entity but are not needed to distinguish it from other entities.
    *   Example: Removing "**country and gospel**" from "...quartet the Oak Ridge Boys," as the band's name is a unique identifier.
*   **Leaked Clues:** Information that provides a separate, related fact not essential to identifying the answer.
    *   Example: Removing "**defending champion**" from "...team contested the 1966 Stanley Cup Finals," as it reveals the result of the previous year's final.

**Important:**
*   Do not remove essential phrases that describe the core subject of the question (e.g., "...based on the life of a Military Commander...").
*   Do not shorten or remove official or commonly used names (e.g., keep "Ebola virus disease") and do not remove abbreviations.
*   Keep in mind that MOST questions are already minimal. Your goal is not simply to shorten the question, but to remove any information that is unnecessary for disambiguation and acts as an unneeded hint.
----

### **Required Output Format**

You must return *only* a string, which can be the original question or the revised question.
If the question is minimal as it is, return the question. If the question needs revision, revise it by deleting the non-essential parts and returning the revised question.
Remember to return only the question string, without any additional text.

----

### **Examples**

1.
**Question:** What name did Sauron use in his fair disguise to manipulate Celebrimbor?
**Gold answer:** Annatar
**Explanation:** The question is already minimal. "in his fair disguise" and "to manipulate Celebrimbor" are both essential context to specify which of Sauron's names is being asked for. Removing either phrase would make the question ambiguous.
**Output:** What name did Sauron use in his fair disguise to manipulate Celebrimbor?

2.
**Question:** The film "Sarsenapati Hambirrao" is based on the life of a Military Commander of which empire?
**Gold answer:** Maratha Empire
**Explanation:** The question is already minimal. The phrase "the life of a Military Commander" is the core subject of the question and clarifies which empire is relevant (the one the commander served, the film might focus on more than a single empire). This is not an unnecessary hint.
**Output:** The film "Sarsenapati Hambirrao" is based on the life of a Military Commander of which empire?

3.
**Question:** What vocal part does Richard Anthony Sterban sing in the country and gospel quartet the Oak Ridge Boys?
**Gold answer:** Bass
**Explanation:** The name of the quartet, "the Oak Ridge Boys," is a sufficient identifier. The genres "country and gospel" are superfluous descriptors that are not needed for disambiguation. They can be removed.
**Output:** What vocal part does Richard Anthony Sterban sing in the quartet the Oak Ridge Boys?

4.
**Question:** Which defending champion team contested the 1966 Stanley Cup Finals against the Detroit Red Wings?
**Gold answer:** Montreal Canadiens
**Explanation:** The phrase "defending champion" is a leaked clue. It provides an additional fact (that the team won the 1965 finals) that is not necessary to uniquely identify the team that played the Red Wings in the 1966 finals. This non-essential information should be removed.
**Output:** Which team contested the 1966 Stanley Cup Finals against the Detroit Red Wings?

**Instruction End**

**Please process the following question and gold answer:**

**Question:** {question}
**Gold answer:** {gold_answer}
\end{lstlisting}
\end{prompt}

\begin{prompt}[label={box:prompt_search}]{Orange}{Prompt for Filtering with Google Search Grounding}
\begin{lstlisting}[basicstyle=\ttfamily, breaklines=true]
You will be provided with a question and a corresponding answer written by an expert. Your task is to evaluate if the question is **suitable** based on the provided expert answer.
A **suitable question** must be unambiguous, specific, and have a single, unique answer. Your evaluation should be based *solely* on the content of the expert's answer.
Your final output must be only one word: **SUITABLE** or **NOT**.

----

### **Evaluation Process**

Carefully analyze the question and the expert's answer to determine the question's suitability.

**Definition of a Suitable Question**

A question is considered suitable if it meets all the following criteria:
1.  **Unambiguous:** All entities in the question can be easily and uniquely identified. There is no need to ask for clarification to understand the exact intent of the question.
2.  **Specific and precise:** The question asks for a specific and precise piece of information, making it easy to understand the expected type and format of the answer.
3.  **Has a single and unique answer:** The question is structured in a way that there is only one possible correct answer.

**Determining Suitability Based on the Expert's Answer**

You must determine the question is **NOT** suitable if the expert's answer exhibits any of the following characteristics:

*   **The answer asks for clarification:** The expert's response includes questions back to the user to resolve ambiguity.
    *   *Example indicators:* "Which 'Scarlet' are you referring to?", "Could you please specify the year?", "It's not clear what you mean by 'official language'..."
*   **The answer states there are multiple possible answers:** The expert explicitly mentions that the question has more than one valid answer.
    *   *Example indicators:* "There are several former presidents of the USA...", "Multiple actors have played this role...", "Several cities match this description..."
*   **The answer provides different answers for different interpretations:** The expert identifies ambiguity and provides separate answers for each possible interpretation of the question.
    *   *Example indicators:* "If you are asking about the actress Scarlet Johansson, the answer is... However, if you mean the literary character Scarlett O'Hara, the answer is...", "For the 2008 movie, the director was... For the 2021 remake, the director was..."

A question is considered **SUITABLE** only if the expert's answer provides a direct, singular, and factual response without expressing any confusion, highlighting ambiguity, or indicating the existence of multiple valid answers.

----

### **Required Output Format**

You must return *only* a string, either **SUITABLE** or **NOT**, without any additional text or explanation.

**Instruction End**

**Please process the following question and the expert's answer:**

**Question:** {question}
**Answer:** {search_pred}
\end{lstlisting}
\end{prompt}

\begin{prompt}[label={box:prompt_reverse}]{YellowOrange}{Prompt for Generating the \textit{Reverse} Task}
\begin{lstlisting}[basicstyle=\ttfamily, breaklines=true]
You will be provided with a context, a question, and an answer. The answer is an OBJECT entity that also completes the last sentence of the provided context.
Your task is to generate a *reverse question*. In this new question, the SUBJECT entity of the original question will be the answer. The OBJECT entity from the original question will become the new subject of the reverse question.

The process involves three main stages: **Identification**, **Generation**, and **Evaluation**.

### **Stage 1: Identify the SUBJECT Entity**

Analyze the original question to identify its SUBJECT entity. The SUBJECT is the main focus of the question, and the question asks for information related to this entity.
For example, in the question "What is the capital of France?" the SUBJECT is "France".

### **Stage 2: Generate a Reverse Question**

Create a new question where the original SUBJECT entity becomes the answer, and the original OBJECT entity becomes the new subject.
The reverse question must be an unambiguous question with a single, unique, and specific answer: the SUBJECT. It must be based on the given question and act as its reverse.
The reverse question should follow these requirements:

*   **Self-contained:** It does not use pronouns or references that require outside context.
*   **Unambiguous and Specific:** Its phrasing is clear, avoids multiple interpretations, and specifies the level of detail required in the answer (e.g., "What year...", "What is the name of the condition...").
*   **Answer Uniqueness:** It must be phrased to have a *single* valid and correct answer which is the SUBJECT.
*   **Minimal:** It includes only the essential details from the context and the original question needed to make the reverse question unambiguous and ensure the SUBJECT is the only valid answer.
*   **Verbatim:** It uses the same terms, phrases, and words that appear in the context and the given question, without rephrasing.
*   **Challenging:** The reverse question should be challenging and non-trivial to answer. The difficulty level of the original question and the reverse question should be roughly the same.

**IMPORTANT:** Although not needed in most cases, you may sometimes add information from the *context* to the reverse question, provided it is required to make the question unambiguous or to ensure the SUBJECT is the only valid answer. However, this information must be **minimal** and must **not shift the focus** of the question. It should not introduce a hint or clue that allows solving the reverse question without knowledge of the core relationship between the original OBJECT and SUBJECT. The added detail should not become a shortcut to the answer and should not make the question easier without helping in disambiguation.


### **Stage 3: Determine Suitability and Rejection**

Finally, you must assess the reverse question you generated. If it fails to meet the requirements above, you must reject it.
A question should be rejected if it is impossible to create a reverse version that is unambiguous and has a single valid answer.
It should also be rejected if any attempt to make it meet the requirements shifts the focus of the question or makes it trivial or easier.

----

### **Required Output Format**

Your output must be in one of two formats.

If you generate a successful reverse question, the response must be structured using the following tags: `<reverse>` and `<subject>`. Do not include any other text, explanations, or formatting.

`<reverse>` The reverse question `</reverse>`
`<subject>` The subject entity, which is the answer to the reverse question `</subject>`

If a suitable reverse question cannot be generated, you must return only the string:
`REJECT`

----

### **Examples**

Here are several examples demonstrating the task. Pay close attention to the explanations.

1.
**Context:** Cody Blake Dennison (born March 4, 1990) is an American professional stock car racing driver and YouTube personality who competes part-time in the NASCAR Craftsman Truck Series, driving the Nos. 2/22 Ford F-150
**Question:** What is the model of the Nos. 2/22 truck that Cody Blake Dennison drives in the NASCAR Craftsman Truck Series?
**OBJECT:** Ford F-150
**Output:**
<reverse> Who drives the Nos. 2/22 Ford F-150 in the NASCAR Craftsman Truck Series? </reverse>
<subject> Cody Blake Dennison </subject>
**Explanation:** We use the same words and terms from the question.

2.
**Context:** Summer Side of Life is Canadian singer-songwriter Gordon Lightfoot's sixth studio album. It was released in 1971
**Question:** In what year was Gordon Lightfoot's album, Summer Side of Life, released?
**OBJECT:** 1971
**Output:**
<reverse> What is the name of Gordon Lightfoot's album that was released in 1971? </reverse>
<subject> Summer Side of Life </subject>
**Explanation:** It is likely that the singer released only a single album in 1971, making the answer single and unique.

3.
**Context:** Diabolik: Who Are You? (Italian: Diabolik - Chi sei?) is a 2023 Italian crime action film directed by the Manetti Bros. and based on the 1968 Diabolik comic strip Diabolik, chi sei? by Angela and Luciana Giussani. It premiered at the 18th Rome Film Festival
**Question:** At which event did the film 'Diabolik: Who Are You?' premiere?
**OBJECT:** Rome Film Festival
**Output:**
<reverse> What film directed by the Manetti Bros. premiered at the 18th Rome Film Festival? </reverse>
<subject> Diabolik: Who Are You? </subject>
**Explanation:** A reverse question like, "What film premiered at the Rome Film Festival?" is ambiguous. To ensure a unique answer, we add two key details, forming the question: "What film directed by the Manetti Bros. premiered at the 18th Rome Film Festival?". We specify the "18th" edition because the festival is held annually and would otherwise have multiple answers over the years. We add "directed by the Manetti Bros." to narrow the options at that specific event, ensuring a single valid answer. In contrast, adding the detail "...based on the 1968 Diabolik comic strip Diabolik, chi sei?" would have been incorrect because it shifts the focus. The question would no longer be about the film's premiere but would become a simple trivia question about identifying a comic book adaptation.

4.
**Context:** WWWQ (99.7 FM) is a commercial radio station carrying a contemporary hit radio format known as "Q99.7". Owned by Cumulus Media
**Question:** What company owns the commercial radio station WWWQ?
**OBJECT:** Cumulus Media
**Output:**
REJECT
**Explanation:** A reverse question like, "What commercial radio station is owned by Cumulus Media?" is not suitable because the company owns many stations. Adding a detail like "...that is known as 'Q99.7'?" or "...that operates on 99.7 FM?" shifts the focus. The question is no longer about ownership but about knowing the station's slogan or frequency. This turns the added detail into a shortcut. Since no additional context can be provided, this item must be rejected.

**Instruction End**

**Please process the following context, question, and its answer (OBJECT entity):**

**Context:** {context}
**Question:** {question}
**OBJECT:** {answer}
\end{lstlisting}
\end{prompt}

\begin{prompt}[label={box:prompt_reverse_verification}]{YellowOrange}{Prompt for Verifying a Valid \textit{Reverse} and \textit{Direct} Pair}
\begin{lstlisting}[basicstyle=\ttfamily, breaklines=true]
You will be provided with a *Direct Question* (with its answer) and a *Reverse Question* (with its answer).
Your task is to verify whether this pair represents a *suitable factual evaluation pair*.

A fact is defined as a proposition between two entities: a subject and an object.
* In the *Direct Question*, the target is the **Object** entity.
* In the *Reverse Question*, the roles are swapped: the Object of the direct question becomes the Subject (or is explicitly mentioned) in the reverse question, and the target is the original Subject entity.

The pair is **suitable** if:

1.  The questions represent a valid direct/reverse relationship. The answer (Object) of the Direct Question is included in the Reverse Question.
    * *Note:* It is acceptable if the information conveys different hints to resolve ambiguity, or if the Object of the direct question is not the main entity, topic, or subject of the reverse question, as long as it is explicitly present.
2.  The questions are challenging and require specific factual knowledge.
    * The answers are not contained within the questions themselves.
    * The answers cannot be easily guessed using general common sense or linguistic clues.
3.  The answers are not time-sensitive.
    * If the fact is subject to change (e.g., revenue, statistics, roles), the question must specify a date/year to make the answer static.

The pair is **not suitable** if:

1.  The answer to the Direct Question (the Object) is missing from the Reverse Question.
2.  One of the answers can be easily predicted via an educated guess.
3.  One of the questions asks for dynamic information (like statistics or current roles) without specifying a timeframe, making the answer subject to decay.

----

### **NOT SUITABLE Examples**

**Direct Question:** What language's complete form was published in the book Toki Pona: The Language of Good?
**Answer (Object):** Toki Pona
**Reverse Question:** ...
**Answer (Subject):** ...
**Explanation:** The answer "Toki Pona" is explicitly contained within the book title mentioned in the question. Therefore, it is trivial.

**Direct Question:** ...
**Answer (Object):** ...
**Reverse Question:** What is the name of the annual celebration for the date the National flag of Canada officially appeared?
**Answer (Subject):** National Flag of Canada Day
**Explanation:** The answer is trivial because it simply repeats the key terms ("National flag of Canada") found in the question.


**Direct Question:** What is the median household income in the United States according to the Census Bureau?
**Answer (Object):** $67,521
**Reverse Question:** ...
**Answer (Subject):** ...
**Explanation:** The answer is time-sensitive. Without a specific year (e.g., "in 2020"), this value changes over time and is not static.


**Direct Question:** What was the release date of Ice Age: Dawn of the Dinosaurs?
**Answer (Object):** July 1, 2009
**Reverse Question:** What film is the third installment in the Ice Age film series?
**Answer (Subject):** Ice Age: Dawn of the Dinosaurs
**Explanation:** This fails the structural criteria. The answer to the Direct Question ("July 1, 2009") is not included in the Reverse Question.

----

### **SUITABLE Examples**

**Direct Question:** ...
**Answer (Object):** ...
**Reverse Question:** What DVI variant is compatible with VGA?
**Answer (Subject):** DVI-I
**Explanation:** Although the answer shares the term "DVI" with the question, it requires specific technical knowledge to identify the exact variant ("I") compatible with VGA. It is non-trivial.


**Direct Question:** What was the GMV achieved by the Redbus India Private Limited company in 2018?
**Answer (Object):** 50 billion Indian Rupee
**Reverse Question:** ...
**Answer (Subject):** ...
**Explanation:** The answer is not time-sensitive because the context restricts the fact to a specific year (2018), making the value static.


**Direct Question:** With what batting average did Carl Furillo win the batting title?
**Answer (Object):** .344
**Reverse Question:** Who won the 1953 batting title with a .344 average?
**Answer (Subject):** Carl Furillo
**Explanation:** This is suitable. Even though the Direct Answer (.344) is not the main subject of the Reverse Question, it is included in the Reverse Question to resolve ambiguity.

----

### **Required Output Format**

Respond only with "SUITABLE" or "NO".

**Instruction End**

**Please process the following direct and reverse questions and their answers:**

**Direct Question:** {direct}
**Answer (Object):** {object_entity}
**Reverse Question:** {reverse}
**Answer (Subject):** {subject_entity}
\end{lstlisting}
\end{prompt}

\begin{prompt}[label={box:prompt_natural}]{Goldenrod}{Prompt for Generating the \textit{Natural} Questions}
\begin{lstlisting}[basicstyle=\ttfamily, breaklines=true]
You will be provided with a question and its answer. Your task is to rephrase the question.
The question provided to you is an unambiguous question that has a single, unique, and specific answer. It is:

*   **Self-contained:** It does not use pronouns or references that require outside context.
*   **Unambiguous and Specific:** Its phrasing is clear, avoids multiple interpretations, and specifies the level of detail required in the answer (e.g., "What year...", "What is the name of the condition...").
*   **Minimal:** It includes only the essential details needed to make the question unambiguous and ensure there is a single possible answer.

### **Natural Question**

Rephrase the provided question to sound natural and conversational, as if a person were genuinely asking.

**Requirements:**

*   The natural question must include all the critical information from the provided question, but rewritten in a natural way.
*   **Crucially, the natural question must remain unambiguous and lead to the exact same specific answer as the provided question.**
*   To ensure the answer is specific, you might need to add clarifying phrases. For example, if the answer is a date, a vague question like "When did X happen?" must be rephrased as **"When did X happen? Answer with a date."** If the answer is a year, it would be **"When did X happen? Answer with a year."**
*   It is acceptable to use shorter, more common terms (e.g., Ebola instead of Ebola virus disease) or common abbreviations if they are widely understood (e.g., USA, NFL, LLM).
*   While you should generally not remove information, you **can** discard minor words or clauses if they are not necessary for disambiguation and a plausible person would consider the canonical and casual questions to be asking for the exact same thing. This should be done sparingly, only when it significantly improves the naturalness of the question.

----

### **Required Output Format**

You must return *only* a string with the natural question.

----

### **Examples**

1.
*   **Question:** What was the condition that caused Erika Nordby to spend two hours without a heartbeat before she was revived?
*   **Answer:** hypothermia
*   **Output:**
What condition caused Erika Nordby to have no heartbeat for two hours?
*Explanation:* The phrase "before she was revived" is omitted as it is implied by the fact that she had "no heartbeat for two hours" and the context is about her survival. A casual speaker would likely not add this detail.

2.
*   **Question:** What was the release date of When I Look in Your Eyes by Diana Krall?
*   **Answer:** June 8, 1999
*   **Output:**
When did Diana Krall's album "When I Look in Your Eyes" come out? Answer with a date.

3.
*   **Question:** What name did Sauron use in his fair disguise to manipulate Celebrimbor?
*   **Answer:** Annatar
*   **Output:**
What name did Sauron use as part of his disguise to trick Celebrimbor?
*Explanation:* The term "manipulated" is rephrased to the more common "trick." The sentence structure is changed to a more natural flow, but the key disambiguating elements "Sauron," "Celebrimbor," and "disguise" are all retained.

4.
*   **Question:** What is the name the press often uses to refer to the rule forbidding the horse-collar tackle from the NFL?
*   **Answer:** The Roy Williams Rule
*   **Output:**
What is the NFL rule that bans horse-collar tackles called?
*Explanation:* This is an example of carefully removing a clause. The phrase "the press often uses to refer to" is simplified to "what is the common name for". A natural speaker asking this question would assume the common name for the rule is the the name the press call the rule, making both questions the same.

**Instruction End**

**Please process the following question and answer:**

**Question:** {question}
**Answer:** {answer}
\end{lstlisting}
\end{prompt}

\begin{prompt}[label={box:prompt_contextual}]{Goldenrod}{Prompt for Generating the \textit{Contextual} Questions}
\begin{lstlisting}[basicstyle=\ttfamily, breaklines=true]
You will be provided with a context, a question, and an answer. The answer also completes the last sentence of the provided context.
Your task is to augment the question with the context.
The question provided to you is an unambiguous question that has a single, unique, and specific answer. It is:

*   **Self-contained:** It does not use pronouns or references that require outside context.
*   **Unambiguous and Specific:** Its phrasing is clear, avoids multiple interpretations, and specifies the level of detail required in the answer (e.g., "What year...", "What is the name of the condition...").
*   **Minimal:** It includes only the essential details needed to make the question unambiguous and ensure there is a single possible answer.


### **Contextual Question**

Create a question that tests comprehension within the given context. This is achieved by combining the context with the question, making minor edits to enhance fluency and minimize redundancy.

The typical process for generating this question is:
1.  Take the entire context, excluding the final incomplete sentence that the answer completes.
2.  Append the provided question to the end.
3.  Minor rephrasing of the context or question may be needed to improve the flow and remove redundancy.


----

### **Required Output Format**

You must return *only* a string with the contextual question.

----

### **Examples**

1.
*   **Context:** Erika Nordby (born February 2000), also known as Baby Erika, Miracle Baby and Canada's Miracle Child, is a Canadian originally from Edmonton, Alberta known for having been revived after spending two hours without a heartbeat due to
*   **Question:** What was the condition that caused Erika Nordby to spend two hours without a heartbeat before she was revived?
*   **Answer:** hypothermia
*   **Output:** Erika Nordby (born February 2000), also known as Baby Erika, Miracle Baby and Canada's Miracle Child, is a Canadian originally from Edmonton, Alberta. What was the condition that caused Erika Nordby to spend two hours without a heartbeat before she was revived?


2.
*   **Context:** When I Look in Your Eyes is the fifth studio album by singer Diana Krall, released on
*   **Question:** What was the release date of When I Look in Your Eyes by Diana Krall?
*   **Answer:** June 8, 1999
*   **Output:** When I Look in Your Eyes is the fifth studio album by singer Diana Krall. What was the release date of When I Look in Your Eyes?
*Explanation:* We removed the singer's name from the question to improve fluency, as it already appears in the preceding context and its repetition would be redundant.


3.
*   **Context:** Celebrimbor is a fictional character in J. R. R. Tolkien's legendarium. In Tolkien's stories, Celebrimbor was an elven-smith who was manipulated into forging the Rings of Power by the Dark Lord Sauron, in a fair disguise and named
*   **Question:** What name did Sauron use in his fair disguise to manipulate Celebrimbor?
*   **Answer:** Annatar
*   **Output:** Celebrimbor is a fictional character in J. R. R. Tolkien's legendarium. In Tolkien's stories, Celebrimbor was an elven-smith who was manipulated into forging the Rings of Power by the Dark Lord Sauron. What name did Sauron use in his fair disguise to manipulate Celebrimbor?


4.
*   **Context:** The horse-collar tackle is a gridiron football maneuver in which a defender tackles another player by grabbing the back collar or the back-inside of an opponent's shoulder pads and pulling the ball carrier directly downward violently in order to pull his feet from underneath him. The technique is most closely associated with Pro Bowl safety Roy Williams. After being blamed for a series of major injuries in the 2004 season , the horse-collar tackle was banned from the NFL during the 2005 off-season. The rule forbidding it is often referred to in the press as
*   **Question:** What is the name the press often uses to refer to the rule forbidding the horse-collar tackle from the NFL?
*   **Answer:** The Roy Williams Rule
*   **Output:** The horse-collar tackle is a gridiron football maneuver in which a defender tackles another player by grabbing the back collar or the back-inside of an opponent's shoulder pads and pulling the ball carrier directly downward violently. The technique is most closely associated with Pro Bowl safety Roy Williams and is blamed for a series of major injuries in the 2004 season, leading to it being banned. What is the name the press often uses to refer to the rule forbidding the horse-collar tackle from the NFL?

**Instruction End**

**Please process the following context, question and answer:**

**Context:** {context}
**Question:** {question}
**Answer:** {answer}
\end{lstlisting}
\end{prompt}

\begin{prompt}[label={box:prompt_mc}]{Yellow}{Prompt for Generating Options for Multiple-Choice Questions}
\begin{lstlisting}[basicstyle=\ttfamily, breaklines=true]
You will be provided with a context, a question, and its gold answer. Your task is to generate three additional plausible but incorrect answers to create a complete multiple-choice question.
The primary goal is to create challenging distractors that are confusing but definitely incorrect.
The provided question should be used as is, and the provided gold answer must be the first option in your output.

----

### **Requirements**

Your task is to generate three incorrect but confusing answer options. These distractors should make the question challenging. Consider the following strategies when creating them:

*   **Plausibility is key:** The incorrect answers should seem like they *could* be correct. They must share the same type, category, and granularity as the gold answer (e.g., if the answer is a person's name, all options should be names of people, if the answer is a city, all options should be cities).
*   **Must be incorrect:** There must be no ambiguity. Only the provided gold answer can be the correct answer to the question. The distractors must be factually wrong in the given context.
*   **Use contextual hints:** The distractors can be based on other entities or details mentioned in the provided context. For example, if the context is a film's summary and the question asks for the director, a good distractor would be the name of the producer or a lead actor mentioned in the text.
*   **Use similarity:** The distractors can be based on entities with similar names, or concepts that are thematically related. For example, if the answer is "The Battle of Hastings," a distractor could be "The Battle of Stamford Bridge," which occurred in the same year.
*   **Use popular associations:** The distractors can be other very popular options that might come to mind for the general topic, but are incorrect for the specific question asked.

----

### **Required Output Format**

You must return only the four answer options, each on a new line.
*   The first line must be the original gold answer provided to you.
*   The following three lines must be the new incorrect but confusing answer options.
*   Do not include any numbering or lettering (e.g., A., B., C., D.).

Remember to return only the list of strings (each on a new line), without any additional text.

**Instruction End**

**Please process the following context, question, and gold answer:**

**Context:** {context}
**Question:** {question}
**Gold answer:** {gold_answer}
\end{lstlisting}
\end{prompt}

\begin{prompt}[label={box:prompt_answer_ner}]{Yellow}{Prompt for Generating the \textit{Reverse} Task}
\begin{lstlisting}[basicstyle=\ttfamily, breaklines=true]
You will be provided with a question and its corresponding answer.
Your task is to categorize the answer by assigning the most appropriate entity type from the list below:

*   **PERSON:** An individual human, whether real or fictional.
*   **ORGANIZATION:** A structured company, institution, or collective with a common purpose.
*   **LOCATION:** A specific geographical place, region, or feature on Earth.
*   **EVENT:** A notable occurrence or incident that happens at a particular time and place.
*   **WORK_OF_ART:** A specific creative work, such as a book, movie, song, or painting.
*   **PRODUCT:** A commercially produced good, food, service, or piece of technology.
*   **ORGANISM:** A non-human biological life form, including animals, plants, and microbes.
*   **SUBSTANCE:** A material with a distinct physical or chemical composition.
*   **DATE:** A specific point in time, including full dates, years, or historical periods.
*   **NUMERIC_VALUE:** A number representing a specific quantity, measurement, or other quantifiable value.
*   **CONDITION:** A specific state of being, typically a medical, physical, or operational status.
*   **TITLE:** An official position, role, award, or honorific held by a person or group.
*   **LANGUAGE:** Natural languages, dialects, writing systems, including phrases.
*   **GROUP:** A collective of people defined by a shared nationality, ethnicity, or cultural affiliation.
*   **LAW:** A specific, formally enacted rule, regulation, or statute created by a governing body.
*   **IDEA:** An abstract concept, belief system, named phenomenon, or cultural practice.
*   **CLASSIFICATION:** A specific type, genre, format, class or list used to categorize other entities.
*   **IDENTIFIER:** A unique name, code, or alias used as a formal reference or label for another entity
*   **OTHER:** A catch-all category for named entities that do not fit into any of the other defined types.

----

### **Required Output Format**

You must return only a string containing a single entity type (e.g., "GROUP", "DATE"). Do not include any additional text or punctuation.

**Instruction End**

Please categorize the answer to the following question:

**Question:** {question}
**Answer:** {answer}
\end{lstlisting}
\end{prompt}

\subsection{Response Instructions and Grader Prompts}
\label{app_sub:graders}

\begin{prompt}[label={box:responses}]{RoyalPurple}{Task Instructions}
\textbf{Completions:}
\begin{lstlisting}[basicstyle=\ttfamily, breaklines=true]
Respond only with the completion of the last sentence
\end{lstlisting}

\textbf{Completions CoT:}
\begin{lstlisting}[basicstyle=\ttfamily, breaklines=true]
Think step by step. Complete the last sentence of the text. End your response with Answer: [Your Completion]
\end{lstlisting}

\textbf{Questions:}
\begin{lstlisting}[basicstyle=\ttfamily, breaklines=true]
Respond only with the answer to the question
\end{lstlisting}

\textbf{Questions CoT:}
\begin{lstlisting}[basicstyle=\ttfamily, breaklines=true]
Think step by step. End your response with Answer: [Your Answer]
\end{lstlisting}

\textbf{Multiple-Choice:}
\begin{lstlisting}[basicstyle=\ttfamily, breaklines=true]
Respond only with the letter of the correct answer
\end{lstlisting}

\textbf{Multiple-Choice CoT:}
\begin{lstlisting}[basicstyle=\ttfamily, breaklines=true]
Think step by step. End your response with: Answer: [Letter]
\end{lstlisting}
\end{prompt}

\begin{prompt}[label={box:autorater_comp}]{Violet}{Autorater Prompt for Text Completion Tasks}
\begin{lstlisting}[basicstyle=\ttfamily, breaklines=true]
>>> SYSTEM:
You will be provided with an incomplete text (referred to as the `left context`), a gold entity that represents the expected continuation or completion (`gold completion`), and several predicted completions generated by a model. Your task is to evaluate each predicted completion in relation to the gold completion and the left context, and assign one of the following labels:
["CORRECT", "PARTIALLY", "INCORRECT", "OTHER"].

First, I will give examples of each grade, and then you will grade a new example.

----

The following are examples of CORRECT predicted completions.

```
Left context: The song "Walk This Way" was peaked at number 10 on the
Gold completion: Billboard Hot 100
Completion 1: on the Billboard Hot 100
Completion 2: Billboard hot 100 in early 1976.
Completion 3: prestigious U.S. chart, Billboard Hot 100 in 1977.
```

```
Left context: Frequently portions of the term would remain only upon a U.S. senator's death or resignation. From 1914 and the enforcement of the
Gold completion: Seventeenth Amendment to the United States Constitution
Completion 1: Seventeenth Amendment to the United States Constitution, senators have been popularly elected.
Completion 2: 17th Amendment, U.S.
Completion 3: Seventeenth Amendment
Completion 4: Amendment XVII
```

```
Left context: Nova Scotia lies in the mid-temperate zone, and although the province is almost surrounded by water, the climate is closer to
Gold completion: continental climate
Completion 1: continental than to maritime.
Completion 2: continental than to maritime because the prevailing winds are from the continent.
```

These predicted completions are all CORRECT because:
    - The predicted completion contains the important information of the gold completion.
    - The predicted completion does not contain any information that contradicts the gold completion or the left context.
    - The gold completion may appear anywhere within the predicted completion as long as it clearly plays the same semantic role as the focus of the completion.
    - Minor omissions to the gold completion are allowed if the missing information is either easily inferable from the left context or not critical to identifying the gold completion (e.g., dropping a redundant modifier or title).
    - Only the semantic meaning matters: The completion may contain a variant, abbreviation, or alternative name for the gold completion. Differences in capitalization, punctuation, spelling, or minor grammatical variation (e.g., articles, pluralization) are allowed.
    - Completions may contain small factual inaccuracies or hallucinated details, as long as they do not contradict the left context.
    - Hedging and guessing are permissible, provided that the gold completion is included and the response does not contradict it or the left context.

----

The following are examples of INCORRECT predicted completions.

```
Left context: The Grand Caravan was reported missing, and a search was made, ending with the wreckage being found the next day, with all
Gold completion: 10
Completion 1: nine occupants of the aircraft dead.
Completion 2: five people on board deceased.
Completion 3: 11
```

```
Left context:  Andrew Paul Gosden (born 10 July 1993) disappeared from Central London on
Gold completion: 14 September 2007
Completion 1: 19 March 2007, aged 14.
Completion 2: 2007-09-15
```

```
Left context:  The Toplica Uprising was a mass uprising by Serbian rebels against the Bulgarian occupation forces that took place in Bulgarian-occupied Serbia during the First World War. The revolt was supported by Serbian guerrilla fighters known as
Gold completion: Chetniks
Completion 1: Komitadjis
Completion 2: Komitadjis who were the enemies of the Chetnik bands
```

These predicted completions are all INCORRECT because:
    - A factual statement in the predicted completion contradicts the left context or the gold completion.
    - Incorrect statements that have some hedging (e.g., "it is possible that", "although i'm not sure, i think") are also considered incorrect.

----

The following are examples of PARTIALLY predicted completions.

```
Left context:  IgA vasculitis (HSP), previously known as Henoch-Schonlein purpura, is an autoimmune disease that most commonly affects children. With kidney involvement, there may be a loss of small amounts of blood and protein in the urine (hematuria and proteinuria), but this usually goes unnoticed; in a small proportion of cases, the kidney involvement proceeds to
Gold completion: chronic kidney disease
Completion 1: end-stage kidney disease
Completion 2: kidney disease
Completion 3: chronic kidney disease and kidney failure - This is CORRECT.
```

```
Left context:  Dopamine dysregulation syndrome (DDS) is a dysfunction of the reward system observed in some individuals taking dopaminergic medications for an extended length of time. It typically occurs in people with Parkinson's disease (PD) or restless legs syndrome (RLS) who have taken
Gold completion: dopamine agonist medications
Completion 1: dopaminergic medication
Completion 2: dopaminergic medication for more than six months.
```

```
Left context:  Franki Raffles (17 October 1955 - 6 December 1994) was an English feminist and a
Gold completion: documentary photographer
Completion 1: photographer
Completion 2: photographer and artist.
Completion 3: professional artist and photographer.
```

These predicted answers are all PARTIALLY because:
    - The predicted completion contains a more general or less specific version of the gold completion (e.g., "dopamine agonist medications" is a specific type of "dopaminergic medication").
    - The predicted completion identifies the core entity and the primary meaning of the gold completion, but omits a descriptive detail or modifier (e.g., "photographer" instead of "documentary photographer").
    - The predicted completion is closely related to the gold completion and represents a common or expected outcome or progression, even if not strictly identical in every case (e.g., "end-stage kidney disease" as a severe form often associated with "chronic kidney disease").

----

The following are examples of OTHER predicted completions.

```
Left context:  Dirk Koperlager van Erp (1862-1933) was a Dutch American artisan,
Gold completion: coppersmith
Completion 1: noted for his distinctive work.
Completion 2: designer, and teacher.
Completion 3: I don't know.
Completion 4: I need more context about Dirk Koperlager.
Completion 5: Without researching the web, I cannot answer this. However, I can tell you that Dirk Koperlager van Erp was an artisan.
Completion 6: who became known for his innovative metalwork, particularly his use of silver.
```

```
Left context:  T33 and CP3 are disability sport classification for disability athletics. Athletes in this class have moderate
Gold completion: quadriplegia
Completion 1: to severe hypertonia, ataxia, or athetosis
Completion 2: athetosis, ataxia, or dystonia.
Completion 3: or good functional strength and range of motion in their trunk
```

These predicted completions are all OTHER because:
    - The important information in the gold completion is not included in the predicted completion.
    - No statements in the predicted completion contradict the gold completion or the left context.

----

Also note the following things:
- For completions where the gold completion is a number, the predicted completion needs to be correct to the last significant figure in the gold completion. For example, consider the left context "The number of citations the Transformer paper has is?" with gold completion "120k".
    - Predictions "120k", "124k", and 115k" are all CORRECT.
    - Predictions "100k" and "113k" are INCORRECT.
    - Predictions "around 100k" and "more than 50k" are considered OTHER because they neither confirm nor contradict the gold completion.
- Do not punish predicted completions if they omit information that would be clearly inferred from the left context. For example, consider the left context: "The paper A pretrainer's guide to training data: Measuring the effects of data age, domain coverage, quality, & toxicity" won the NAACL '24", and the gold completion "Outstanding Paper Award". The predicted completion "Outstanding Paper" would be considered CORRECT, because "award" is presumed in the left context.
- Do not punish typos in people's names if it's clearly the same name.
    - For example, if the gold completion is "Hyung Won Chung", you can consider the following predicted completions as correct: "Hyoong Won Choong", "Hyungwon Chung", or "Hyun Won Chung".

----

**Required Output Format**

Your entire response must be structured using the following tags:
<completion_{i}>the grade of Completion {i}</completion_{i}>

>>> USER:
Here is a new example. Use the output format above to grade each completion with either "CORRECT", "INCORRECT", "PARTIALLY", or "OTHER". Don't apologize or correct yourself if there was a mistake; we are just trying to grade the completions.

```
Left context:  {left_context}
Gold completion: {gold_target}
{completions}
\end{lstlisting}
\end{prompt}

\begin{prompt}[label={box:autorater_qa}]{Periwinkle}{Autorater Prompt for Closed-book QA Tasks}
\begin{lstlisting}[basicstyle=\ttfamily, breaklines=true]
>>> SYSTEM:
You will be provided with a question, a gold target, and several predicted answers generated by a model. Your task is to evaluate each predicted answer in relation to the gold target, and assign one of the following labels:
["CORRECT", "PARTIALLY", "INCORRECT", "OTHER"].

First, I will give examples of each grade, and then you will grade a new example.

----

The following are examples of CORRECT predicted answers.

```
Question: What chart did Aerosmith's "Walk This Way" hit number 10 on?
Gold target: Billboard Hot 100
Answer 1: On the Billboard Hot 100
Answer 2: Billboard hot 100 in early 1976.
Answer 3: The prestigious U.S. chart, Billboard Hot 100 in 1977.
```

```
Question: What was enforced from 1914 that affected the election of U.S. senators?
Gold target: Seventeenth Amendment to the United States Constitution
Answer 1: Seventeenth Amendment to the United States Constitution, senators have been popularly elected.
Answer 2: 17th Amendment, U.S.
Answer 3: Seventeenth Amendment
Answer 4: Amendment XVII
```

```
Question: Nova Scotia lies in the mid-temperate zone. What type of climate is Nova Scotia's climate closer to, despite the province being almost surrounded by water?
Gold target: continental climate
Answer 1: Continental than to maritime.
Answer 2: Continental than to maritime because the prevailing winds are from the continent.
```

These predicted answers are all CORRECT because:
    - The predicted answer contains the important information of the gold target.
    - The predicted answer does not contain any information that contradicts the gold target or the question.
    - The gold target may appear anywhere within the predicted answer as long as it clearly plays the same semantic role as intended by the question.
    - Minor omissions to the gold target in the predicted answer are allowed if the missing information is either easily inferable from the question or not critical to identifying the gold target (e.g., dropping a redundant modifier or title).
    - Only the semantic meaning matters: The predicted answer may contain a variant, abbreviation, or alternative name for the gold target. Differences in capitalization, punctuation, spelling, or minor grammatical variation (e.g., articles, pluralization) are allowed.
    - Predicted answers may contain small factual inaccuracies or hallucinated details, as long as they do not contradict the question.
    - Hedging and guessing are permissible, provided that the gold target is included and the predicted answer does not contradict it or the question.

----

The following are examples of INCORRECT predicted answers.

```
Question: How many people were on Bering Air Flight 445 when it crashed on February 6, 2025?
Gold target: 10
Answer 1: Nine occupants of the aircraft dead.
Answer 2: Five people on board deceased.
Answer 3: 11
```

```
Question:  Andrew Paul Gosden (born 10 July 1993) disappeared from Central London. On what date did this disappearance occur?
Gold target: 14 September 2007
Answer 1: 19 March 2007, aged 14.
Answer 2: 2007-09-15
```

```
Question:  What was the name of the Serbian guerrilla fighters who supported the Toplica Uprising?
Gold target: Chetniks
Answer 1: Komitadjis
Answer 2: Komitadjis who were the enemies of the Chetnik bands
```

These predicted answers are all INCORRECT because:
    - A factual statement in the predicted answer contradicts the question or the gold target.
    - Incorrect answers that have some hedging (e.g., "it is possible that", "although i'm not sure, i think") are also considered incorrect.

----

The following are examples of PARTIALLY predicted answers.

```
Question:  What is a rare but serious kidney complication of IgA vasculitis?
Gold target: chronic kidney disease
Answer 1: End-stage kidney disease
Answer 2: Kidney disease
Answer 3: Chronic kidney disease and kidney failure - This is CORRECT.
```

```
Question:  What drugs can lead to Dopamine dysregulation syndrome in patients with Parkinson's or restless legs syndrome?
Gold target: dopamine agonist medications
Answer 1: Dopaminergic medication
Answer 2: Dopaminergic medication for more than six months.
```

```
Question:  What was Franki Raffles' profession?
Gold target: documentary photographer
Answer 1: Photographer
Answer 2: Photographer and artist.
Answer 3: Professional artist and photographer.
```

These predicted answers are all PARTIALLY because:
    - The predicted answer contains a more general or less specific version of the gold target (e.g., "dopamine agonist medications" is a specific type of "dopaminergic medication").
    - The predicted answer identifies the core entity and the primary meaning of the gold target, but omits a descriptive detail or modifier (e.g., "photographer" instead of "documentary photographer").
    - The predicted answer is closely related to the gold target and represents a common or expected outcome or progression, even if not strictly identical in every case (e.g., "end-stage kidney disease" as a severe form often associated with "chronic kidney disease").

----

The following are examples of OTHER predicted completions.

```
Question:  Dirk Koperlager van Erp (1862-1933) was a Dutch American artisan. What type of artisan was he?
Gold completion: coppersmith
Answer 1: Noted for his distinctive work.
Answer 2: Designer, and teacher.
Answer 3: I don't know.
Answer 4: I need more context about Dirk Koperlager.
Answer 5: Without researching the web, I cannot answer this. However, I can tell you that Dirk Koperlager van Erp was an artisan.
Answer 6: He became known for his innovative metalwork, particularly his use of silver.
```

```
Question:  What condition is described as moderate in athletes in the T33 and CP3 disability sport classification?
Gold completion: quadriplegia
Answer 1: Severe hypertonia, ataxia, or athetosis
Answer 2: Athetosis, ataxia, or dystonia.
Answer 3: Good functional strength and range of motion in their trunk
```

These predicted completions are all OTHER because:
    - The important information in the gold target is not included in the predicted answer.
    - No statements in the predicted answer contradict the gold target or the question.

----

Also note the following things:
- For answers where the gold target is a number, the predicted answer needs to be correct to the last significant figure in the gold target. For example, consider the question "What is the number of citations the Transformer paper has?" with gold target "120k".
    - Predictions "120k", "124k", and 115k" are all CORRECT.
    - Predictions "100k" and "113k" are INCORRECT.
    - Predictions "around 100k" and "more than 50k" are considered OTHER because they neither confirm nor contradict the gold target.
- Do not punish predicted answers if they omit information that would be clearly inferred from the question. For example, consider the left context: "What NAACL '24 award did the paper A pretrainer's guide to training data: Measuring the effects of data age, domain coverage, quality, & toxicity" won?", and the gold target "Outstanding Paper Award". The predicted answer "Outstanding Paper" would be considered CORRECT, because "award" is presumed in the question.
- Do not punish typos in people's names if it's clearly the same name.
    - For example, if the gold target is "Hyung Won Chung", you can consider the following predicted answers as correct: "Hyoong Won Choong", "Hyungwon Chung", or "Hyun Won Chung".

----

**Required Output Format**

Your entire response must be structured using the following tags:
<answer_{i}>the grade of Answer {i}</answer_{i}>

>>> USER:
Here is a new example. Use the output format above to grade each answer with either "CORRECT", "INCORRECT", "PARTIALLY", or "OTHER". Don't apologize or correct yourself if there was a mistake; we are just trying to grade the answers.

```
Question:  {question}
Gold target: {gold_target}
{answers}
\end{lstlisting}
\end{prompt}

\twocolumn


\end{document}